\documentclass{article} 
\usepackage{iclr2024_conference,times}

\usepackage{amsmath,amsfonts,bm}

\def\eqref#1{equation~\ref{#1}}

\def\1{\bm{1}}

\DeclareMathAlphabet{\mathsfit}{\encodingdefault}{\sfdefault}{m}{sl}
\SetMathAlphabet{\mathsfit}{bold}{\encodingdefault}{\sfdefault}{bx}{n}

\DeclareMathOperator*{\argmin}{arg\,min}

\usepackage{hyperref}
\usepackage{url}

\usepackage{amsmath,amsfonts,amssymb,amsthm, mathtools}
\usepackage{cleveref}
\usepackage{multirow}
\usepackage{adjustbox}
\usepackage{wrapfig}
\usepackage{booktabs}
\usepackage{algorithm}
\usepackage{algpseudocode}

\newtheorem{theorem}{Theorem}

\title{Understanding Reconstruction Attacks with the Neural Tangent Kernel and Dataset Distillation}

\author{Noel Loo, Ramin Hasani, Mathias Lechner, Alexander Amini and Daniela Rus \\
MIT CSAIL\\
Cambridge, Massachussetts, USA \\
\texttt{\{loo, rhasani, mlechner, amini, rus\}@mit.edu} \\
}

\iclrfinalcopy 
\begin{document}

\maketitle

\begin{abstract}
  Modern deep learning requires large volumes of data, which could contain sensitive or private information that cannot be leaked. Recent work has shown for homogeneous neural networks a large portion of this training data could be reconstructed with only access to the trained network parameters. While the attack was shown to work empirically, there exists little formal understanding of its effective regime which datapoints are susceptible to reconstruction. In this work, we first build a stronger version of the dataset reconstruction attack and show how it can provably recover the \emph{entire training set} in the infinite width regime. We then empirically study the characteristics of this attack on two-layer networks and reveal that its success heavily depends on deviations from the frozen infinite-width Neural Tangent Kernel limit. Next, we study the nature of easily-reconstructed images. We show that both theoretically and empirically, reconstructed images tend to ``outliers'' in the dataset, and that these reconstruction attacks can be used for \textit{dataset distillation}, that is, we can retrain on reconstructed images and obtain high predictive accuracy.
\end{abstract}
\section{Introduction}


Neural networks have been shown to perform well and even generalize on a range of tasks, despite achieving zero loss on training data \citep{rethinking_generalization, still_rethinking}. But this performance is useless if neural networks cannot be used in practice due to security issues. A fundamental question in the security of neural networks is how much information is \emph{leaked} via this training procedure, that is, can adversaries with access to trained models, or predictions from a model, infer what data was used to train the model? Ideally, we want to ensure that our models are resistant to such attacks. However, in practice, we see that this ideal is commonly violated. One heinous violation of this principle is the phenomenon of memorization \citep{memorization_p1, why_do_they_memorize, do_we_need_to_memorize, carlini_memorize_1}, where trained networks can be shown to replicate their training data at test time in generative models. A more extreme example of memorization is presented in \citet{reconstruction}, where the authors show that it is possible to recover a large subset of the training data given only the trained network parameters.

The existence of this attack begs many follow-up questions: ``\textit{Under what circumstances is this attack successful?}"; and ``\textit{What are properties of these recovered images?}" In this paper, we consider a stronger variant of the attack presented in \citet{reconstruction}, provide novel theoretical and empirical insights about dataset reconstruction attacks, and provide answers to the above questions. In particular, we make the following new \textbf{contributions}:

\textbf{We design a stronger version of \citet{reconstruction}'s dataset reconstruction attack} that can provably reconstruct the \textit{entire} training set for networks in the neural tangent kernel (NTK) \citep{jacotntk} regime when trained under mean squared error (MSE) loss. This attack transfers to finite networks with its success dependent on deviations from the NTK regime.

\textbf{We show that outlier datapoints are prone to reconstruction} under our attack, corroborating prior work observing this property. Additionally, we show that removing easily reconstructed images can improve predictive accuracy.

\textbf{We formally prove and empirically show that a dataset reconstruction attack is a variant of \textit{dataset distillation}.} The reconstruction loss is equal to the loss of the kernel-inducing points (KIP)  \citep{KIP1, KIP2} dataset distillation algorithm, under a different norm, plus a variance-controlled term. Furthermore, we can retrain models using recovered images and achieve high performance. 

\section{Background and Related Works}

\textbf{Machine Learning Privacy.} A large body of work studies how to extract sensitive information from trained models. This is problematic as legislation such as HIPAA and GDPR enforce what data can and cannot be published or used \citep{hipaa, gdpr}. Quantifying the influence of training examples leads to the topic of influence functions \citep{understanding_black_bow_with_influence}, and Membership-inference attacks \citep{MIA, mia_first_principles}, which try to infer whether particular examples were used in training. Defending against these attacks is the study of \textit{differential privacy}, which quantifies and limits the sensitivity of models to small changes in training data \citep{Original_DP_paper, deep_learning_with_DP_OG_paper}. Likewise, the field of \emph{machine unlearning} tries to remove the influence of training examples post-training \citep{unlearning}. Without defense techniques, trained networks have been shown to leak information \citep{privacy_attacks_survey}. For generative language models, \citep{carlini_memorize_1} show large language models often reproduce examples in the training corpus verbatim. \textit{Model inversion} techniques aim to recreate training examples by looking at model activations \citep{inversion_1, inversion_2, inversion_3}. This memorization phenomenon can be shown to be necessary to achieve high performance under certain circumstances \citep{why_do_they_memorize,memorization_necessary}.

\textbf{Dataset Reconstruction.}
Recent work \citep{reconstruction} has shown that one can reconstruct a large subset of the training data from trained networks by exploiting the implicit biases of neural nets. They note that homogeneous neural networks trained under a logistic loss converge in direction to the solution of the following max-margin problem \citep{kkt_1, kkt_2}:
\begin{align}
\argmin_{\theta'}\frac{1}{2}\|\theta'\|^2_2 \quad \text{s.t.} \quad \forall i\in [n],  y_i f_{\theta'}(x_i) \geq 1,
\label{eq:haim_max-margin}
\end{align}
Where $\{x_i, y_i\}$ is the training set with images $x_i$ and labels $y_i \in \{+1, -1\}$, and $f_{\theta'}(x)$ the neural network output with parameters $\theta'$.
\citep{reconstruction} shows that by taking a trained neural network and optimizing images (and dual parameters) to match the Karush–Kuhn–Tucker (KKT) conditions of the max-margin problem, it is possible to reconstruct training data. This is an attack that causes leakage of training data. Here, we consider a stronger variant of the attack that requires training under mean-squared error (MSE) loss.

\textbf{Neural Tangent Kernel.} To investigate the generalization in neural networks we can use the neural tangent kernel (NTK) theory \citep{jacotntk, Arora_Exact_NTK_calc}. NTK theory states that networks behave like first-order Taylor expansions of network parameters about their initialization as network width approaches infinity \citep{wide_linear_models}. Furthermore, the resulting feature map and kernel converge to the NTK, and this kernel is frozen throughout training \citep{jacotntk, Arora_Exact_NTK_calc}. As a result, wide neural networks are analogous to kernel machines, and when trained with MSE loss using a support set $X_S$ with labels $y_S$ result in test predictions given by:
\[\hat{y}_T = K_{TS}K_{SS}^{-1}y_S,\]
with $K$ being the NTK. For fully-connected networks, this kernel can be computed exactly very quickly (as they reduce to arc-cosine kernels), but for larger convolutional networks, exact computation slows down dramatically \citep{Arora_Exact_NTK_calc, NTK_features_via_sketching}. In practice, it has been shown that networks often deviate far from the frozen-kernel theoretical regime, with the resulting empirical NTKs varying greatly within the first few epochs of training before freezing for the rest \citep{Hanin2020Finite, finite_dynamcis2, fortman, loo2022evolution, nyu_peeps_paper}. In this paper, we use the NTK theory to gain a better understanding of these reconstruction attacks.

\textbf{Dataset Distillation.} Dataset distillation aims to construct smaller synthetic datasets which accurately represent larger datasets. Specifically, training on substantially smaller \emph{distilled dataset} achieves performance comparable to the full dataset, and far above random sampling of the dataset \citep{wang2018dataset, zhao2021DC, zhao2021dsa, KIP1, KIP2, frepo, RFAD}. There are many algorithms for this, ranging from methods that directly unroll computation \citep{wang2018dataset}, try to efficiently approximate the inner unrolled computation associated with training on distilled data \citep{frepo, RFAD, KIP2}, and other heuristics \citep{zhao2021DC, zhao2021dsa}. One algorithm is kernel-induced points (KIP) \citep{KIP1, KIP2}, which leverages NTK theory to derive the following loss:
\[\mathcal{L}_{KIP} = \frac{1}{2}\|y_t - K_{TS}K_{SS}^{-1}y_S\|^2_2.\]
The loss indicates the prediction error of infinite width networks on distilled images $X_S$ and labels $y_S$, which are then optimized. We bring up dataset distillation as we show in this paper that our dataset reconstruction attack is a generalization of KIP, and that dataset distillation can be used to defend against the attack.

\section{A Neural Tangent Kernel Reconstruction Attack}
\label{sec:attack_description}

\citet{reconstruction} considers the scenario where the attacker only has access to the final trained network parameters. This attack requires that the networks are homogeneous and are trained for many epochs until convergence so that the network converges in direction to the final KKT point of Eq. \ref{eq:haim_max-margin}. While it is a good proof-of-concept for such attacks, there are several theoretical and practical limitations of this attack. We find that the attack presented in \citet{reconstruction} is brittle. Namely, we were unable to reliably reproduce their results without careful hyperparameter tuning, and \textbf{careful network initialization strategies}. Their attack also requires training until directional convergence, which \textbf{requires network parameters to tend to infinity}, and requires homogenous networks. Furthermore, outside of the top few reconstructions, the overwhelming majority of reconstructions ($>70\%$) are of poor quality (more in-depth discussions in \cref{app:comparison_to_haim}). Here we present an attack which is compatible with early stopping, does not require special initialization strategies, and can provable reconstruct the \textit{entire training set} under certain assumptions, the \textbf{first guarantee of reconstruction in any regime}. However, our attack requires access to the model initialization or a previous training checkpoint. Access to the initialization or earlier training checkpoints arises naturally in many settings, such as fine tuning from public models, or in federated learning where clients receive period updates of the model parameters.

Note that we cannot compare our attack to other ones such as gradient leakage attacks \citep{gradient_leakage}, membership inference attacks \citep{MIA} and generative model attacks \citep{llm_generative_attack}, as these attacks either require gradient access (which also requires parameter access) in the setting on gradient leakage, specific query data points in the case of membership inference, or a generative model and query points for generative model attacks. Our attack requires no a priori knowledge of the dataset. With this context in mind, the attack desribed in this paper contributes to the literature on parameter-only based attacks.

Consider a neural network trained under MSE loss, $\mathcal{L} = \frac{1}{2}\sum_{i = 0}^{N-1} (y_i - f_\theta(x_i))^2$, for $x_i, y_i \in X_T, y_T$, being the training set datapoints and labels. Now further assume that the network is trained under gradient flow and that the network is approximately in the lazy/NTK regime, that is, it behaves like a first-order Taylor expansion of the network outputs \citep{lazy_training}: 
\begin{align}
    f_\theta(x) \approx f_{lin, \theta}(x) = f_{\theta_0}(x) + (\theta - \theta_0)^\intercal \nabla_\theta f_{\theta_0}(x) \label{eq:linearized_dyanmics}
\end{align}

\citet{wide_linear_models} shows that the time evolution of the network parameters in this regime is given by:
{\small
    \begin{align*}
        \theta(t) = \theta_0 -\nabla_\theta f_{\theta_0}(X_T)^\intercal K_0^{-1} \left(I - e^{-\eta K_0 t}\right)(f_{\theta_0}(X_T) - y_T)
    \end{align*}
}
With $\eta$ the learning rate and $K_0$ the finite-width/empirical NTK evaluated at $\theta_0$. Namely, the final change in parameters is given by:
\begin{align}
    \Delta \theta = \theta_f - \theta_0 = \nabla_\theta f_{\theta_0}(X_T)^\intercal K_0^{-1}(y_T - f_{\theta_0}(X_T))
    \label{eq:recon_solution_to_opt_problem}
\end{align}
Notably, this is the solution to the following optimization problem:
\begin{align}
    \argmin_{\Delta \theta} \frac{1}{2} \|\Delta \theta\|^2_2 \quad &\text{s.t.} \quad \Delta\theta^\intercal\nabla_\theta f_{\theta_0}(X_T) = y_T - f_{\theta_0}(X_T) \label{eq:recon_opt_problem}
\end{align}
The corresponding KKT conditions are:
\begin{align}
    \Delta \theta &= \alpha^\intercal \nabla_\theta f_{\theta_0}(X_T) \label{eq:kkt_1} \\ 
    \Delta\theta^\intercal \nabla_\theta f_{\theta_0}(X_T) &= y_T - f_{\theta_0}(X_T) \label{eq:kkt_2}
\end{align}
With $\alpha$ being the set of dual parameters. In our formulation, \cref{eq:kkt_1} ensures we are at a stationary point, while \cref{eq:kkt_2} ensures that the labels are correct. Like with \citep{reconstruction}, we can directly optimize the reconstruction images and dual parameters ($X$ and $\alpha$, respectively) to match these KKT conditions, given a network's final parameters and initialization to get $\Delta \theta$. In practice, we only need to optimize \cref{eq:kkt_1}, for reasons we will describe next, leading to our reconstruction loss:
\begin{align}
    \mathcal{L}_{\textrm{Reconstruction}} = \| \Delta \theta - \alpha^\intercal \nabla_\theta f_{\theta_0}(X_T)\|^2_2 \label{eq:recon_loss}
\end{align}

\begin{figure*}[t]
\vskip 0.1in
\begin{center}
\includegraphics[width = 0.7 \linewidth]{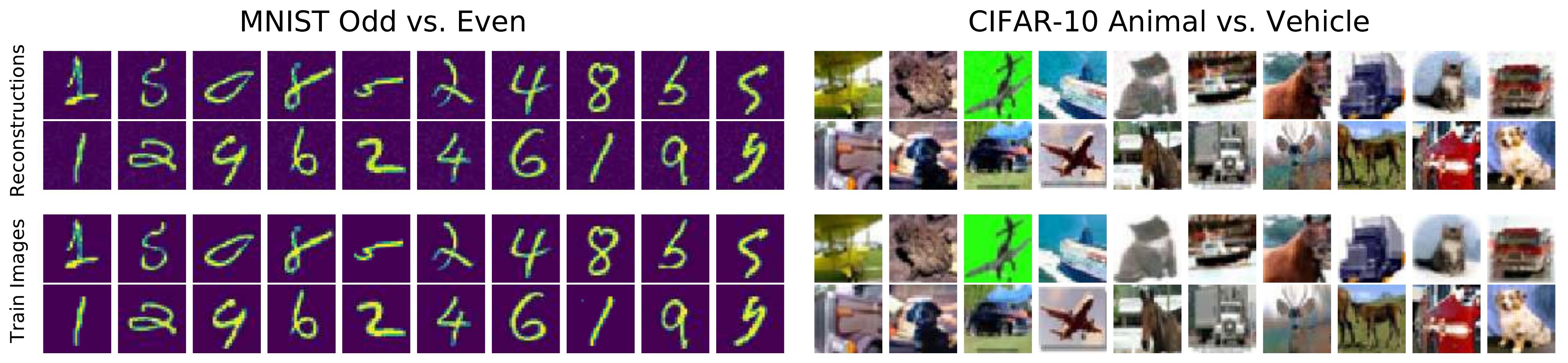}
\vskip -0.1in
\caption{Reconstructed images (top) vs closest training images (bottom) for MNIST Odd vs. Even, and CIFAR-10 Animal vs. Vehicle Classification. Reconstructions were made from 4096-width two hidden layer fully-connected networks trained with standard dynamics and low learning rates. Apart from small amounts of noise, the original images and their reconstructions are visually indistinguishable.}
\label{fig:reconstruction_main_banner}
\end{center}
\vskip -0.2in
\end{figure*}

\begin{figure*}[t]
\begin{center}
\includegraphics[height=0.7in]{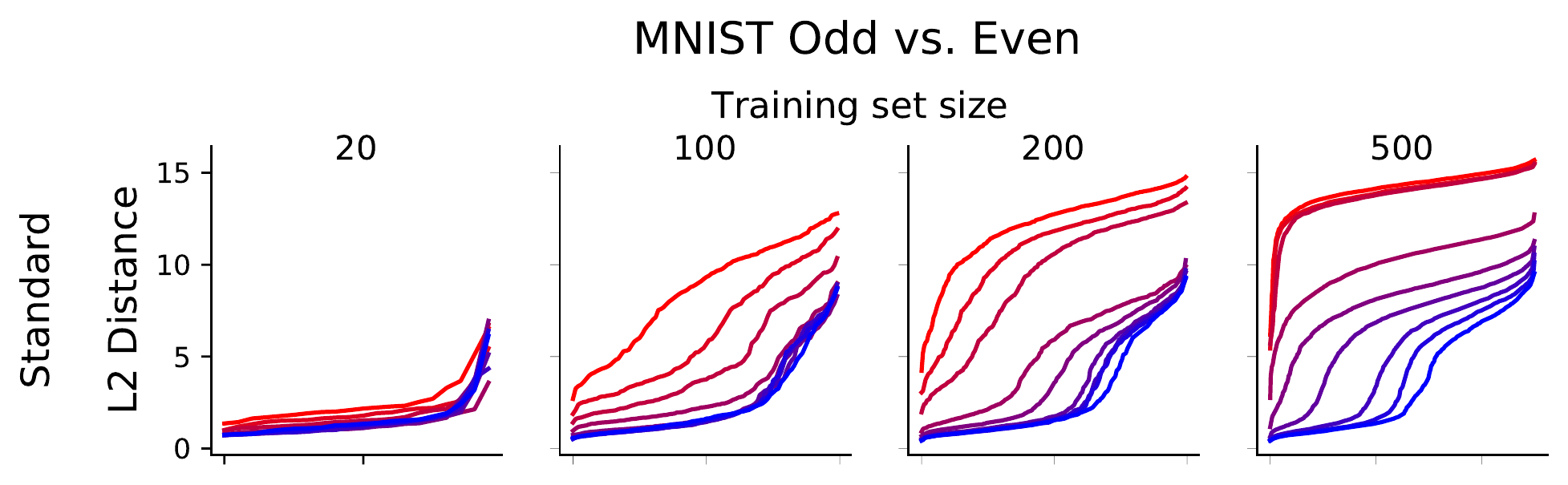}
\includegraphics[height=0.7in]{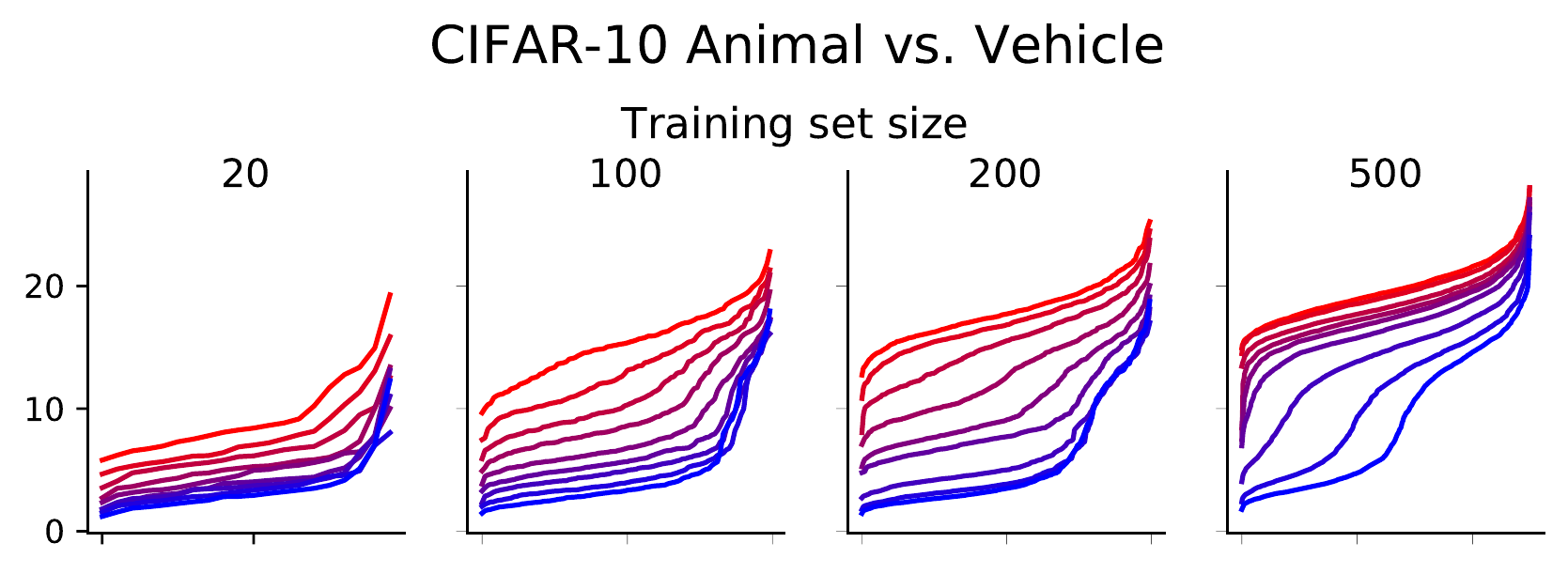}
\includegraphics[height=0.64in]{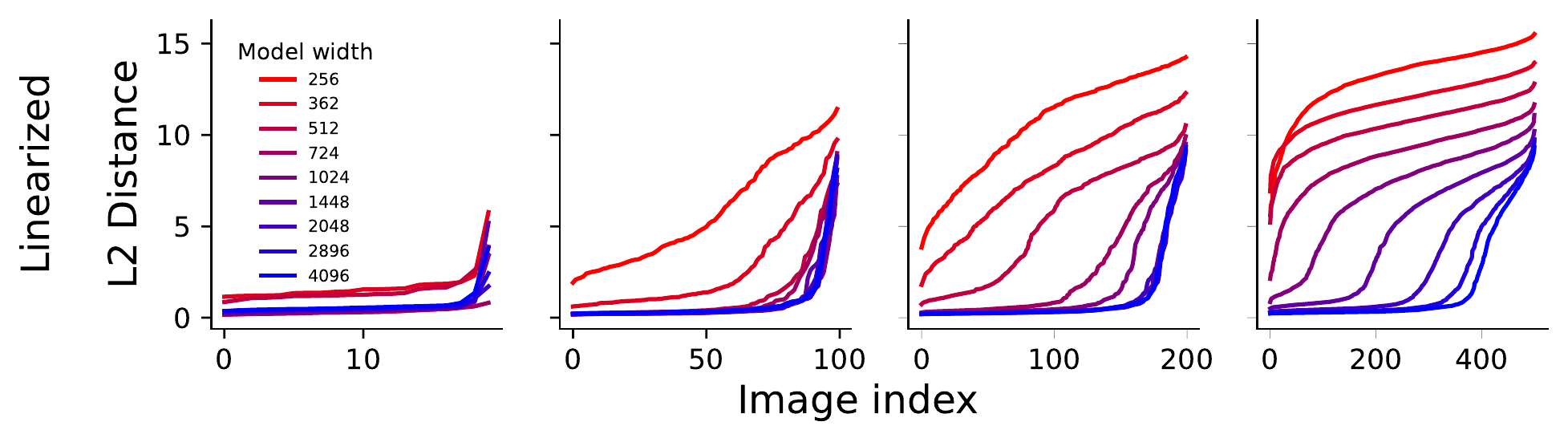}
\includegraphics[height=0.64in]{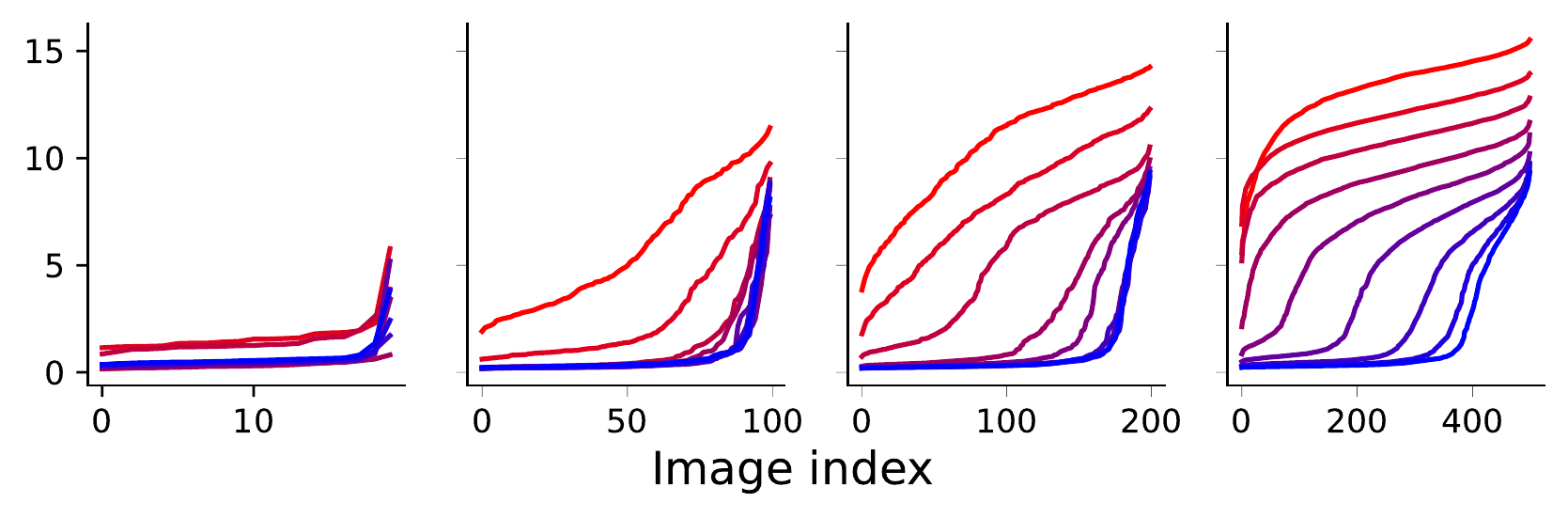}
\vskip -0.1in
\caption{Reconstruction quality curves for the MNIST Odd vs. Even and for CIFAR-10 Animal vs. Vehicle classification. We either used standard dynamics (top) or linearized dynamics (bottom) and varied both the size of the training set and model width. Smaller datasets are easier to reconstruct while wider models can reconstruct more images, with linearization helping in both scenarios.}
\label{fig:reconstruction_curve_main}
\end{center}
\vskip -0.2in
\end{figure*}

\textbf{Reconstruction in Infinite Width.}
\label{sec:infinite_width_attack}
Next, we show that this formulation of the attack recovers the \textit{entire} training set for infinite-width models. We further assume that the training data lies on the unit hypersphere. 

\begin{theorem}
\label{thm:reconstruct_entire}
If $\mathcal{L}_{\textrm{reconstruction}} = 0$ (from Eq. \ref{eq:recon_loss}), then we reconstruct the entire training set in the infinite-width limit, assuming that training data lies on the unit hypersphere.
\end{theorem}

\begin{proof}
Define $k_{\theta}(x, x') = \nabla_{\theta}f_\theta(x)^\intercal \nabla_{\theta}f_\theta(x')$, that is, the finite-width/empirical NTK function. We know as network width $w \to \infty$, $\Delta \theta = \sum_{\alpha_i, x_i \in \alpha^T, X_{T}}{\alpha_i \nabla_{\theta_0}f_{\theta_0}(x_i)}$, with $\alpha^T = K_{\theta_0, TT}^{-1}y_T$, with $X_T$ being the training set, $y_T$ the training labels, and $K_{\theta_0, TT}$ the finite-width NTK evaluated on the training set. Our attack then becomes:
{\small \begin{align}
    &\quad \Big\|\Delta \theta - \sum_{\mathclap{\alpha_j x_j \in \alpha^R, X_R}} \alpha_j \nabla_{\theta_f} f_{\theta_f} (x_j)\Big\|^2_2 =\Big\|\sum_{\mathclap{\alpha_i, x_i \in \alpha^T, X_{T}}}{\alpha_i \nabla_{\theta_0}f(x_i)} - \sum_{\mathclap{\alpha_j x_j \in \alpha^R, X_R}} \alpha_j \nabla_{\theta_f} f_{\theta_f} (x_j)\Big\|^2_2 \\
    &= \Big\|\sum_{\mathclap{\alpha_i, x_i \in \alpha^T, X_{T}}}{\alpha_i k_{\theta_0}(x_i, \cdot)} - \sum_{\mathclap{\alpha_j x_j \in \alpha^R, X_R}} \alpha_j k_{\theta_f}(x_j, \cdot)\Big\|^2_2 
\end{align}}
With $T$ and $R$ referring to the training and reconstruction set, respectively. As $w \to \infty$ we know that $k_{\theta_0}, k_{\theta_f} \to k_{NTK}$.
Furthermore, define 
{\small \begin{align*}
    P_T = \sum_{\alpha_i, x_i \in \alpha^T, X_{T}}{\alpha_i\delta(x_i)},~~~~~~~P_R = \sum_{\alpha_j, x_j \in \alpha^R, X_{R}}{\alpha_j\delta(x_j)}
\end{align*}}
as measures associated with our trained network and reconstruction, respectively, and $\mu_* = \int_\Omega k_{NTK}(x, \cdot) dP_*(x)$, with $\Omega = S^d$, with $d$ being the data dimension (Assuming data lies on the unit hypersphere). $\mu_T$ and $\mu_R$ are now kernel embeddings of our trained network and reconstruction, respectively. Our reconstruction loss becomes:
$\|\mu_T - \mu_R\|^2_{\mathcal{H}_{NTK}}$. This is the maximum-mean discrepancy (MMD) \citep{MMD}. We note that $P_T, P_R$ are signed Borel measures (since $\alpha$ are finite and our reconstruction/training sets are on the unit sphere). The NTK is universal over the unit sphere \citep{jacotntk}, implying that the map $\mu:\{\textrm{Family of signed Borel measures}\} \to \mathcal{H}$ is injective \citep{universality_of_measures}, meaning that we are able to \textbf{recover the entire training set}, provided that $\alpha_i \neq 0$, which happens almost surely (see \cref{app:label_conditions}). 
\end{proof}

Note that in practice we do not enforce the unit sphere requirement on the data, and we still see high reconstruction quality, which we show in \cref{sec:finite_width_attack}. This mapping from the network tangent space to image space also sheds light on the success of gradient leakage attacks \citep{gradient_leakage}, in which gradients are used to find training batch examples.

\section{Dataset Reconstruction for Finite Networks}
\label{sec:finite_width_attack}
While the attack outlined in Theorem \ref{thm:reconstruct_entire} carries fundamental theoretical insights in the infinite-width limit, it has limited practicality as it requires access to the training images themselves to compute the kernel inner products. How does the attack work for finite-width neural networks, and under what circumstances is this attack successful?

\begin{table*}[t!]
\centering
\caption{Performance of KIP, Recon-KIP (RKIP), RKIP from a trained network (RKIP-finite), on distilling 500 images down to 20 images. KIP and RKIP provide the best infinite-width performance, while KIP fails for finite models. (n=7) 
}
 \begin{adjustbox}{width=0.9\textwidth}
\begin{tabular}{lcccccc}
\toprule
\multirow{2}{*}{Distillation Algorithm} & \multicolumn{3}{c}{MNIST Odd/Even}       & \multicolumn{3}{c}{CIFAR-10 Animal/Vehicle}  \\ \cmidrule{2-7}
                                        & Standard & Linearized & Infinite Width & Standard & Linearized & Infinite Width                           \\ \midrule
Full dataset (500 images)	&	$92.85 \pm 0.42$	&	$92.91 \pm 0.33$	&	$93.18 \pm 0.37$	&	$75.06 \pm 0.21$	&	$74.60 \pm 0.21$	&	$75.42 \pm 0.28$	\\
KIP	&	$57.42 \pm 8.41$	&	$55.62 \pm 7.48$	&	$\mathbf{91.53 \pm 0.57}$	&	$35.26 \pm 5.67$	&	$32.37 \pm 3.60$	&	$70.98 \pm 0.43$	\\
RKIP	&	$\mathbf{89.61 \pm 1.18}$	&	$\mathbf{89.99 \pm 1.11}$	&	$91.44 \pm 0.48$	&	$\mathbf{72.23 \pm 3.61}$	&	$\mathbf{72.76 \pm 3.74}$	&	$\mathbf{74.66 \pm 0.93}$	\\
RKIP-finite	&	$88.45 \pm 0.89$	&	$86.15 \pm 3.39$	&	$87.31 \pm 3.24$	&	$71.96 \pm 1.14$	&	$63.99 \pm 4.02$	&	$62.05 \pm 4.17$	\\
Random images	&	$73.52 \pm 3.60$	&	$73.54 \pm 3.61$	&	$74.12 \pm 3.73$	&	$70.36 \pm 2.53$	&	$70.18 \pm 2.54$	&	$70.77 \pm 2.04$	\\
\bottomrule
\end{tabular}
\end{adjustbox}
\label{tab:kip_rkip_performances}
\vskip -0.1in
\end{table*}

\begin{wrapfigure}[15]{r}{0.6\textwidth}
\vspace{-10mm}
\begin{center}
\includegraphics[width = 0.45\linewidth]{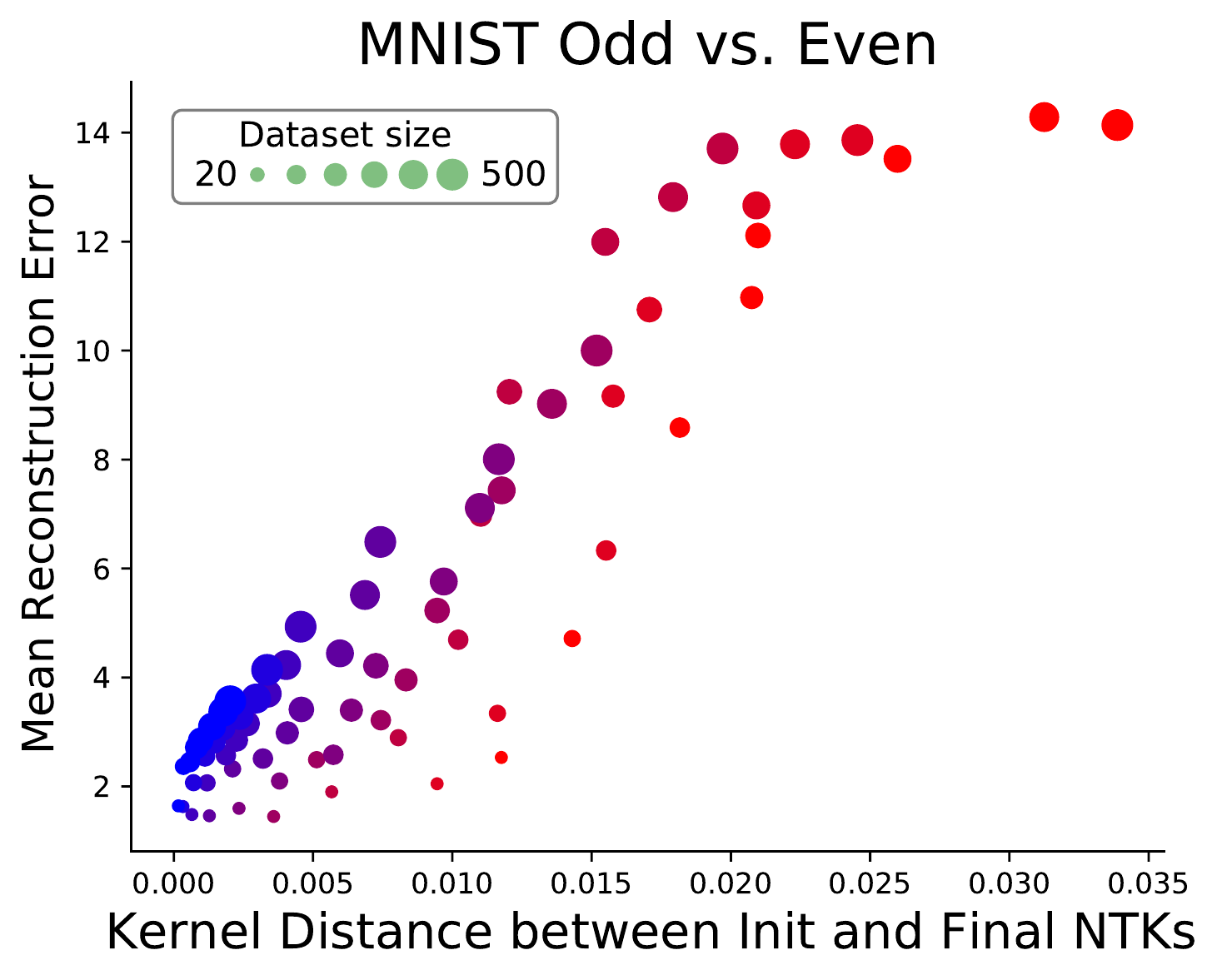}
\includegraphics[width = 0.45\linewidth]{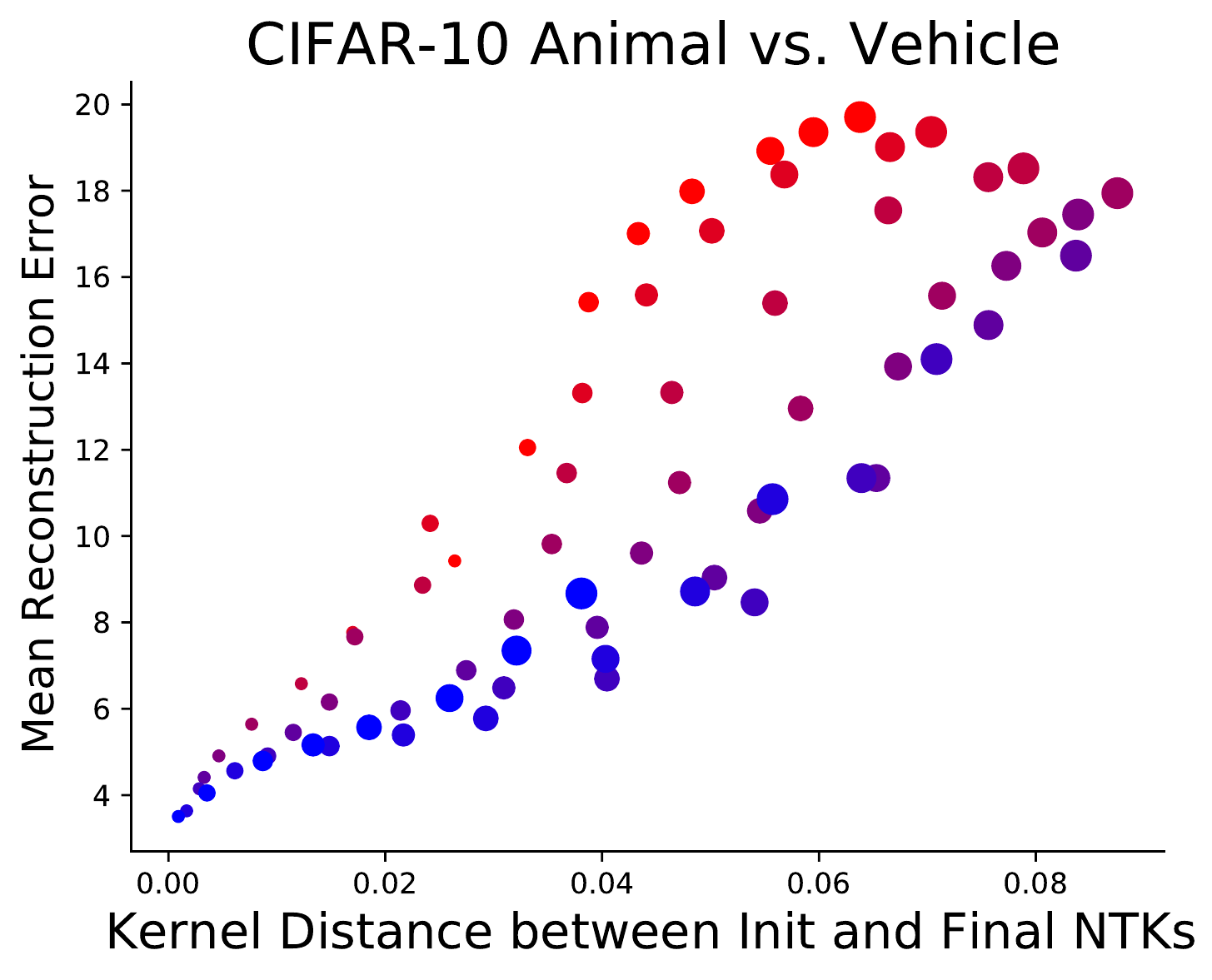}
\vskip -0.1in
\caption{Mean reconstruction error vs. the kernel distance from the initialization to the final kernel. The mean reconstruction error, measured as the average value of the reconstruction curve, is strongly correlated with how much the finite-width NTK evolves over training. Dataset size is given by dot size, while the color indicates model width (see \cref{fig:reconstruction_curve_main}).}
\label{fig:reconstruction_kernel_dist_scatter}
\end{center}
\vspace{10mm}
\end{wrapfigure}
To answer these questions, we follow the experimental protocol of \citet{reconstruction}, where we try to recover images from the MNIST and CIFAR-10 datasets on the task of odd/even digit or animal/vehicle classification for MNIST and CIFAR-10, respectively. We vary the size of the training set from 10 images per class to 250 images per class (500 total training set size). We consider two hidden layer neural networks with biases using standard initialization (as opposed to NTK parameterization or the initialization scheme proposed in \citet{reconstruction}). We vary the width of the neural networks between 256 and 4096 to see how deviations from the infinite-width regime affect the reconstruction quality. Furthermore, it is known that for finite-width networks the finite-width NTK varies over the course of training, deviating from the infinite-width regime. We can force the kernel to be frozen by considering \emph{linearized} training, where we train a first-order Taylor expansion of the network parameters around its initialization (see \cref{eq:linearized_dyanmics}). We consider both networks under standard (unmodified) dynamics and linearized dynamics. In \cref{app:conv_results} and \cref{app:timagenet_results} we consider convolutional architectures and high-resolution datasets, respectively, but we restrict our attention to MLPs on lower-resolution images in the main text.

We train these networks for $10^6$ iteration using full-batch gradient descent with a low learning rate, and during the reconstruction, we make $M = 2N$ reconstructions with $N$ being the training set size. A full description of our experimental parameters is available in \cref{app:experiment_details} and algorithmic runtime details in \cref{app:alg_details}. To measure reconstruction quality we consider the following metric. We first measure the squared $L_2$ distance in pixel space from each training image to each reconstruction. We select the pair of training images and reconstruction which has the lowest distance and remove it from the pool, considering it pair of image/reconstruction. We repeat this process until we have a full set of $N$ training images and reconstructions  (See Fig. \ref{fig:reconstruction_main_banner}). We then order the $L_2$ distances into an ascending list of distances and plot this function. We call this the \emph{reconstruction curve} associated with a particular reconstruction set. We plot these reconstruction curves for varying dataset sizes and model widths in \cref{fig:reconstruction_curve_main}. From \cref{fig:reconstruction_curve_main} we have the following three observations:

\textbf{Smaller training sets are easier to reconstruct.} We see that the reconstruction curve for smaller datasets has low values for all model widths.
\textbf{Wider models can resolve larger datasets.} We observe that for a given model width, there is a threshold image index at which the quality of reconstructions severely decreases. For example, for MNIST Odd/Even, 200 images and a width of 1024, this is 80 images. As we increase the model width this threshold increases almost monotonically.

\begin{wrapfigure}[12]{r}{0.6\textwidth}
\vspace{-8mm}
\vskip 0.1in
\begin{center}
\includegraphics[height=1.1in]{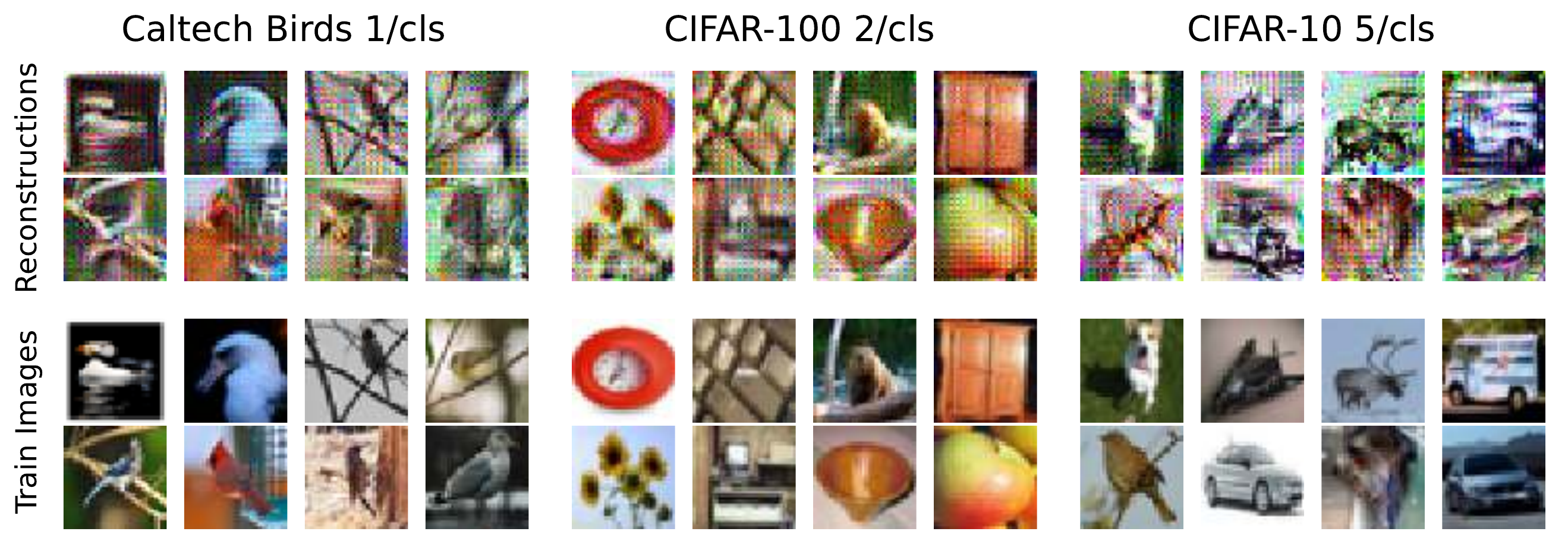}
\caption{Reconstructions of training data for few-shot fine tuning on a ResNet-18 pretrained on ImageNet on Caltech Birds (1/cls), CIFAR-100 (2/cls) and CIFAR-10 (5/cls).}
\label{fig:fine_tune_recon}
\end{center}
\vskip -0.2in
\end{wrapfigure}
\textbf{Linearization improves reconstruction quality.} We see that linearized networks can resolve more images and have better images compared to their same-width counterparts. The success of linearization suggests that deviations from the frozen NTK regime affect reconstruction quality. We can measure the deviation from the frozen kernel regime by measuring the \emph{kernel distance} of the network's initialization NTK and its final NTK, given by the following: $d(K_0, K_f) = 1 - \frac{\textrm{Tr}(K_0^\intercal K_f)}{\|K_0\|_F \|K_f\|_F}$.

Intuitively, this distance tells us how well the initialization and final kernel align. Large values indicate that the kernel has changed substantially, meaning the network is deviating far from the NTK regime. We plot these kernel distances against the mean value of the reconstruction curves in figure \cref{fig:reconstruction_kernel_dist_scatter}. We see immediately that reconstruction quality is strongly correlated with kernel distance, and that smaller datasets and wider models have a lower kernel distance. In \cref{app:early_stopping_xent}, we discuss how our attack is \textbf{compatible with early stopping and cross-entropy loss}, unlike \citep{reconstruction}, which requires training until convergence. A more detailed discussion of the effect of early stopping is available in \cref{app:early_stopping_xent}.

\begin{figure}[t!]
\begin{center}
\includegraphics[width = 0.45\linewidth]{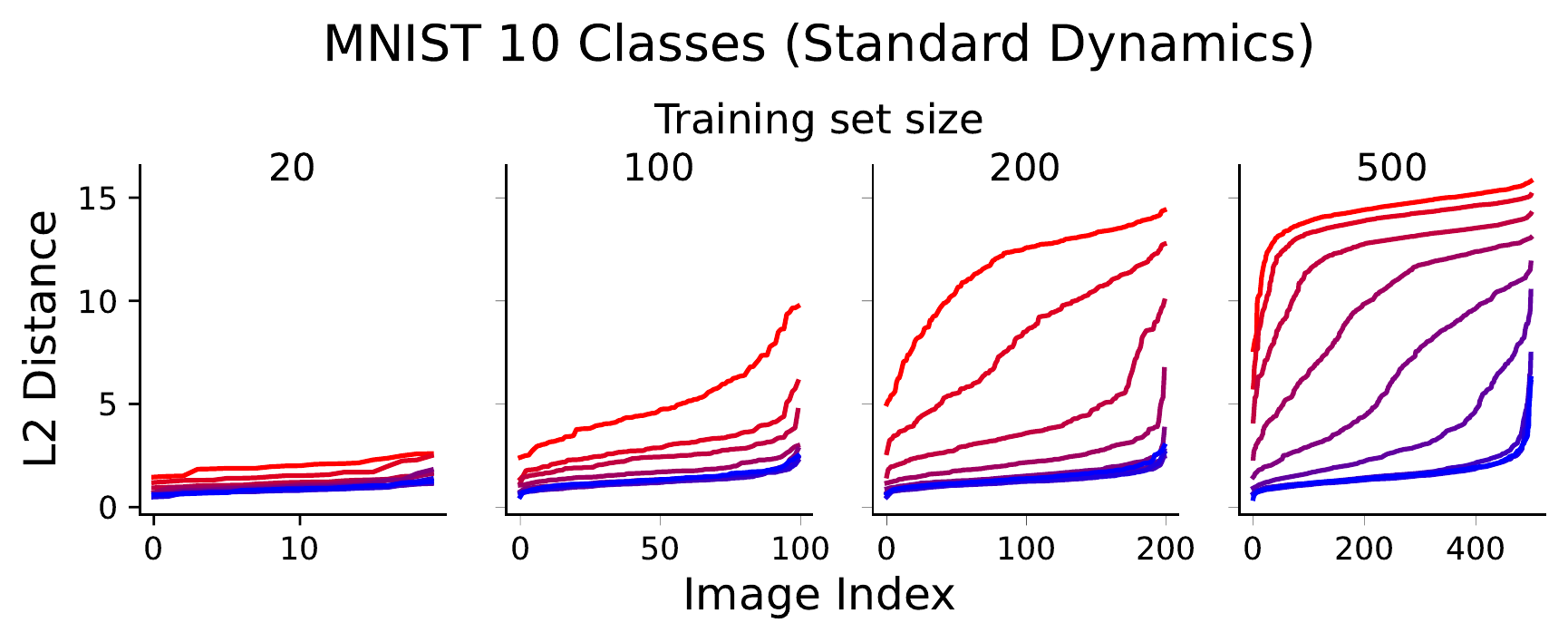}
\includegraphics[width = 0.45\linewidth]{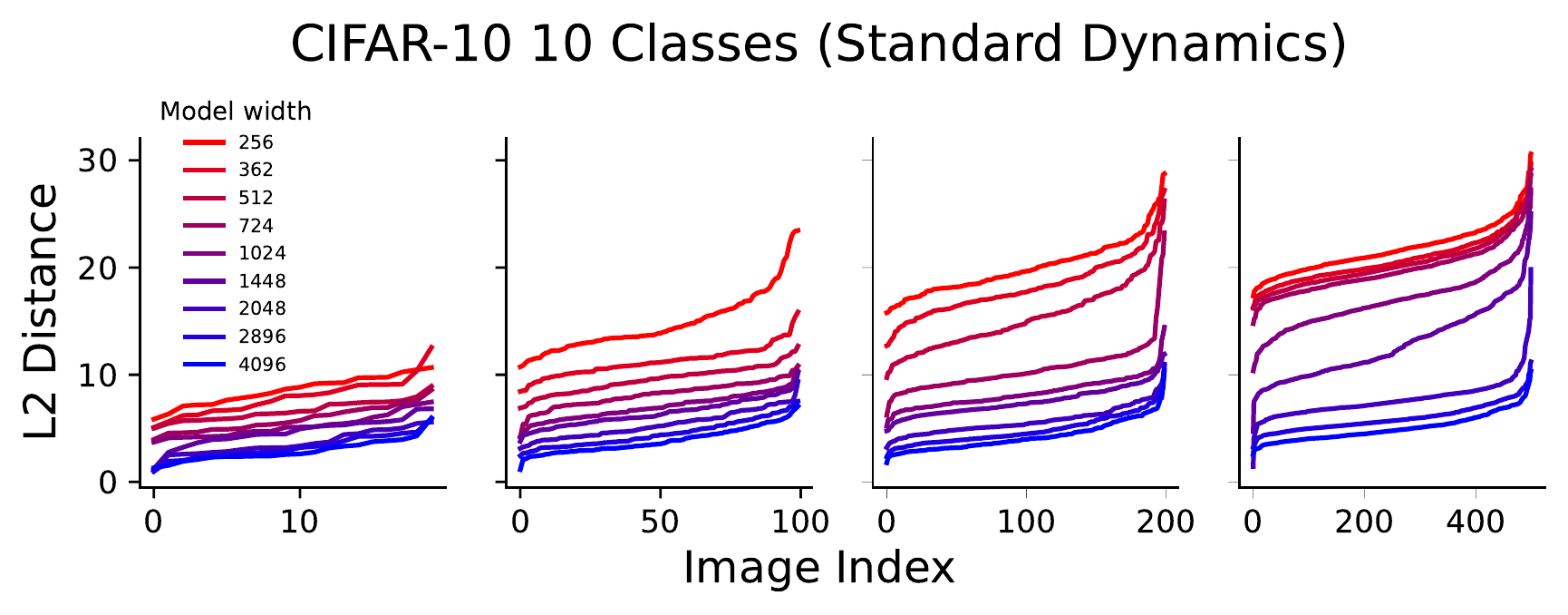}
\vskip -0.1in
\caption{Reconstruction curves for networks trained on multiclass MNIST/CIFAR-10. 
}
\label{fig:reconstruction_curve_multiclass}
\end{center}
\vskip -0.4in
\end{figure}


\textbf{Multiclass Classification.} In previous sections, we showed the validity of the attack on binary classification. Here we verify that the attack works with multiple classes. We repeat the same procedure as in \cref{sec:finite_width_attack}, but will all 10 classes. Details are given in \cref{app:alg_details}, as well as additional results on 200-way classification on Tiny-ImageNet in \cref{app:timagenet_results}. Results are shown by reconstruction curves in \cref{fig:reconstruction_curve_multiclass}. We observe that this attack has \textbf{improved} reconstruction quality with more classes. In \cref{app:kernel_distance_scatter_multiclass}, we observe that multi-class classification leads to lower kernel distances, suggesting it behaves more in the kernel regimes, explaining the better reconstruction quality. Future work could investigate this further.

\section{Dataset Reconstruction in Fine Tuning}
\label{sec:fine_tuning}
A key requirement of the attack is the model initialization. When training from scratch, attackers will not have access to this, making the attack useless. However, practitioners often do not train from scratch, but rather fine-tune large publicly available pre-trained models. Furthermore, users often do not have access to large amounts of data, effectively making the task few-shot. With evidence suggesting that training neural networks later during fine-tuning is well approximated by the frozen finite-NTK theory \citep{fine_tuning_linear, continual_learning_linear, meta_learning_ntk, fine_tuning_ntk_llm}, this makes the few-shot fine-tuning setting an easy target for this attack. To evaluate our attack in this setting, we fine-tuned publically available ResNet-18s pretrained on ImageNet on few-shot image classification on Caltech Birds (200 classes) \citep{cub_200}, CIFAR-100, CIFAR-10 with 1, 2, and 5 samples per class, respectively. For $\theta_0$, we use the initial fine-tuned model parameters. We see the best reconstructions in \cref{fig:fine_tune_recon}. We see that our attack is able to recover some training images but with limited quality. Future work could look at improving these attacks.

\section{What Datapoints are susceptible to Reconstruction?}
\label{sec:which_get_attacked}
It has been observed in previous work that no datapoints are equally susceptible to privacy attacks \citep{privacy_onion, influence_outliers, mia_first_principles, dp_accuracy_disparate}. In particular, \textit{outlier} images tend to be leaked more easily than others. In this section, we show that this occurs for our attack, and provide theoretical justification for this.

\begin{wrapfigure}[16]{r}{0.5\textwidth}
\vspace{-5mm}
\begin{center}
\includegraphics[height=1.0in]{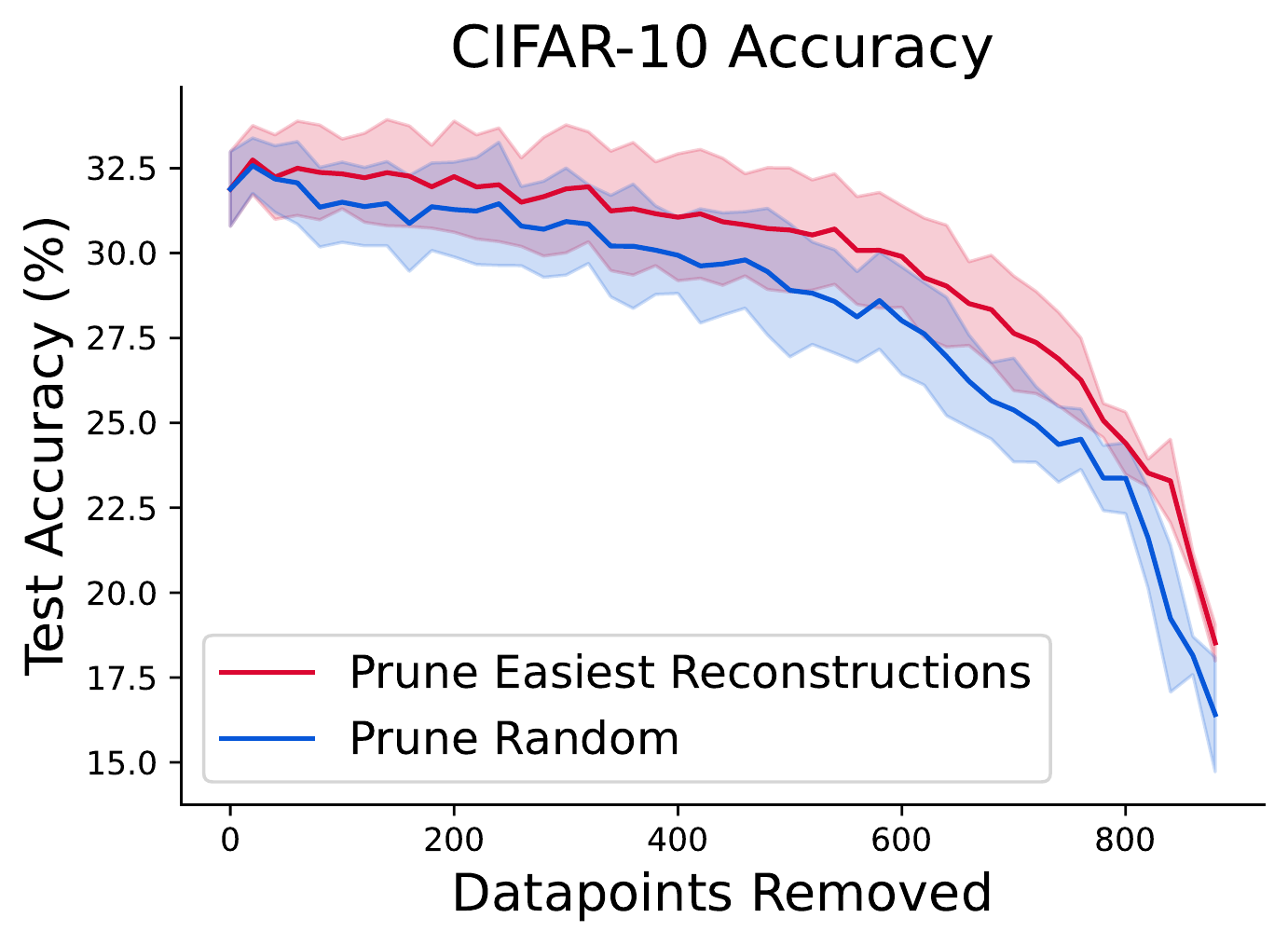}
\includegraphics[height=1.0in]{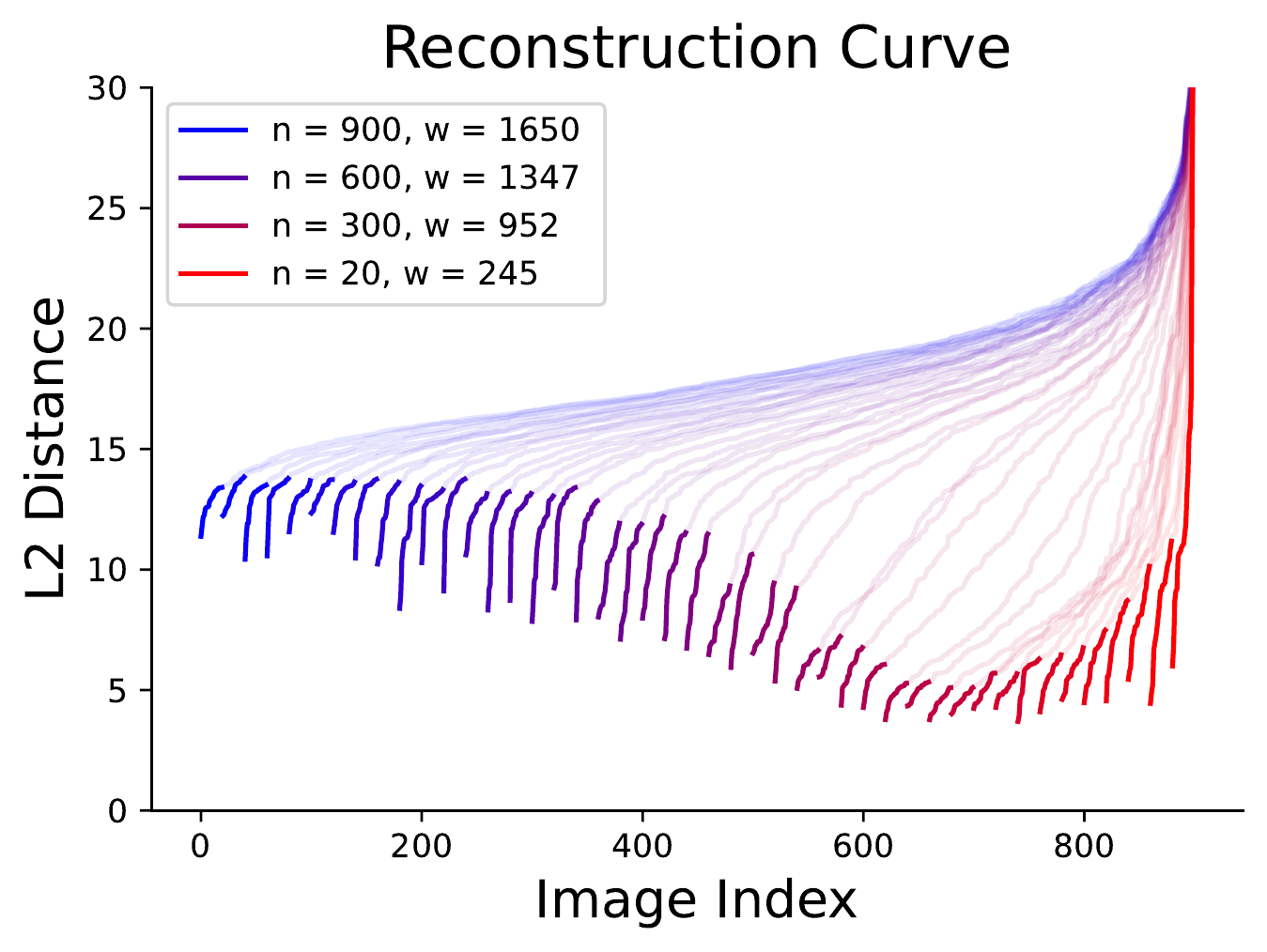}
\vskip -0.1in
\caption{Test accuracy of iteratively pruned CIFAR-10 using either random pruning or pruning based on easily reconstructed images (Left), and reconstruction curves for pruned CIFAR-10 (Right). We see that easily reconstructed datapoints can be removed without harming accuracy. We observe a privacy ``onion" effect where removing easily reconstructed images reveals other images which are easy to reconstruct. Bolded lines indicate the 20 images removed after pruning.}
\label{fig:onion_figure}
\end{center}
\vskip -0.2in
\end{wrapfigure}
\textbf{6.1~~~Hard to fit Implies Easy to Reconstruct}

\begin{figure*}[t]
\begin{center}
\includegraphics[height=1.05in]{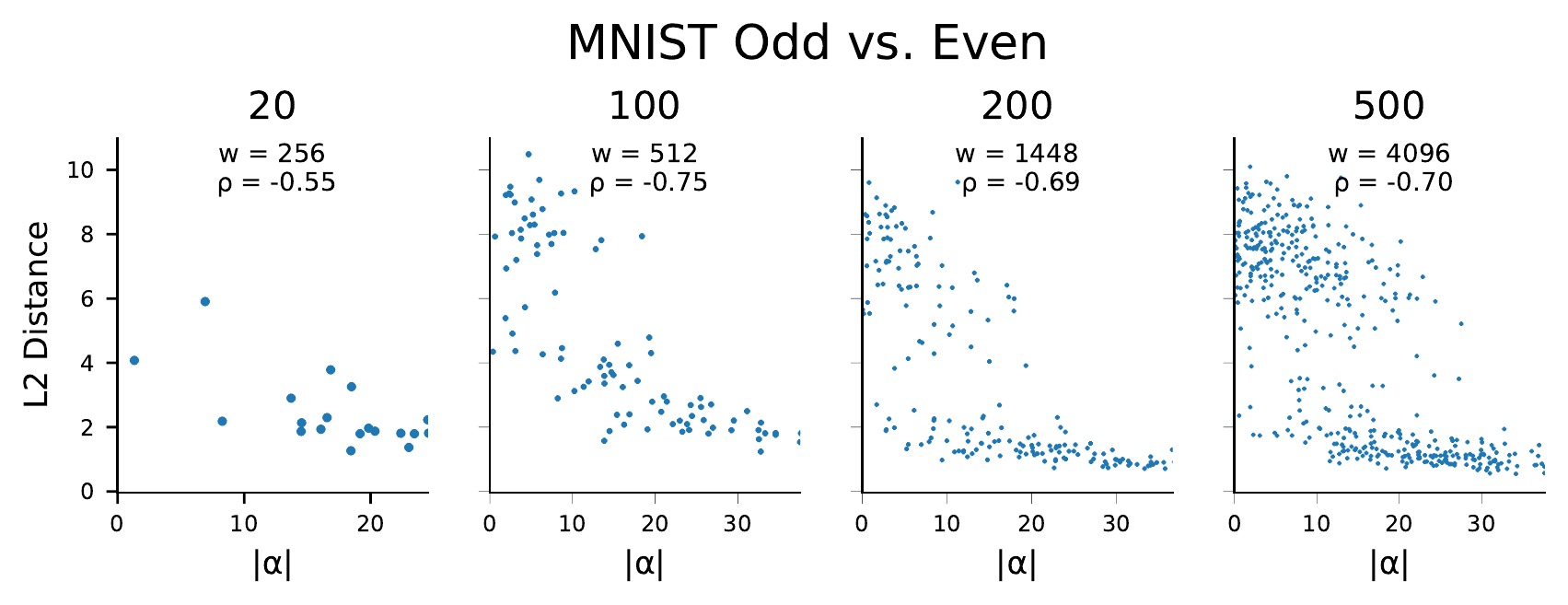}
\includegraphics[height=1.05in]{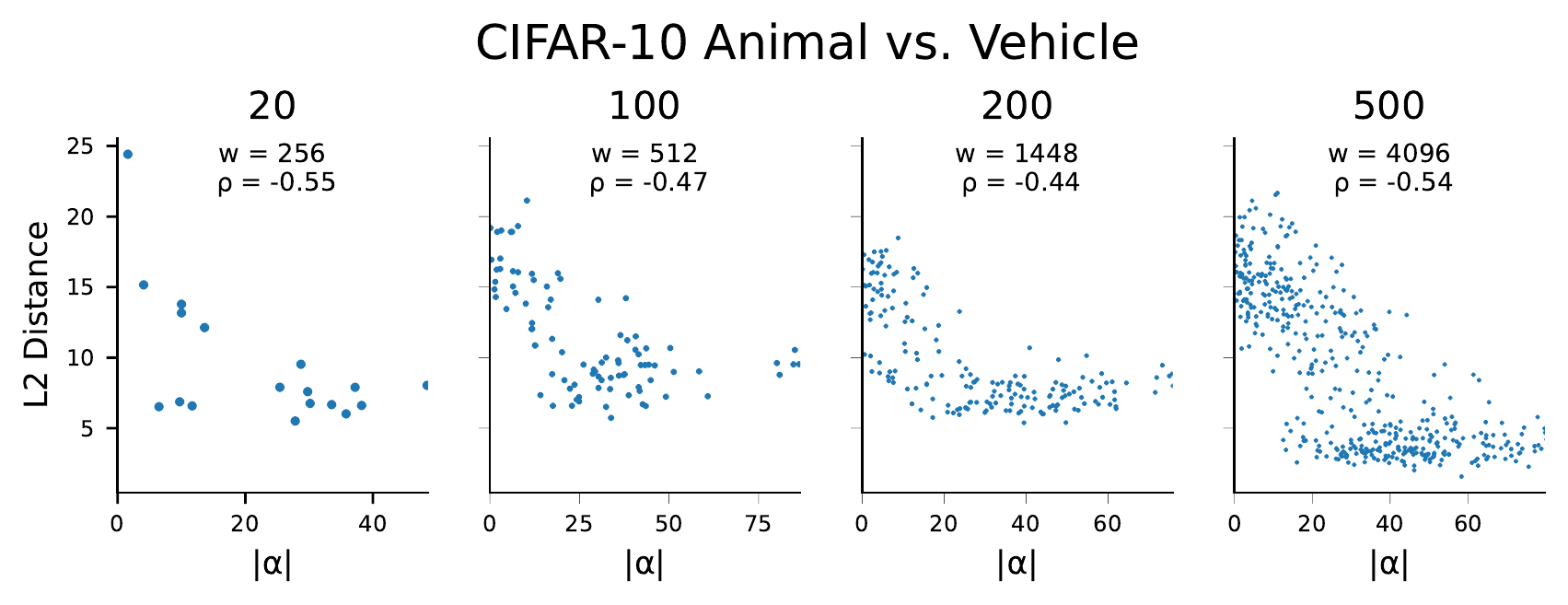}
\vskip -0.1in
\caption{Scatter plots of reconstruction quality measured in l2 distance and corresponding $|\alpha|$ values for images. We vary the width with $n$ so that we observe a range of reconstruction qualities. $|\alpha|$ is negatively correlated with reconstruction error.}
\label{fig:alpha_scatter}
\end{center}
\vskip -0.2in
\end{figure*}

By considering our reconstruction loss $\Big\|\Delta \theta - \sum_{j} \alpha_j \phi(x_j)\Big\|^2_2 $ with $\phi(x_j) = \nabla_{\theta_f} f_{\theta_f} (x_j)$ we aim to learn a basis to ``explain" $\Delta \theta$, we see that this could be cast as a sparse coding problem. Assuming that all $\phi(x_j)$ are of roughly the same magnitude, we expect the the parameters which larger $\alpha$ parameters to contribute more to $\Delta \theta$, and thus be more easily reconstructed. This is closely related to how data points with high influence are likely to be memorized \citep{influence_outliers}. We verify this heuristic holds empirically by plotting a scatter plot of reconstruction error vs. the corresponding $|\alpha|$ values calculated for infinite width in \cref{fig:alpha_scatter}. We see that images with small $\alpha$ values are ``protected" from reconstruction since their contribution to $\Delta \theta$ is small and could be written off as noise. From \cref{app:label_conditions}, we know that $\alpha = 0$ corresponds to an image/label which does not alter the prediction at all, so this suggests that well-predicted datapoints are safe from reconstruction. Aside from the definition of $\alpha = K^{-1}y$, we can alternatively show that $\alpha$ is closely related to how \textit{quickly} the model fits that datapoint. We can write (see \cref{app:alpha_as_loss} for a derivation) that $\alpha_j = \int_0^\infty (y_j - f_{\theta_t}(x_j)) dt$ implying that datapoints which are slow to fit will have large $\alpha$ values, further strengthening the claim that outliers are easier to reconstruct.

\textbf{6.2~~~A Reconstruction Privacy Onion}

\citet{privacy_onion} showed that removing ``vulnerable" training points reveals another set of training points which are susceptible to inference attacks. Here, we verify that our reconstruction attacks sees a similar phenomenon. Specifically, we consider reconstructing CIFAR-10 training images after training on $n$ datapoints, with $n = 900$ initially. We train and attack networks with $w \propto \sqrt{n}$, and then iterative remove the 20 most easily reconstructed datapoints based on the reconstrution curve. We scale the network capacity with $n$ so that our attack is unable to reconstruct the entire training set. We see that in \cref{fig:onion_figure}, that despite our network and attack capacity decreasing as we remove more datapoints, we are still able to reconstruct data points with increasing attack quality, replicating the ``privacy onion" effect. Future work could look at how the interaction of $\alpha$ parameters affects which items are susceptible to reconstruction post-datapoint removal. Likewise, we evaluate the test accuracy on these pruned subsets in \cref{fig:onion_figure}. We see that as, these easily reconstructed datapoints tend to be outliers, removing them has a reduced effect on the test accuracy, however as \citet{power_law_shit} discusses, the decision to remove easy vs. hard datapoints during pruning is dependent on other factors such as the size of the dataset and the complexity of the task.

\section{Unifying Reconstruction and Distillation}
In the previous sections, we considered the task of reconstructing the \textit{entire} training set. To do this, we set the reconstruction image count $M > N$. What happens if we set $M < N$? Do we reconstruct a subset of a few training images, or do we form images that are \emph{averages} of the training set?

\begin{wrapfigure}[15]{r}{0.5\textwidth}
\vspace{-7.5mm}
\begin{center}
\includegraphics[width = 0.5\textwidth]{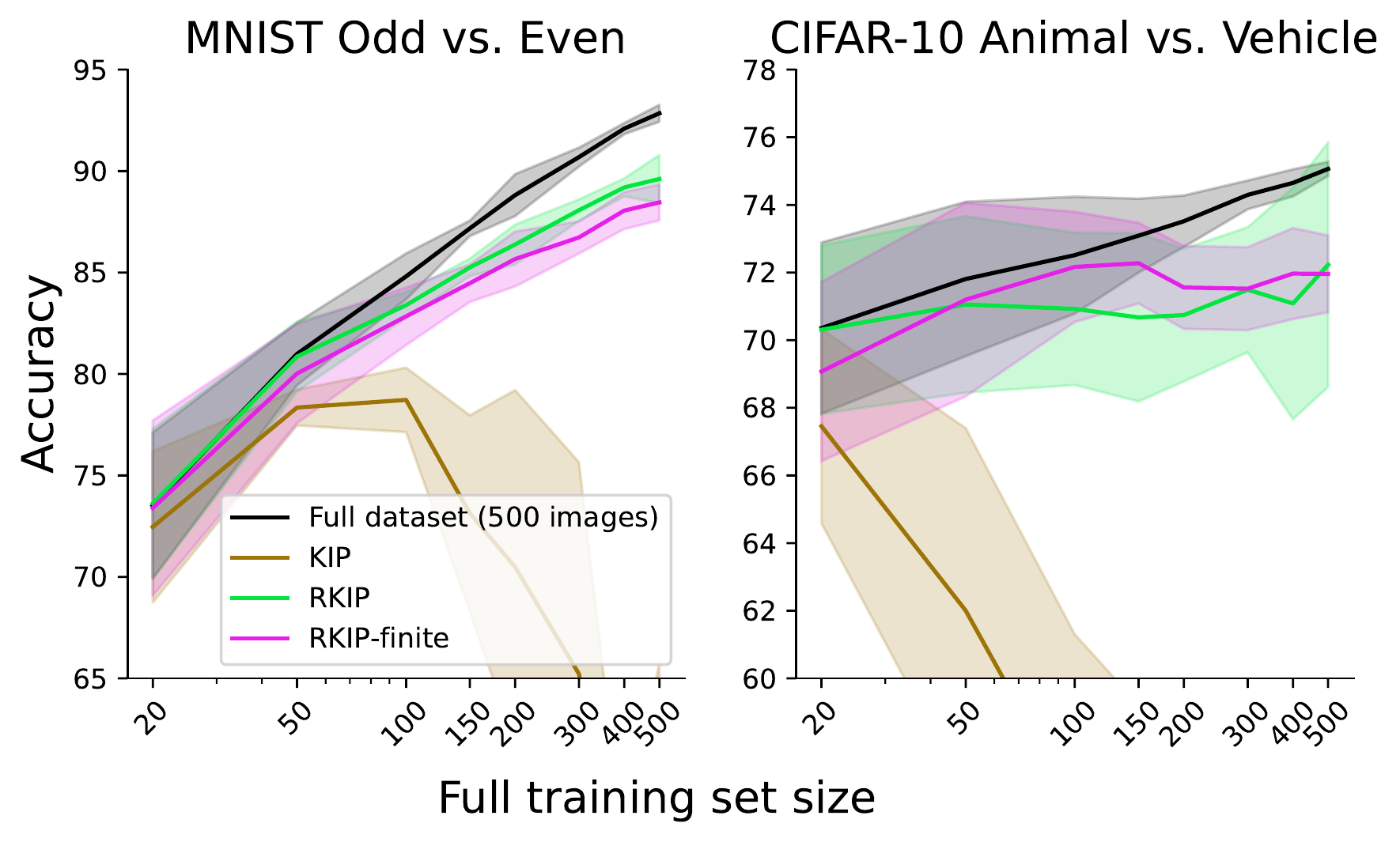}
\vskip -0.2in
\caption{Performance of KIP, RKIP, and RKIP-finite on distill N images down to 20 images, trained on a 4096 width network with standard dynamics. KIP fails to transfer to finite networks while RKIP variations succeed.}
\label{fig:distillation_accs}
\end{center}
\vskip -0.2in
\end{wrapfigure}
We perform this experiment on the CIFAR-10 Animal/Vehicle task with 500 training images for a 4096-width model with linearized dynamics, aiming to reconstruct only 20 images. We recover the images shown in \cref{app:finite_rkip_vis}. With a few exceptions, these images are not items of the training set, but rather these images look like \emph{averages} of classes.

Now, what if we retrain a network on these reconstructed images? Noting that the optimal set of dual parameters for a full reconstruction is given by $\alpha^T = K^{-1}_{TT}y_T$, a natural choice for the training labels for these images is $y_R = K_{RR} \alpha^R$, where we compute the empirical NTK for $K_{RR}$ and use the learned $\alpha^R$ parameters found during reconstruction. Retraining a different network from scratch on these 20 recovered images yields high accuracy, as shown in \cref{tab:kip_rkip_performances} as RKIP-finite. This suggests that by doing this reconstruction we performed \textit{dataset distillation}, that is we constructed a smaller set of images that accurately approximates the full dataset. This is not a coincidence, and the two algorithms are in fact the same. More formally:

\begin{theorem}
\label{thm:reconst_equals_distill}
The reconstruction scheme of Eq. \ref{eq:recon_opt_problem} with KKT points of Eq. \ref{eq:kkt_1} and Eq. \ref{eq:kkt_2}, with $M \leq N$ where $M$ is the reconstruction image counts and $N$ is the dataset size, can be written as a kernel inducing point distillation loss under a different norm plus a variance-controlled error as follows:
\begin{align*}
    \mathcal{L}_{\textrm{Reconstruction}} &= \overbrace{\|y_T - K_{TR}K_{RR}^{-1}y_R\|^2_{K^{-1}_{TT}}  + \lambda_{\text{var of } R | T}}^{\textrm{RKIP loss}}
\end{align*} \label{eq:rkip_loss}
\end{theorem}
The full proof is given in \cref{app:rkip_derivation}. $\lambda_{\text{var of } R | T}$ is proportional to the variance of the reconstruction data points conditioned on the training data, based on the NTK (see \cref{app:rkip_derivation}). Intuitively, it ensures that training images provide ``information" about the reconstructions.
Compared to the loss of a well-known dataset-distillation algorithm, KIP (with $S$ referring to the distilled dataset):
\[\mathcal{L}_{\textrm{KIP}} = \|y_T - K_{TS}K_{SS}^{-1}y_S\|^2_2\]

The connection is apparent: the reconstruction loss is equal to the KIP dataset distillation loss under a different norm, where, rather than weighting each datapoint equally, we weight training images by their inverse similarity measured by the NTK, plus $\lambda_{\text{var of } r | T}$. This leads to a variant of KIP which we call Recon-KIP (RKIP) which uses the reconstruction loss in \cref{eq:rkip_loss}. Note that for large datasets, RKIP is not practically feasible since it requires computing $K^{-1}_{TT}$, which is typical $N\times N$. We deal with small datasets in this work so it is still tractable.

\begin{figure*}[t]
\begin{center}
\includegraphics[height=1.0in]{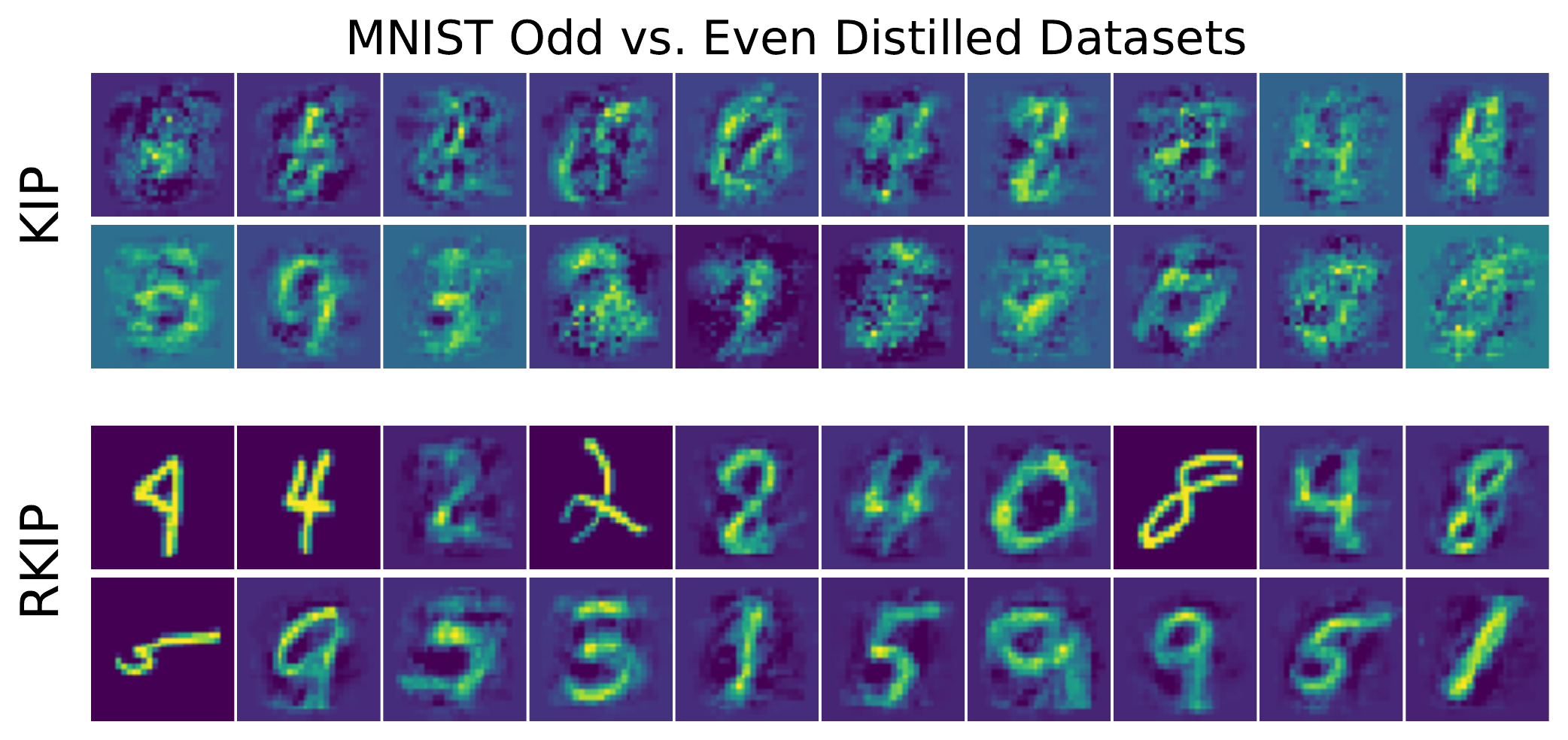}
\includegraphics[height=1.0in]{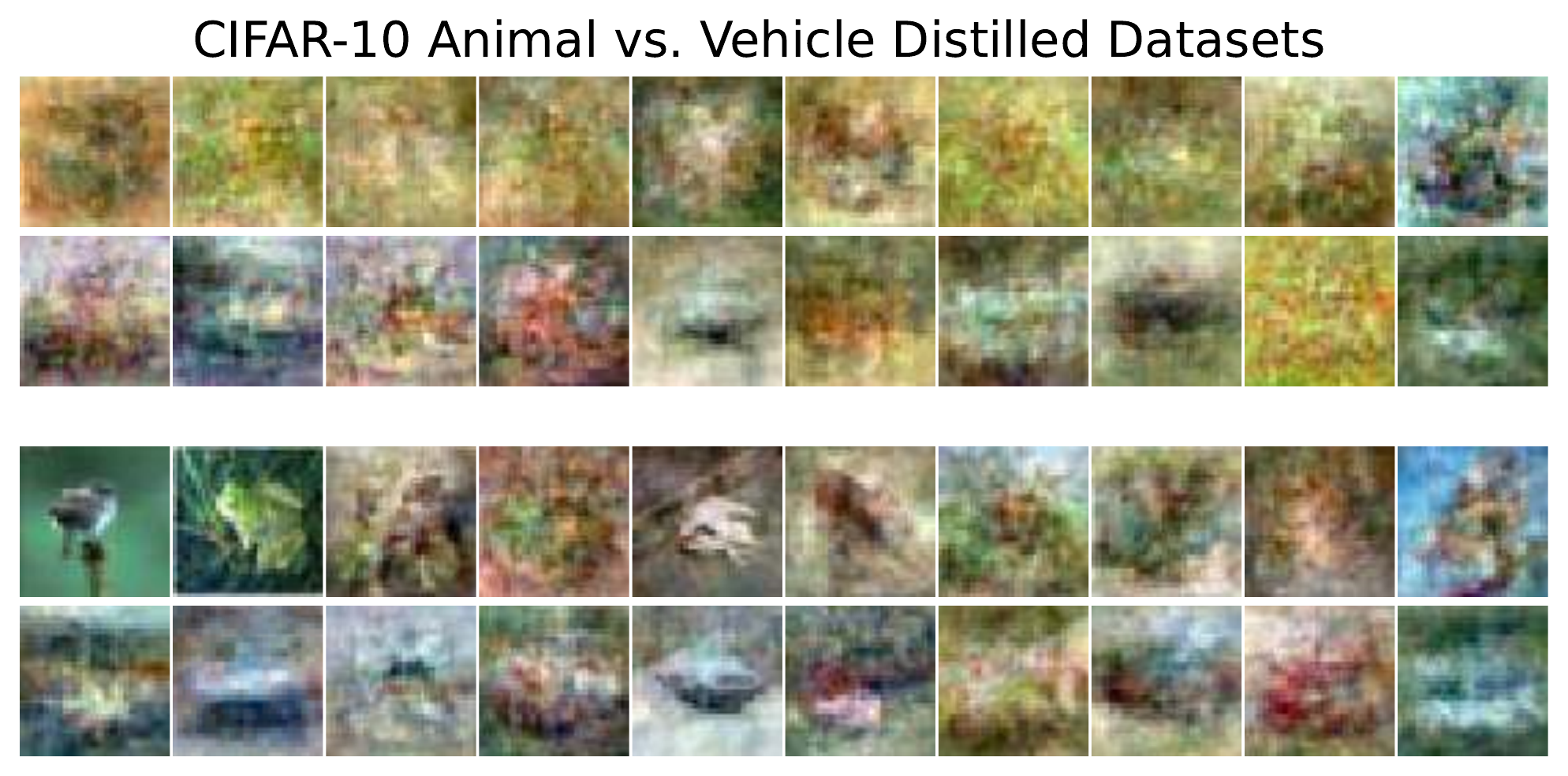}
\caption{Visualizations of distilled datasets, on MNIST Odd/Even and CIFAR-10 Animal/Vehicle classification made with KIP, and RKIP. We distill datasets of 500 original images to 20 images (shown). KIP does not copy the original training images, while RKIP occasionally reproduces training images.}
\label{fig:visualize_distilled_datasets}
\end{center}
\vskip -0.2in
\end{figure*}

We summarize the performance on KIP and RKIP in \cref{tab:kip_rkip_performances} on the MNIST Odd/Even and CIFAR-10 Animal/Vehicle task, distilling 500 images down to 20. We evaluate 4096-width networks with standard or linearized dynamics, and infinite width using the NTK. Additionally, we consider using the images/labels made from reconstructing dataset points using a finite network trained on the full dataset and call this RKIP-finite. Note in this case the labels are not necessarily $\{ +1, -1\}$, as $K_{0, RR} \alpha^R$ are not guaranteed to be one-hot labels. Similar results for distilling fewer images (20 - 500 training images) to 20 distilled images are shown in figure \cref{fig:visualize_distilled_datasets}. 

We observe in \cref{tab:kip_rkip_performances} that both KIP and RKIP have high infinite-width accuracies, but KIP sees a significant performance drop when transferring to finite networks. For example, while KIP achieves 91.53\% infinite-width test accuracy on the MNIST odd/even task, its finite-width performance is $55.62\%$, not significantly better than a random guess. Interestingly, this performance gap increases as we distill more images, as seen in \cref{fig:distillation_accs}. For small distilled datasets there is little to no performance drop but for larger ones the difference is significant. In contrast, RKIP surprisingly sees almost no performance drop in the finite-width settings. We hypothesize that this finite-width transfer performance difference for KIP and not RKIP could be due to the contribution of $\lambda_{\text{var of } r | T}$, which we discuss in \cref{app:rkip_derivation}. We leave it to future work to explore this further. 
Additionally, RKIP-finite performs nearly as well as RKIP, despite distilling using the information from a single finite-width neural network.

\text{\bfseries \large 8~~~Discussion, Limitations, and Conclusion}

In this work we showed that a stronger variant of the attack given in \citet{reconstruction} which requires wide neural networks trained under MSE loss can provably reconstruct the entire training set, owing to the injectivity of the NTK kernel measure embedding. We showed that this attack works in practice for finite-width networks, with deviations from the infinite-width regime weakening the attack. We looked at how outlier datapoints are more likely to be reconstructed under our attack, and that these easily reconstructed images can be detrimental to learning. Finally, we made a novel connection between this reconstruction attack and dataset distillation. While this sheds light on dataset reconstruction attacks, and their theoretical underpinnings, there are still many avenues to explore.

In this work, we primarily explored 2-layer fully connected networks, where neural networks are known to behave similarly to their infinite-width counterparts. Meanwhile, deeper convolutional networks are known to deviate significantly, and it is unclear how well the attacks in this paper would transfer to those settings, and what adjustments would need to be made. Secondly, while we observed increasing model width increases the network's ``resolving capacity" (i.e. how many images it could reconstruct), future work could look at how this quantity arises from deviations in the finite-width NTK from the infinite width one. Finally, we still need to resolve how the dataset reconstruction notion of privacy connects with more established notions such as differential privacy, which is the subject of future work.

We believe this work provides an important step toward understanding the strengths and weaknesses of dataset reconstruction attacks and provide novel connections to existing literature such as the Neural Tangent Kernel and dataset distillation.

\section{Reproducbility Statement}
This work uses open source datasets and models. Experimental details such as hyperparameters are described in \cref{app:experiment_details}. We additionally provide code for running the experiments in the supplementary material.

\bibliography{iclr2024_conference}

\begin{thebibliography}{61}
\providecommand{\natexlab}[1]{#1}
\providecommand{\url}[1]{\texttt{#1}}
\expandafter\ifx\csname urlstyle\endcsname\relax
  \providecommand{\doi}[1]{doi: #1}\else
  \providecommand{\doi}{doi: \begingroup \urlstyle{rm}\Url}\fi

\bibitem[Abadi et~al.(2016)Abadi, Chu, Goodfellow, McMahan, Mironov, Talwar, and Zhang]{deep_learning_with_DP_OG_paper}
Martin Abadi, Andy Chu, Ian Goodfellow, H.~Brendan McMahan, Ilya Mironov, Kunal Talwar, and Li~Zhang.
\newblock Deep learning with differential privacy.
\newblock In \emph{Proceedings of the 2016 ACM SIGSAC Conference on Computer and Communications Security}, CCS '16, pp.\  308–318, New York, NY, USA, 2016. Association for Computing Machinery.
\newblock ISBN 9781450341394.
\newblock \doi{10.1145/2976749.2978318}.
\newblock URL \url{https://doi.org/10.1145/2976749.2978318}.

\bibitem[Aitken \& Gur-Ari(2020)Aitken and Gur-Ari]{finite_dynamcis2}
Kyle Aitken and Guy Gur-Ari.
\newblock On the asymptotics of wide networks with polynomial activations.
\newblock \emph{ArXiv}, abs/2006.06687, 2020.

\bibitem[Arora et~al.(2019)Arora, Du, Hu, Li, Salakhutdinov, and Wang]{Arora_Exact_NTK_calc}
Sanjeev Arora, Simon~S Du, Wei Hu, Zhiyuan Li, Russ~R Salakhutdinov, and Ruosong Wang.
\newblock On exact computation with an infinitely wide neural net.
\newblock In \emph{Advances in Neural Information Processing Systems}, pp.\  8141--8150. Curran Associates, Inc., 2019.

\bibitem[Arpit et~al.(2017)Arpit, Jastrzebski, Ballas, Krueger, Bengio, Kanwal, Maharaj, Fischer, Courville, Bengio, et~al.]{memorization_p1}
Devansh Arpit, Stanis{\l}aw Jastrzebski, Nicolas Ballas, David Krueger, Emmanuel Bengio, Maxinder~S Kanwal, Tegan Maharaj, Asja Fischer, Aaron Courville, Yoshua Bengio, et~al.
\newblock A closer look at memorization in deep networks.
\newblock In \emph{International conference on machine learning}, pp.\  233--242. PMLR, 2017.

\bibitem[Babuschkin et~al.(2020)Babuschkin, Baumli, Bell, Bhupatiraju, Bruce, Buchlovsky, Budden, Cai, Clark, Danihelka, Fantacci, Godwin, Jones, Hemsley, Hennigan, Hessel, Hou, Kapturowski, Keck, Kemaev, King, Kunesch, Martens, Merzic, Mikulik, Norman, Quan, Papamakarios, Ring, Ruiz, Sanchez, Schneider, Sezener, Spencer, Srinivasan, Wang, Stokowiec, and Viola]{deepmind2020jax}
Igor Babuschkin, Kate Baumli, Alison Bell, Surya Bhupatiraju, Jake Bruce, Peter Buchlovsky, David Budden, Trevor Cai, Aidan Clark, Ivo Danihelka, Claudio Fantacci, Jonathan Godwin, Chris Jones, Ross Hemsley, Tom Hennigan, Matteo Hessel, Shaobo Hou, Steven Kapturowski, Thomas Keck, Iurii Kemaev, Michael King, Markus Kunesch, Lena Martens, Hamza Merzic, Vladimir Mikulik, Tamara Norman, John Quan, George Papamakarios, Roman Ring, Francisco Ruiz, Alvaro Sanchez, Rosalia Schneider, Eren Sezener, Stephen Spencer, Srivatsan Srinivasan, Luyu Wang, Wojciech Stokowiec, and Fabio Viola.
\newblock The {D}eep{M}ind {JAX} {E}cosystem, 2020.
\newblock URL \url{http://github.com/deepmind}.

\bibitem[Bagdasaryan \& Shmatikov(2019)Bagdasaryan and Shmatikov]{dp_accuracy_disparate}
Eugene Bagdasaryan and Vitaly Shmatikov.
\newblock Differential privacy has disparate impact on model accuracy.
\newblock \emph{CoRR}, abs/1905.12101, 2019.
\newblock URL \url{http://arxiv.org/abs/1905.12101}.

\bibitem[Bourtoule et~al.(2019)Bourtoule, Chandrasekaran, Choquette{-}Choo, Jia, Travers, Zhang, Lie, and Papernot]{unlearning}
Lucas Bourtoule, Varun Chandrasekaran, Christopher~A. Choquette{-}Choo, Hengrui Jia, Adelin Travers, Baiwu Zhang, David Lie, and Nicolas Papernot.
\newblock Machine unlearning.
\newblock \emph{CoRR}, abs/1912.03817, 2019.
\newblock URL \url{http://arxiv.org/abs/1912.03817}.

\bibitem[Bradbury et~al.(2018)Bradbury, Frostig, Hawkins, Johnson, Leary, Maclaurin, Necula, Paszke, Vander{P}las, Wanderman-{M}ilne, and Zhang]{jax2018github}
James Bradbury, Roy Frostig, Peter Hawkins, Matthew~James Johnson, Chris Leary, Dougal Maclaurin, George Necula, Adam Paszke, Jake Vander{P}las, Skye Wanderman-{M}ilne, and Qiao Zhang.
\newblock {JAX}: composable transformations of {P}ython+{N}um{P}y programs, 2018.
\newblock URL \url{http://github.com/google/jax}.

\bibitem[Brown et~al.(2021)Brown, Bun, Feldman, Smith, and Talwar]{memorization_necessary}
Gavin Brown, Mark Bun, Vitaly Feldman, Adam Smith, and Kunal Talwar.
\newblock When is memorization of irrelevant training data necessary for high-accuracy learning?
\newblock In \emph{Proceedings of the 53rd Annual ACM SIGACT Symposium on Theory of Computing}, pp.\  123--132, 2021.

\bibitem[Carlini et~al.(2019)Carlini, Liu, Erlingsson, Kos, and Song]{carlini_memorize_1}
Nicholas Carlini, Chang Liu, \'{U}lfar Erlingsson, Jernej Kos, and Dawn Song.
\newblock The secret sharer: Evaluating and testing unintended memorization in neural networks.
\newblock In \emph{Proceedings of the 28th USENIX Conference on Security Symposium}, SEC'19, pp.\  267–284, USA, 2019. USENIX Association.
\newblock ISBN 9781939133069.

\bibitem[Carlini et~al.(2020)Carlini, Tram{\`{e}}r, Wallace, Jagielski, Herbert{-}Voss, Lee, Roberts, Brown, Song, Erlingsson, Oprea, and Raffel]{llm_generative_attack}
Nicholas Carlini, Florian Tram{\`{e}}r, Eric Wallace, Matthew Jagielski, Ariel Herbert{-}Voss, Katherine Lee, Adam Roberts, Tom~B. Brown, Dawn Song, {\'{U}}lfar Erlingsson, Alina Oprea, and Colin Raffel.
\newblock Extracting training data from large language models.
\newblock \emph{CoRR}, abs/2012.07805, 2020.
\newblock URL \url{https://arxiv.org/abs/2012.07805}.

\bibitem[Carlini et~al.(2021)Carlini, Chien, Nasr, Song, Terzis, and Tram{\`{e}}r]{mia_first_principles}
Nicholas Carlini, Steve Chien, Milad Nasr, Shuang Song, Andreas Terzis, and Florian Tram{\`{e}}r.
\newblock Membership inference attacks from first principles.
\newblock \emph{CoRR}, abs/2112.03570, 2021.
\newblock URL \url{https://arxiv.org/abs/2112.03570}.

\bibitem[Carlini et~al.(2022)Carlini, Jagielski, Zhang, Papernot, Terzis, and Tramer]{privacy_onion}
Nicholas Carlini, Matthew Jagielski, Chiyuan Zhang, Nicolas Papernot, Andreas Terzis, and Florian Tramer.
\newblock The privacy onion effect: Memorization is relative.
\newblock In S.~Koyejo, S.~Mohamed, A.~Agarwal, D.~Belgrave, K.~Cho, and A.~Oh (eds.), \emph{Advances in Neural Information Processing Systems}, volume~35, pp.\  13263--13276. Curran Associates, Inc., 2022.
\newblock URL \url{https://proceedings.neurips.cc/paper_files/paper/2022/file/564b5f8289ba846ebc498417e834c253-Paper-Conference.pdf}.

\bibitem[{Centers for Medicare \& Medicaid Services}(1996)]{hipaa}
{Centers for Medicare \& Medicaid Services}.
\newblock {The Health Insurance Portability and Accountability Act of 1996 (HIPAA)}.
\newblock Online at http://www.cms.hhs.gov/hipaa/, 1996.

\bibitem[Chizat et~al.(2019)Chizat, Oyallon, and Bach]{lazy_training}
L{\'e}na{\"i}c Chizat, Edouard Oyallon, and Francis~R. Bach.
\newblock On lazy training in differentiable programming.
\newblock In \emph{NeurIPS}, 2019.

\bibitem[de~Azevedo (https://math.stackexchange.com/users/339790/rodrigo-de azevedo)()]{least_norm_stackexchange}
Rodrigo de~Azevedo (https://math.stackexchange.com/users/339790/rodrigo-de azevedo).
\newblock Does gradient descent converge to a minimum-norm solution in least-squares problems?
\newblock Mathematics Stack Exchange.
\newblock URL \url{https://math.stackexchange.com/q/3499305}.
\newblock URL:https://math.stackexchange.com/q/3499305 (version: 2022-02-18).

\bibitem[Dwork et~al.(2006)Dwork, McSherry, Nissim, and Smith]{Original_DP_paper}
Cynthia Dwork, Frank McSherry, Kobbi Nissim, and Adam Smith.
\newblock Calibrating noise to sensitivity in private data analysis.
\newblock In \emph{Proceedings of the Third Conference on Theory of Cryptography}, TCC'06, pp.\  265–284, Berlin, Heidelberg, 2006. Springer-Verlag.
\newblock ISBN 3540327312.
\newblock \doi{10.1007/11681878_14}.
\newblock URL \url{https://doi.org/10.1007/11681878_14}.

\bibitem[{European Commission}(2016)]{gdpr}
{European Commission}.
\newblock Regulation ({EU}) 2016/679 of the {European} {Parliament} and of the {Council} of 27 {April} 2016 on the protection of natural persons with regard to the processing of personal data and on the free movement of such data, and repealing {Directive} 95/46/{EC} ({General} {Data} {Protection} {Regulation}) ({Text} with {EEA} relevance), 2016.
\newblock URL \url{https://eur-lex.europa.eu/eli/reg/2016/679/oj}.

\bibitem[Feldman(2020)]{do_we_need_to_memorize}
Vitaly Feldman.
\newblock Does learning require memorization? a short tale about a long tail.
\newblock In \emph{Proceedings of the 52nd Annual ACM SIGACT Symposium on Theory of Computing}, STOC 2020, pp.\  954–959, New York, NY, USA, 2020. Association for Computing Machinery.
\newblock ISBN 9781450369794.
\newblock \doi{10.1145/3357713.3384290}.
\newblock URL \url{https://doi.org/10.1145/3357713.3384290}.

\bibitem[Feldman \& Zhang(2020{\natexlab{a}})Feldman and Zhang]{influence_outliers}
Vitaly Feldman and Chiyuan Zhang.
\newblock What neural networks memorize and why: Discovering the long tail via influence estimation.
\newblock \emph{CoRR}, abs/2008.03703, 2020{\natexlab{a}}.
\newblock URL \url{https://arxiv.org/abs/2008.03703}.

\bibitem[Feldman \& Zhang(2020{\natexlab{b}})Feldman and Zhang]{why_do_they_memorize}
Vitaly Feldman and Chiyuan Zhang.
\newblock What neural networks memorize and why: Discovering the long tail via influence estimation.
\newblock In \emph{Proceedings of the 34th International Conference on Neural Information Processing Systems}, NIPS'20, Red Hook, NY, USA, 2020{\natexlab{b}}. Curran Associates Inc.
\newblock ISBN 9781713829546.

\bibitem[Fort et~al.(2020)Fort, Dziugaite, Paul, Kharaghani, Roy, and Ganguli]{fortman}
Stanislav Fort, Gintare~Karolina Dziugaite, Mansheej Paul, Sepideh Kharaghani, Daniel~M. Roy, and Surya Ganguli.
\newblock Deep learning versus kernel learning: an empirical study of loss landscape geometry and the time evolution of the neural tangent kernel.
\newblock In \emph{NeurIPS}, 2020.
\newblock URL \url{https://proceedings.neurips.cc/paper/2020/hash/405075699f065e43581f27d67bb68478-Abstract.html}.

\bibitem[Fredrikson et~al.(2015)Fredrikson, Jha, and Ristenpart]{inversion_1}
Matt Fredrikson, Somesh Jha, and Thomas Ristenpart.
\newblock Model inversion attacks that exploit confidence information and basic countermeasures.
\newblock In \emph{Proceedings of the 22nd ACM SIGSAC conference on computer and communications security}, pp.\  1322--1333, 2015.

\bibitem[Gretton et~al.(2012)Gretton, Borgwardt, Rasch, Sch\"{o}lkopf, and Smola]{MMD}
Arthur Gretton, Karsten~M. Borgwardt, Malte~J. Rasch, Bernhard Sch\"{o}lkopf, and Alexander Smola.
\newblock A kernel two-sample test.
\newblock \emph{J. Mach. Learn. Res.}, 13\penalty0 (null):\penalty0 723–773, mar 2012.
\newblock ISSN 1532-4435.

\bibitem[Haim et~al.(2022)Haim, Vardi, Yehudai, michal Irani, and Shamir]{reconstruction}
Niv Haim, Gal Vardi, Gilad Yehudai, michal Irani, and Ohad Shamir.
\newblock Reconstructing training data from trained neural networks.
\newblock In Alice~H. Oh, Alekh Agarwal, Danielle Belgrave, and Kyunghyun Cho (eds.), \emph{Advances in Neural Information Processing Systems}, 2022.
\newblock URL \url{https://openreview.net/forum?id=Sxk8Bse3RKO}.

\bibitem[Hanin \& Nica(2020)Hanin and Nica]{Hanin2020Finite}
Boris Hanin and Mihai Nica.
\newblock Finite depth and width corrections to the neural tangent kernel.
\newblock In \emph{International Conference on Learning Representations}, 2020.
\newblock URL \url{https://openreview.net/forum?id=SJgndT4KwB}.

\bibitem[He et~al.(2019)He, Zhang, and Lee]{inversion_3}
Zecheng He, Tianwei Zhang, and Ruby~B. Lee.
\newblock Model inversion attacks against collaborative inference.
\newblock In \emph{Proceedings of the 35th Annual Computer Security Applications Conference}, ACSAC '19, pp.\  148–162, New York, NY, USA, 2019. Association for Computing Machinery.
\newblock ISBN 9781450376280.
\newblock \doi{10.1145/3359789.3359824}.
\newblock URL \url{https://doi.org/10.1145/3359789.3359824}.

\bibitem[Heek et~al.(2020)Heek, Levskaya, Oliver, Ritter, Rondepierre, Steiner, and van {Z}ee]{flax}
Jonathan Heek, Anselm Levskaya, Avital Oliver, Marvin Ritter, Bertrand Rondepierre, Andreas Steiner, and Marc van {Z}ee.
\newblock {F}lax: A neural network library and ecosystem for {JAX}, 2020.
\newblock URL \url{http://github.com/google/flax}.

\bibitem[Jacot et~al.(2018)Jacot, Gabriel, and Hongler]{jacotntk}
Arthur Jacot, Franck Gabriel, and Clement Hongler.
\newblock Neural tangent kernel: Convergence and generalization in neural networks.
\newblock In S.~Bengio, H.~Wallach, H.~Larochelle, K.~Grauman, N.~Cesa-Bianchi, and R.~Garnett (eds.), \emph{Advances in Neural Information Processing Systems}, volume~31. Curran Associates, Inc., 2018.
\newblock URL \url{https://proceedings.neurips.cc/paper/2018/file/5a4be1fa34e62bb8a6ec6b91d2462f5a-Paper.pdf}.

\bibitem[Ji \& Telgarsky(2020)Ji and Telgarsky]{kkt_2}
Ziwei Ji and Matus Telgarsky.
\newblock Directional convergence and alignment in deep learning.
\newblock In H.~Larochelle, M.~Ranzato, R.~Hadsell, M.F. Balcan, and H.~Lin (eds.), \emph{Advances in Neural Information Processing Systems}, volume~33, pp.\  17176--17186. Curran Associates, Inc., 2020.
\newblock URL \url{https://proceedings.neurips.cc/paper/2020/file/c76e4b2fa54f8506719a5c0dc14c2eb9-Paper.pdf}.

\bibitem[Kingma \& Ba(2015)Kingma and Ba]{adam}
Diederik~P. Kingma and Jimmy Ba.
\newblock Adam: {A} method for stochastic optimization.
\newblock In Yoshua Bengio and Yann LeCun (eds.), \emph{3rd International Conference on Learning Representations, {ICLR} 2015, San Diego, CA, USA, May 7-9, 2015, Conference Track Proceedings}, 2015.
\newblock URL \url{http://arxiv.org/abs/1412.6980}.

\bibitem[Koh \& Liang(2017)Koh and Liang]{understanding_black_bow_with_influence}
Pang~Wei Koh and Percy Liang.
\newblock Understanding black-box predictions via influence functions.
\newblock In \emph{Proceedings of the 34th International Conference on Machine Learning - Volume 70}, ICML'17, pp.\  1885–1894. JMLR.org, 2017.

\bibitem[Le \& Yang(2015)Le and Yang]{tinyimagenet}
Ya~Le and Xuan~S. Yang.
\newblock Tiny imagenet visual recognition challenge.
\newblock 2015.
\newblock URL \url{https://api.semanticscholar.org/CorpusID:16664790}.

\bibitem[Lee et~al.(2019)Lee, Xiao, Schoenholz, Bahri, Novak, Sohl-Dickstein, and Pennington]{wide_linear_models}
Jaehoon Lee, Lechao Xiao, Samuel Schoenholz, Yasaman Bahri, Roman Novak, Jascha Sohl-Dickstein, and Jeffrey Pennington.
\newblock Wide neural networks of any depth evolve as linear models under gradient descent.
\newblock \emph{Advances in neural information processing systems}, 32, 2019.

\bibitem[Loo et~al.(2022{\natexlab{a}})Loo, Hasani, Amini, and Rus]{RFAD}
Noel Loo, Ramin Hasani, Alexander Amini, and Daniela Rus.
\newblock Efficient dataset distillation using random feature approximation.
\newblock \emph{Advances in Neural Information Processing Systems}, 2022{\natexlab{a}}.

\bibitem[Loo et~al.(2022{\natexlab{b}})Loo, Hasani, Amini, and Rus]{loo2022evolution}
Noel Loo, Ramin Hasani, Alexander Amini, and Daniela Rus.
\newblock Evolution of neural tangent kernels under benign and adversarial training.
\newblock In \emph{Advances in Neural Information Processing Systems}, 2022{\natexlab{b}}.

\bibitem[Lyu \& Li(2020)Lyu and Li]{kkt_1}
Kaifeng Lyu and Jian Li.
\newblock Gradient descent maximizes the margin of homogeneous neural networks.
\newblock In \emph{International Conference on Learning Representations}, 2020.
\newblock URL \url{https://openreview.net/forum?id=SJeLIgBKPS}.

\bibitem[Malladi et~al.(2023)Malladi, Wettig, Yu, Chen, and Arora]{fine_tuning_ntk_llm}
Sadhika Malladi, Alexander Wettig, Dingli Yu, Danqi Chen, and Sanjeev Arora.
\newblock A kernel-based view of language model fine-tuning, 2023.
\newblock URL \url{https://openreview.net/forum?id=erHaiO9gz3m}.

\bibitem[Nguyen et~al.(2021{\natexlab{a}})Nguyen, Chen, and Lee]{KIP1}
Timothy Nguyen, Zhourong Chen, and Jaehoon Lee.
\newblock Dataset meta-learning from kernel ridge-regression.
\newblock In \emph{International Conference on Learning Representations}, 2021{\natexlab{a}}.
\newblock URL \url{https://openreview.net/forum?id=l-PrrQrK0QR}.

\bibitem[Nguyen et~al.(2021{\natexlab{b}})Nguyen, Novak, Xiao, and Lee]{KIP2}
Timothy Nguyen, Roman Novak, Lechao Xiao, and Jaehoon Lee.
\newblock Dataset distillation with infinitely wide convolutional networks.
\newblock In \emph{Thirty-Fifth Conference on Neural Information Processing Systems}, 2021{\natexlab{b}}.
\newblock URL \url{https://openreview.net/forum?id=hXWPpJedrVP}.

\bibitem[Novak et~al.(2020)Novak, Xiao, Hron, Lee, Alemi, Sohl-Dickstein, and Schoenholz]{neural_tangents}
Roman Novak, Lechao Xiao, Jiri Hron, Jaehoon Lee, Alexander~A. Alemi, Jascha Sohl-Dickstein, and Samuel~S. Schoenholz.
\newblock Neural tangents: Fast and easy infinite neural networks in python.
\newblock In \emph{International Conference on Learning Representations}, 2020.
\newblock URL \url{https://openreview.net/forum?id=SklD9yrFPS}.

\bibitem[Novak et~al.(2022)Novak, Sohl{-}Dickstein, and Schoenholz]{fast_finite_NTK}
Roman Novak, Jascha Sohl{-}Dickstein, and Samuel~S. Schoenholz.
\newblock Fast finite width neural tangent kernel.
\newblock In Kamalika Chaudhuri, Stefanie Jegelka, Le~Song, Csaba Szepesv{\'{a}}ri, Gang Niu, and Sivan Sabato (eds.), \emph{International Conference on Machine Learning, {ICML} 2022, 17-23 July 2022, Baltimore, Maryland, {USA}}, volume 162 of \emph{Proceedings of Machine Learning Research}, pp.\  17018--17044. {PMLR}, 2022.
\newblock URL \url{https://proceedings.mlr.press/v162/novak22a.html}.

\bibitem[Rigaki \& Garcia(2020)Rigaki and Garcia]{privacy_attacks_survey}
Maria Rigaki and Sebastian Garcia.
\newblock A survey of privacy attacks in machine learning.
\newblock \emph{CoRR}, abs/2007.07646, 2020.
\newblock URL \url{https://arxiv.org/abs/2007.07646}.

\bibitem[Shokri et~al.(2016)Shokri, Stronati, and Shmatikov]{MIA}
Reza Shokri, Marco Stronati, and Vitaly Shmatikov.
\newblock Membership inference attacks against machine learning models.
\newblock \emph{CoRR}, abs/1610.05820, 2016.
\newblock URL \url{http://arxiv.org/abs/1610.05820}.

\bibitem[Shon et~al.(2022)Shon, Lee, Kim, and Kim]{continual_learning_linear}
Hyounguk Shon, Janghyeon Lee, Seung~Hwan Kim, and Junmo Kim.
\newblock {DLCFT:} deep linear continual fine-tuning for general incremental learning.
\newblock \emph{CoRR}, abs/2208.08112, 2022.
\newblock \doi{10.48550/arXiv.2208.08112}.
\newblock URL \url{https://doi.org/10.48550/arXiv.2208.08112}.

\bibitem[Sorscher et~al.(2022)Sorscher, Geirhos, Shekhar, Ganguli, and Morcos]{power_law_shit}
Ben Sorscher, Robert Geirhos, Shashank Shekhar, Surya Ganguli, and Ari~S. Morcos.
\newblock Beyond neural scaling laws: beating power law scaling via data pruning.
\newblock \emph{ArXiv}, abs/2206.14486, 2022.
\newblock URL \url{https://api.semanticscholar.org/CorpusID:250113273}.

\bibitem[Sriperumbudur et~al.(2011)Sriperumbudur, Fukumizu, and Lanckriet]{universality_of_measures}
Bharath~K. Sriperumbudur, Kenji Fukumizu, and Gert~R.G. Lanckriet.
\newblock Universality, characteristic kernels and rkhs embedding of measures.
\newblock \emph{Journal of Machine Learning Research}, 12\penalty0 (70):\penalty0 2389--2410, 2011.
\newblock URL \url{http://jmlr.org/papers/v12/sriperumbudur11a.html}.

\bibitem[Tsilivis \& Kempe(2022)Tsilivis and Kempe]{nyu_peeps_paper}
Nikolaos Tsilivis and Julia Kempe.
\newblock What can the neural tangent kernel tell us about adversarial robustness?
\newblock In Alice~H. Oh, Alekh Agarwal, Danielle Belgrave, and Kyunghyun Cho (eds.), \emph{Advances in Neural Information Processing Systems}, 2022.
\newblock URL \url{https://openreview.net/forum?id=KBUgVv8z7OA}.

\bibitem[Wang et~al.(2018)Wang, Zhu, Torralba, and Efros]{wang2018dataset}
Tongzhou Wang, Jun-Yan Zhu, Antonio Torralba, and Alexei~A Efros.
\newblock Dataset distillation.
\newblock \emph{arXiv preprint arXiv:1811.10959}, 2018.

\bibitem[Welinder et~al.(2010)Welinder, Branson, Mita, Wah, Schroff, Belongie, and Perona]{cub_200}
P.~Welinder, S.~Branson, T.~Mita, C.~Wah, F.~Schroff, S.~Belongie, and P.~Perona.
\newblock {Caltech-UCSD Birds 200}.
\newblock Technical Report CNS-TR-2010-001, California Institute of Technology, 2010.

\bibitem[Wright(2022)]{flaxmodels}
Matthias Wright.
\newblock Flax models.
\newblock \url{https://github.com/matthias-wright/flaxmodels}, 2022.

\bibitem[Yang et~al.(2019)Yang, Zhang, Chang, and Liang]{inversion_2}
Ziqi Yang, Jiyi Zhang, Ee-Chien Chang, and Zhenkai Liang.
\newblock Neural network inversion in adversarial setting via background knowledge alignment.
\newblock In \emph{Proceedings of the 2019 ACM SIGSAC Conference on Computer and Communications Security}, CCS '19, pp.\  225–240, New York, NY, USA, 2019. Association for Computing Machinery.
\newblock ISBN 9781450367479.
\newblock \doi{10.1145/3319535.3354261}.
\newblock URL \url{https://doi.org/10.1145/3319535.3354261}.

\bibitem[Zancato et~al.(2020)Zancato, Achille, Ravichandran, Bhotika, and Soatto]{fine_tuning_linear}
Luca Zancato, Alessandro Achille, Avinash Ravichandran, Rahul Bhotika, and Stefano Soatto.
\newblock Predicting training time without training.
\newblock In H.~Larochelle, M.~Ranzato, R.~Hadsell, M.F. Balcan, and H.~Lin (eds.), \emph{Advances in Neural Information Processing Systems}, volume~33, pp.\  6136--6146. Curran Associates, Inc., 2020.
\newblock URL \url{https://proceedings.neurips.cc/paper_files/paper/2020/file/440e7c3eb9bbcd4c33c3535354a51605-Paper.pdf}.

\bibitem[Zandieh et~al.(2021)Zandieh, Han, Avron, Shoham, Kim, and Shin]{NTK_features_via_sketching}
Amir Zandieh, Insu Han, Haim Avron, Neta Shoham, Chaewon Kim, and Jinwoo Shin.
\newblock Scaling neural tangent kernels via sketching and random features.
\newblock In A.~Beygelzimer, Y.~Dauphin, P.~Liang, and J.~Wortman Vaughan (eds.), \emph{Advances in Neural Information Processing Systems}, 2021.
\newblock URL \url{https://openreview.net/forum?id=vIRFiA658rh}.

\bibitem[Zhang et~al.(2017)Zhang, Bengio, Hardt, Recht, and Vinyals]{rethinking_generalization}
Chiyuan Zhang, Samy Bengio, Moritz Hardt, Benjamin Recht, and Oriol Vinyals.
\newblock Understanding deep learning requires rethinking generalization.
\newblock In \emph{International Conference on Learning Representations}, 2017.
\newblock URL \url{https://openreview.net/forum?id=Sy8gdB9xx}.

\bibitem[Zhang et~al.(2021)Zhang, Bengio, Hardt, Recht, and Vinyals]{still_rethinking}
Chiyuan Zhang, Samy Bengio, Moritz Hardt, Benjamin Recht, and Oriol Vinyals.
\newblock Understanding deep learning (still) requires rethinking generalization.
\newblock \emph{Commun. ACM}, 64\penalty0 (3):\penalty0 107–115, feb 2021.
\newblock ISSN 0001-0782.
\newblock \doi{10.1145/3446776}.
\newblock URL \url{https://doi.org/10.1145/3446776}.

\bibitem[Zhao \& Bilen(2021)Zhao and Bilen]{zhao2021dsa}
Bo~Zhao and Hakan Bilen.
\newblock Dataset condensation with differentiable siamese augmentation.
\newblock \emph{arXiv preprint arXiv:2102.08259}, 2021.

\bibitem[Zhao et~al.(2021)Zhao, Mopuri, and Bilen]{zhao2021DC}
Bo~Zhao, Konda~Reddy Mopuri, and Hakan Bilen.
\newblock Dataset condensation with gradient matching.
\newblock In \emph{International Conference on Learning Representations}, 2021.
\newblock URL \url{https://openreview.net/forum?id=mSAKhLYLSsl}.

\bibitem[Zhou et~al.(2022)Zhou, Nezhadarya, and Ba]{frepo}
Yongchao Zhou, Ehsan Nezhadarya, and Jimmy Ba.
\newblock Dataset distillation using neural feature regression.
\newblock In \emph{Proceedings of the Advances in Neural Information Processing Systems (NeurIPS)}, 2022.

\bibitem[Zhou et~al.(2021)Zhou, Wang, Xian, Chen, and Xu]{meta_learning_ntk}
Yufan Zhou, Zhenyi Wang, Jiayi Xian, Changyou Chen, and Jinhui Xu.
\newblock Meta-learning with neural tangent kernels.
\newblock \emph{CoRR}, abs/2102.03909, 2021.
\newblock URL \url{https://arxiv.org/abs/2102.03909}.

\bibitem[Zhu et~al.(2019)Zhu, Liu, , and Han]{gradient_leakage}
Ligeng Zhu, Zhijian Liu, , and Song Han.
\newblock Deep leakage from gradients.
\newblock In \emph{Annual Conference on Neural Information Processing Systems (NeurIPS)}, 2019.

\end{thebibliography}
\bibliographystyle{iclr2024_conference}

\newpage
\appendix
\section*{Appendix}
\section{Comparison to \citet{reconstruction}}
\label{app:comparison_to_haim}

In \cref{sec:attack_description}, we mentioned that we had problems reproducing \citet{reconstruction}'s attack. Here, we compare the two attacks and discuss the issues we found with theirs.

Firstly, we compare the quality of the two attacks. \citet{reconstruction} open sourced their code, as well as gave the best two sets of reconstructions for both CIFAR-10 and MNIST-10. We plot the reconstruction curves of these reconstructions here:

\begin{figure*}[h]
\vskip 0.1in
\begin{center}
\includegraphics[height = 2.0in]{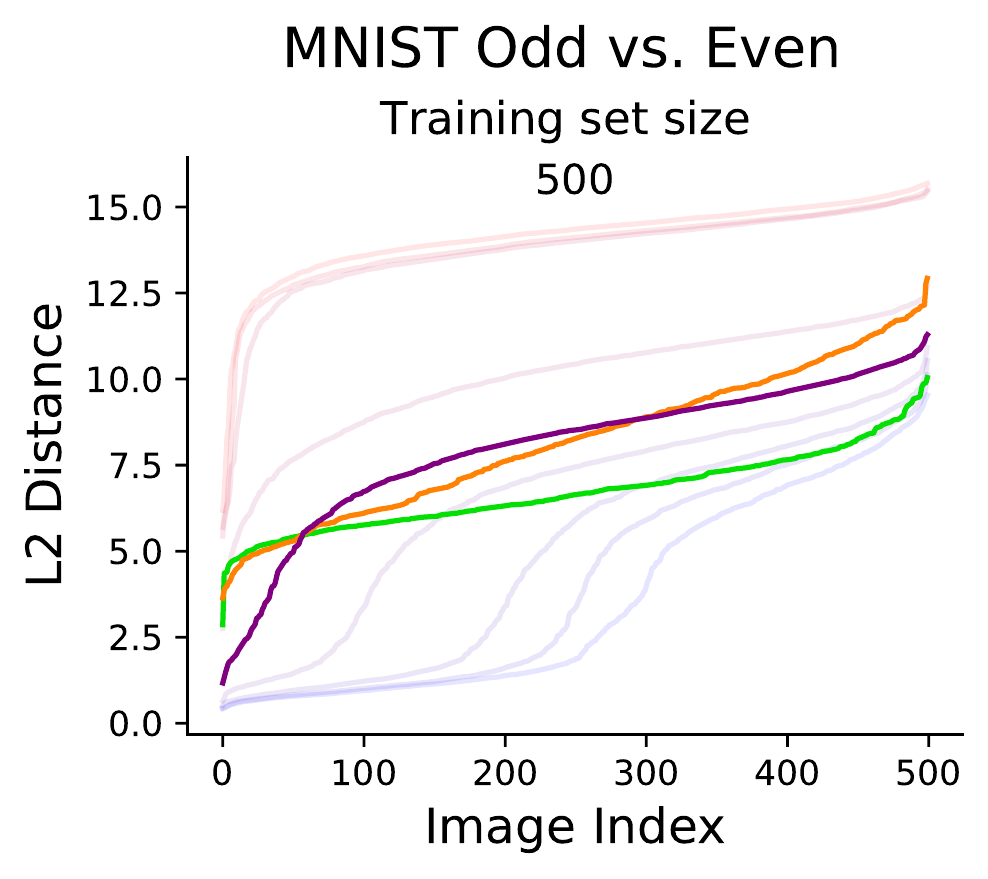}
\includegraphics[height = 2.0in]{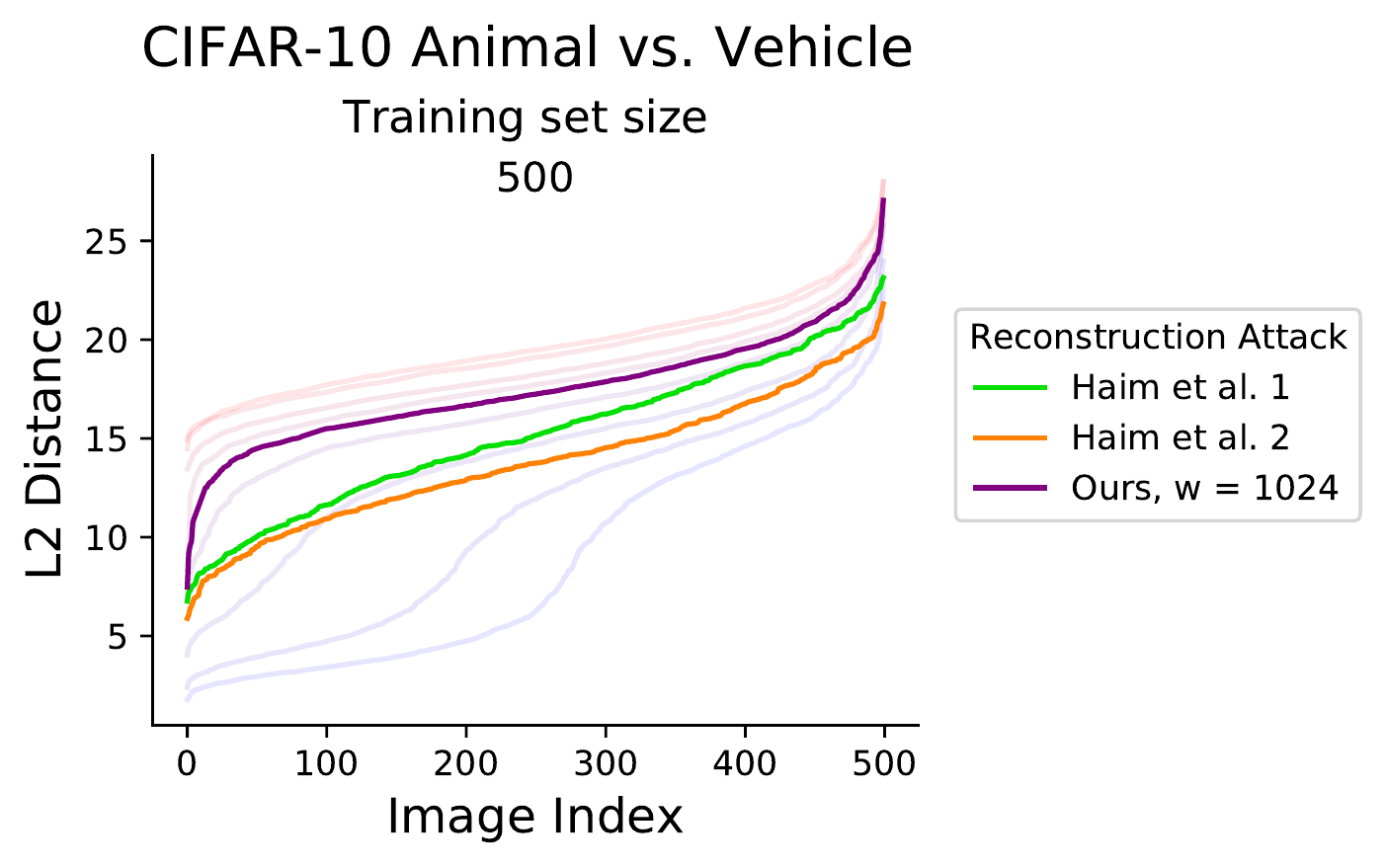}
\vskip -0.1in
\caption{Reconstruction curves for the attacks given in \citet{reconstruction}, in comparison to our reconstruction attacks, with a comparable width of 1024.}
\label{fig:reconstruction_kernel_dist_scatter_1}
\end{center}
\vspace{-6mm}
\end{figure*}

In their paper, they consider networks of a width 1000, so for a fair comparison, we highlight our reconstruction with a comparable width of 1024, under standard dynamics. We see that for MNIST, our attack has significantly better quality until image index 50 in which case both attacks perform poorly. For CIFAR-10, our attack performs worse but can achieve better performance with wider width or linearization.

Note that the two reconstruction curves presented from \citet{reconstruction} correspond to the two \textit{best} reconstructions with carefully chosen hyperparameters. These hyperparameters are chosen to maximize reconstruction quality, which requires access to the training data to measure the quality. A priori, an attacker would not be able to assess reconstruction quality, and thus would not be able to do such hyperparameter tuning. In contrast, attack parameters such as learning rate/initialization were not fine-tuned for ours, and we use the same hyperparameters for every attack. Further gains could likely be seen with more careful tuning. We would like to emphasize that the goal of this work is not necessarily to create the strongest attack, but more so to explore the properties of the attack, and conditions for failure/success.

A further limitation of their attack is that it requires homogenous neural networks (i.e. no biases), in comparison to ours which uses biases, which is closer to practice. The largest limitation of their attack is that they require \textbf{a careful initialization scheme}, in which the weights of the first layer are initialized with significantly smaller variance. In \citet{reconstruction} they discuss that this is \textbf{essential} to the success of their attack. In contrast, we use the default initialization given in the Flax neural network library \citep{flax}.

We also observed that the KKT condition given in \cref{eq:haim_max-margin}, which is required for their attack to work often is not reached in practice. To reiterate the definition of directional convergence, it requires that $\lim_{t\to\infty} \frac{\theta}{\|\theta\|_2} \to \frac{\theta'}{\|\theta'\|_2}$ for trained network parameters $\theta$, and $\theta'$ the solution to \cref{eq:haim_max-margin}. It is clear that if $\frac{\|\theta - \theta_0\|^2_2}{\|\theta_0\|^2_2} < 1$, i.e. the parameters have not drifted far from their initialization, then, we cannot hope the KKT point is reached (of course, with unlikely exceptions such as $\theta_0$ already being close to the KKT point). In practice, we found that $\|\theta_0\|_2^2 \approx 333$ and $\|\theta\|_2^2 \approx 377$, when running their attack, which suggests that the initialization still contributes a significant amount to the final parameters, suggesting that the KKT was not reached. Of course, our attack is not without limitations as well, the most notable being that we require network initialization. We leave it to future work to alleviate this requirement.

\section{Reconstruction attack algorithm details}
\label{app:alg_details}

\begin{algorithm}
\caption{Standard Reconstruction Attack}\label{alg:recon_attack}
\begin{algorithmic}
\Require Initial Parameters $\theta_0$, final parameters $\theta_f$, network function $f_\theta$, randomly initialized reconstruction images and dual parameters $\{X_R, \alpha_R\}$, optimizer $\texttt{Optim(\textrm{params}, \textrm{gradients}}$, number of steps $T$
\State $\Delta\theta = \theta_f - \theta_0$
\State $t \gets 1$
\While{$t < T$}
    \State $G \gets \sum_i \alpha_i \nabla_\theta f_{\theta_f}(x_i)$ for $\alpha_i, x_i \in \alpha_R, X_R$ \Comment{Compute reconstruction gradient}
    \State $\mathcal{L}_{\textrm{recon}} = ||\Delta \theta - G ||^2_2$ \Comment{Compute reconstruction loss}
    \State ${\alpha_R, X_R} \gets \texttt{Optim}\left( \{ \alpha_R, X_R \}, \frac{\partial L_{recon}}{\partial \{ \alpha_R, X_R \}} \right)$ \Comment{Update Reconstruction Images}
    \State $t \gets t+1$
\EndWhile
\end{algorithmic}
\end{algorithm}

\begin{algorithm}
\caption{Batched Reconstruction Attack}\label{alg:recon_attack_batched}
\begin{algorithmic}
\Require Initial Parameters $\theta_0$, final parameters $\theta_f$, network function $f_\theta$, randomly initialized reconstruction images and dual parameters $\{X_R, \alpha_R\}$, optimizer $\texttt{Optim(\textrm{params}, \textrm{gradients}}$, number of steps $T$, batch size $|B|$
\State $\Delta\theta = \theta_f - \theta_0$
\State $G_R = \sum_i \alpha_i \nabla_\theta f_{\theta_f}(x_i)$ for $\alpha_i, x_i \in \alpha_R, X_R$ \Comment{Compute total reconstruction gradient (this step can also be batched)}
\While{$t < T$}
    \State Sample batch $\alpha_B, X_B \subset \alpha_R, X_R$ of batch size $|B|$ uniformly
    \State $G_B \gets \sum_i \alpha_i \nabla_\theta f_{\theta_f}(x_i)$ for $\alpha_i, x_i \in \alpha_B, X_B$ \Comment{Compute reconstruction gradient for batch}
    \State $G_{B, old} \gets \texttt{detach}(G_B) $ \Comment{Store old batch gradient}
    \State $\mathcal{L}_{\textrm{recon}}\gets ||\Delta \theta - (G_R - G_{B, old}  + G_B) ||^2_2 $ \Comment{Compute reconstruction loss}
    \State ${\alpha_B, X_B} \leftarrow \texttt{Optim}(\{ \alpha_B, X_B \}, \partial L_{recon} /\partial \{\alpha_B, X_B \}) $ \Comment{Optimize batch images}
    \State $G_{B, new} \gets \sum_i \alpha_i \nabla_\theta f_{\theta_f}(x_i)$ for $\alpha_i, x_i \in \alpha_B, X_B$ \Comment {Compute new batch gradient}
    \State $G_R \gets G_R - G_{B, old} + \texttt{detach}(G_{B, new})$ \Comment{Update total reconstruction gradient}
    \State $t \gets t+1$
\EndWhile
\end{algorithmic}
\end{algorithm}

Here we discuss the runtime of our attack, given a model with $P$ parameters, $M$ reconstruction images, and $T$ iterations. We present two versions of the attack: the standard version of the attack, given in \cref{alg:recon_attack}, and a minibatched version of the attack \cref{alg:recon_attack_batched}. Both versions of the attack are mathetmatically equivalent, but the batched version allows for larger reconstruction sets that may not fit into memory all at once.

For the standard version of the attack presented in \cref{alg:recon_attack}, $O(MPT)$ time is required and $O(MP)$ memory is required, as naively one needs to store (and backpropagate through) gradients for each of the reconstructed images. Evidently for large datasets, one cannot pass the whole dataset through the model at once and backpropagate, and as we need $M > N$ to reconstruct the full dataset, this seems problematic. For datasets in our paper, this was not a concern, but for larger datasets it would be.

To mitigate this issue, we present a minibatched version of the attack in \cref{alg:recon_attack_batched}, which requires $O(BPT)$ time and $O(BP)$ memory.  The premise of this version that you store the a buffered value of the total sum of gradients $G_R = \sum_{i}^M g_i$ over reconstruction examples, and by carefully using autodiff, you can update only a subset of the gradients (using the buffered value of $G_R$ and subtracting the batch gradients). In \cref{alg:recon_attack_batched}, the $\texttt{detach}$ function refers to the autodiff graph detachment function common in most autodiff libraries. This method has been implemented and performs exactly the same as the original attack, however the results in this paper do not require it as we dealt with small $N$.

\section{Label conditions for full recovery}
\label{app:label_conditions}

As discussed in \cref{sec:attack_description}, we require that $\alpha_i \neq 0$ in order to recover the training image. Here we discuss the conditions for this to occur. We know that $\alpha = K^{-1}y$, and without loss of generality, consider reconstructing the final image at index $N$. Our equation for $\alpha$ becomes, focusing on the value of $\alpha_N$:

\newcommand{\matA}{K_{:N-1, :N-1}}
\newcommand{\matB}{K_{:N-1, N}}
\newcommand{\matD}{K_{N, N}}
{
\tiny

\begin{align*}
    \begin{bmatrix}
    \alpha_{:N-1} \\
    \alpha_N
    \end{bmatrix}
     &= 
     \begin{bmatrix}
    \matA & \matB\\
    \matB^\intercal & \matD
    \end{bmatrix} ^{-1}
    \begin{bmatrix}
    y_{:N-1} \\
    y_N
    \end{bmatrix} \\
    \begin{bmatrix}
    \alpha_{:N-1} \\
    \alpha_N
    \end{bmatrix}
    &= 
    \begin{bmatrix}
    \matA^{-1} + \matA^{-1}\matB Q \matB^\intercal \matA^{-1} & - \matA^{-1}\matB Q \\
    - Q\matB^T \matA^{-1} & Q \\
    \end{bmatrix}
    \begin{bmatrix}
    y_{:N-1} \\
    y_N
    \end{bmatrix}
\end{align*}
}

With $Q = (\matD - \matB \matA ^{-1} \matB^\intercal)^{-1}$. Setting $\alpha_N = 0$:

\begin{align*}
    0 &= - Q\matB^T \matA^{-1}y_{:N-1} + Qy_N \\
    y_N &= \matB^T \matA^{-1}y_{:N-1}
\end{align*}

Noting $\matB^T \matA^{-1}y_{:N-1}$ corresponds to the kernel regression prediction of $y_N$ given $y_{:N-1}$, we see that $\alpha_N = 0$ iff $x_N$ is already perfectly predicted by the remaining data points. Of course, the label corresponding exactly to the prediction occurs with probability 0.

\subsection{Alternative Interpretation of $\alpha$}
\label{app:alpha_as_loss}

In \cref{sec:which_get_attacked}, we discussed how $\alpha$ parameters can be treated as items which are ``hard" vs ``easy" to fit. Here we derive how $\alpha_j = \int_0^\infty (y_j - f_{\theta_t}(x_j)) dt$ for MSE loss (up to some scaling parameter).
\begin{align*}
    \mathcal{L}(\theta_t) &= \frac{1}{2} \sum_i (y_i - f_{\theta_t}(x_i))^2 \\
    \frac{\partial \theta_t}{\partial t} &= -\eta \frac{\partial \mathcal{L}(\theta_t)}{\partial \theta_t} \\
    \frac{\partial \theta_t}{\partial t} &= -\eta\sum_i (y_i - f_{\theta_t}(x_i)) \frac{\partial f_{\theta_t} (x_i)}{\partial \theta_t} \\
    \Delta \theta &= \int_0^\infty \frac{\partial \theta_t}{\partial t} dt \\
     &= \int_0^\infty -\eta\sum_i (y_i - f_{\theta_t}(x_i)) \frac{\partial f_{\theta_t} (x_i)}{\partial \theta_t} dt \\
     &= -\eta\sum_i \left[   \int_0^\infty (y_i - f_{\theta_t}(x_i)) \frac{\partial f_{\theta_t} (x_i)}{\partial \theta_t} dt \right]\\
    &\approx -\eta\sum_i \left[   \int_0^\infty (y_i - f_{\theta_t}(x_i)) dt \frac{\partial f_{\theta_0} (x_i)}{\partial \theta_0} \right] \quad \textrm{(Frozen Kernel approximation)}\\ 
\end{align*}

\begin{figure}[h]
\begin{center}
\includegraphics[width = 0.9\linewidth]{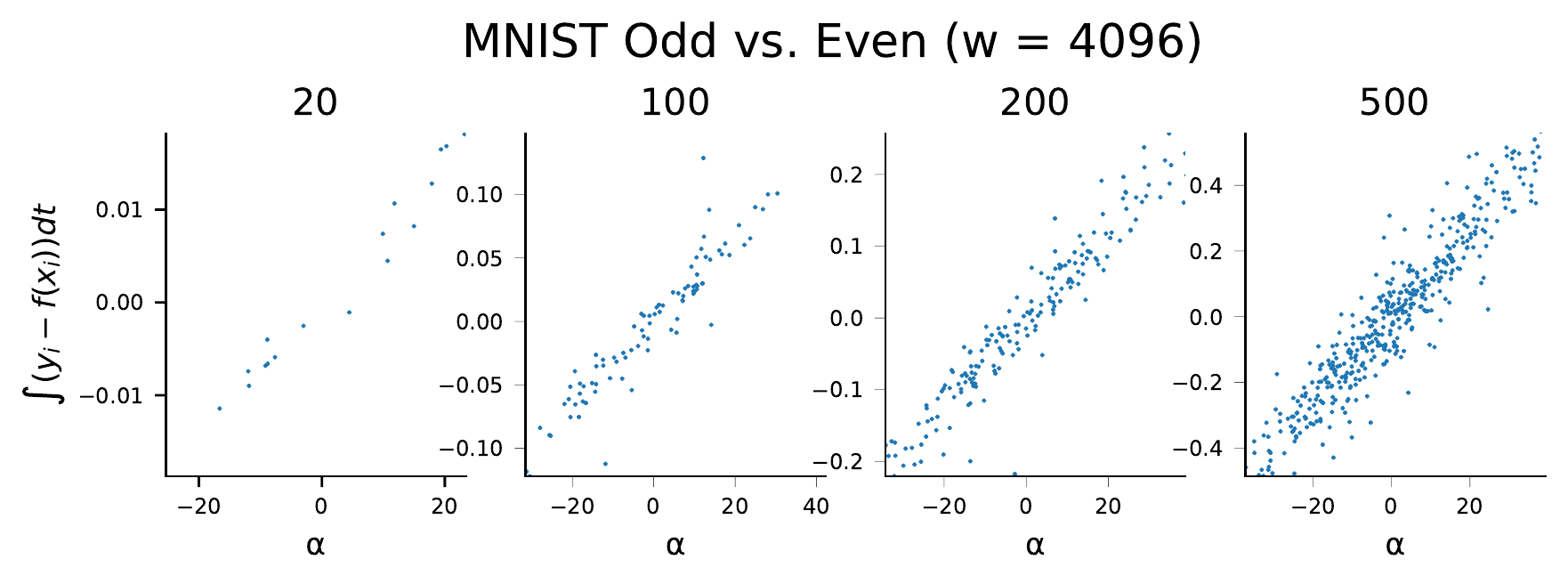}
\includegraphics[width = 0.9\linewidth]{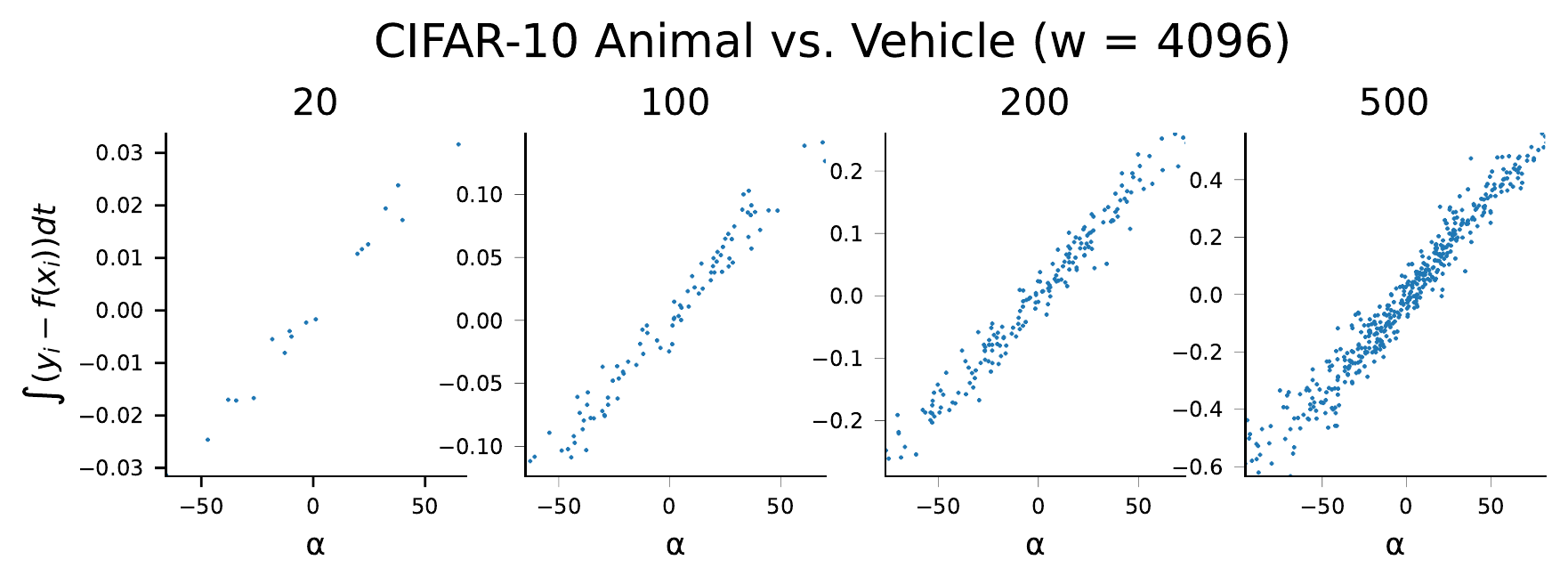}
\vskip -0.1in
\caption{Infinite width values of $\alpha$ vs. error integral formulation of error calculated from 4096-width finite networks. There is a strong correlation between the two values.}
\label{app:fig:app_alpha_theory_to_integral}
\end{center}
\vspace{-5mm}
\end{figure}

Noting that $\Delta \theta = \sum_{i} \alpha_i \frac{\partial f_{\theta_0} (x_i)}{\partial \theta_0}$ (\cref{eq:kkt_1}) and matching terms, also noting that $P > T$ (more parameters than training points), we have the system is uniquely solved when $\alpha_i = -\eta \int_0^\infty (y_j - f_{\theta_t}(x_j)) dt$. We can verify this experimentally, by plotting the calculated values of $\alpha_i$ vs. $\int_0^\infty (y_j - f_{\theta_t}(x_j)) dt$ in \cref{app:fig:app_alpha_theory_to_integral}.

\section{Proof of Theorem 2: RKIP Derivation from Reconstruction loss}
\label{app:rkip_derivation}

\begin{proof}
Here we derive how one gets from the reconstruction loss given in \cref{eq:recon_loss} to \cref{eq:rkip_loss}, the RKIP loss. We repeat the losses here:

\begin{align*}
    \mathcal{L}_{\textrm{Reconstruction}} &= \| \Delta \theta - \alpha^\intercal \nabla_\theta f_{\theta_0}(X_T) \|^2_2 \\
    \mathcal{L}_{\textrm{RKIP}} &= \|y_T - K_{TR}K_{RR}^{-1}y_R\|^2_{K^{-1}_{TT}}
\end{align*}

First we note that $\alpha^T = K_{TT}^{-1}y_T$, and if we trained on reconstructions with labels $y_R$, then $\alpha^R = K_{RR}^{-1}y_R$. For brevity we denote $s_i^{*} = \{\alpha_i^*, x_i^*\}$ and $S^* = \{\alpha^*, X_*\}$. Let $S^T$ and $S^R$ denote the training and reconstruction set, respectively
\begin{align*}
    &\quad \Big\|\Delta \theta - \sum_{\mathclap{s_j^R \in S^R}} \alpha_j^R \nabla_{\theta_f} f_{\theta_f} (x_j^R)\Big\|^2_2 \\
    &=\Big\|\sum_{\mathclap{s_i^T \in S^T}}{\alpha_i^T \nabla_{\theta_0}f(x_i^T)} - \sum_{\mathclap{s_j^R \in S_R}} \alpha_j^R \nabla_{\theta_f} f_{\theta_f} (x_j^R)\Big\|^2_2 \\
    &= \Big\|\sum_{\mathclap{s_i^T \in S^T}}{\alpha_i^T k_{\theta_0}(x_i^T, \cdot)} - \sum_{\mathclap{s_j^R \in S_R}} \alpha_j^R k_{\theta_f}(x_j^R, \cdot)\Big\|^2_2
\end{align*}

Again take the infinite width limit so $k_{\theta_0}, k_{\theta_f} \to k_{NTK}$, we which just write $k$ for simplicity.

\newcommand{\sos}[2]{\sum_{s_{#1}^{#2} \in S^{#2}}}
{\tiny
\begin{align*}
    \Big\|\sum_{\mathclap{s_i^T \in S^T}}{\alpha_i^T k_{\theta_0}(x_i^T, \cdot)} - \sum_{\mathclap{s_j^R \in S_R}} \alpha_j^R k_{\theta_f}(x_j^R, \cdot)\Big\|^2_2
    &\to \Big\|\sum_{\mathclap{s_i^T \in S^T}}{\alpha_i^T k(x_i^T, \cdot)} - \sum_{\mathclap{s_j^R \in S_R}} \alpha_j^R k(x_j^R, \cdot)\Big\|^2_2 \\
    &= \sos{i}{T}\sos{j}{T} \alpha_i^T \alpha_j^T k(x_i^T, x_j^T) - 2 \sos{i}{T}\sos{j}{R} \alpha_i^T \alpha_j^R k(x_i^T, x_j^R) \\
    &\quad + \sos{i}{R}\sos{j}{R} \alpha_i^R \alpha_j^R k(x_i^R, x_j^R) \\
    &= \alpha^{T \intercal} K_{TT} \alpha^T - 2 \alpha^{T \intercal} K_{TR} \alpha^R + \alpha^{R \intercal} K_{RR} \alpha^R \\
    &= y_T^\intercal K_{TT}^{-1} K_{TT} K_{TT}^{-1} y_T - 2 y_T^\intercal K_{TT}^{-1} K_{TR} K_{RR}^{-1} y_R + y_R^\intercal K_{RR}^{-1} K_{RR} K_{RR}^{-1} y_R \\
    &= y_T^\intercal K_{TT}^{-1} y_T - 2 y_T^\intercal K_{TT}^{-1} K_{TR} K_{RR}^{-1} y_R + y_R^\intercal K_{RR}^{-1} y_R \\
    &= y_T^\intercal K_{TT}^{-1} y_T - 2 y_T^\intercal K_{TT}^{-1} K_{TR} K_{RR}^{-1} y_R + y_R^\intercal K_{RR}^{-1} K_{RT} K_{TT}^{-1} K_{TR} K_{RR}^{-1}y_R \\
    &\quad - y_R^\intercal K_{RR}^{-1} K_{RT} K_{TT}^{-1} K_{TR} K_{RR}^{-1} y_R + y_R^\intercal K_{RR}^{-1} y_R \\
    &= \|y_T - K_{TR}K_{RR}^{-1}y_R\|^2_{K^{-1}_{TT}} + y_R^\intercal (K_{RR}^{-1} -  K_{RR}^{-1} K_{RT} K_{TT}^{-1} K_{TR} K_{RR}^{-1} ) y_R \\
    &= \underbrace{\|y_T - K_{TR}K_{RR}^{-1}y_R\|^2_{K^{-1}_{TT}}}_{\textrm{label matching}} + \underbrace{y_R^\intercal K_{RR}^{-1} (K_{RR} -  K_{RT} K_{TT}^{-1} K_{TR}) K_{RR}^{-1}y_R}_{\lambda_{\text{var of } R | T}}\\
\end{align*}
}
\end{proof}

$\lambda_{\text{var of } R|T}$ is proportional to $K_{RR} -  K_{RT} K_{TT}^{-1} K_{TR}$, which is $K_{[T, R], [T, R]} / K_{TT}$, the Schur complement of $K_{[T, R], [T, R]}$ with $K_{TT}$. Note that this is the Gaussian conditional variance formula, assuming we are making predictions of $R$ based on $T$. This regularizer ensures that not only do the reconstructed images result in the correct predictions (ensured by the label matching term) but also that our distilled dataset images do not deviate significantly from the training distribution, as measured by the NTK. We hypothesize this term is what contributes to the success of RKIP over KIP in the finite-width setting. As there is nothing that directly ensures that KIP distilled datapoints remain ``similar" to training images (only that they are predictive), these distilled images may be more susceptible to domain shift, such as moving from the infinite-width setting to finite width. This interesting behavior could be the subject of future work.

\newpage
\section{Finite RKIP Images}
\label{app:finite_rkip_vis}
\begin{figure}[h]
\begin{center}
\includegraphics[width = 0.8\linewidth]{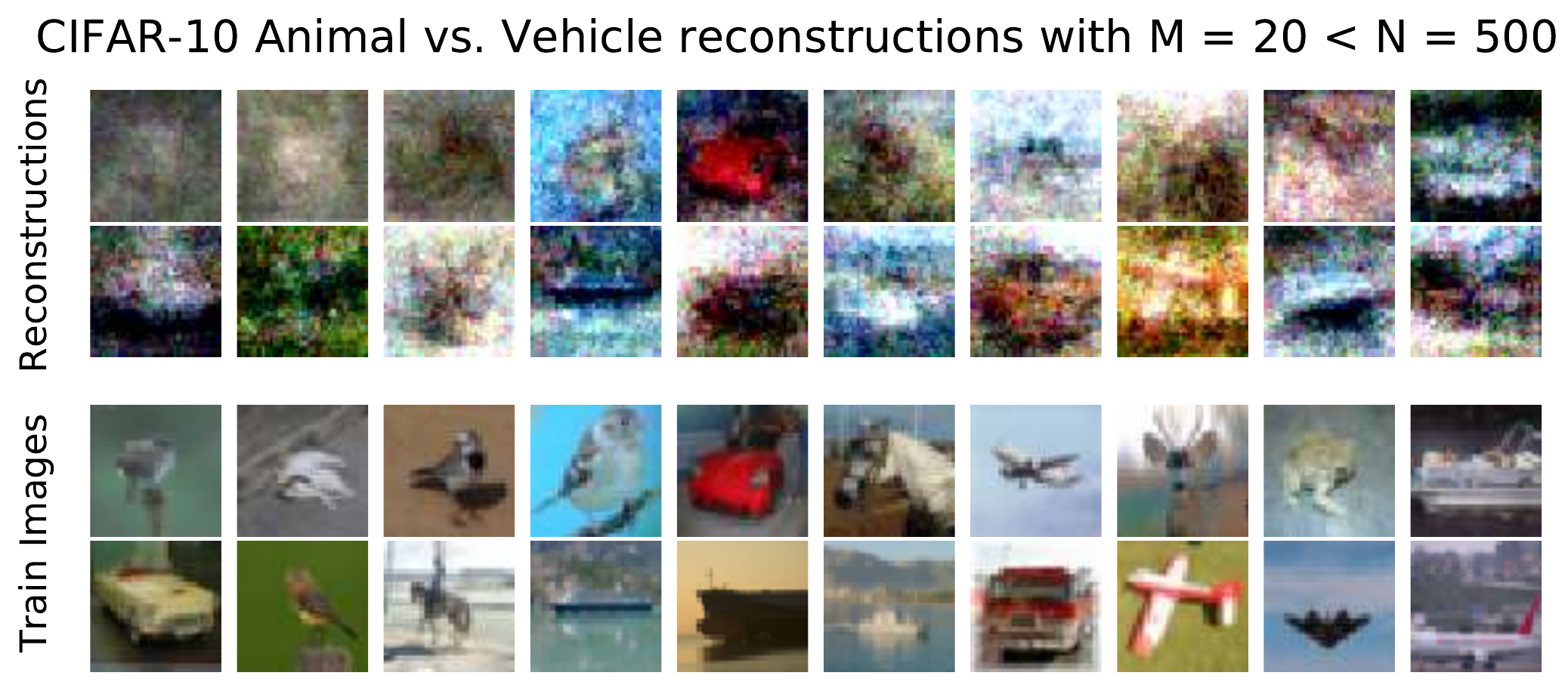}
\caption{Reconstructing 20 images from a network trained on 500 CIFAR-10 images. Reconstructions often do not match actual training images and contain heavy corruption. Retraining on these images yields high accuracy.}
\label{fig:rkip_finite_visualization}
\end{center}
\end{figure}

\begin{figure*}[h]
\vskip 0.1in
\begin{center}
\includegraphics[width = 0.45\linewidth]{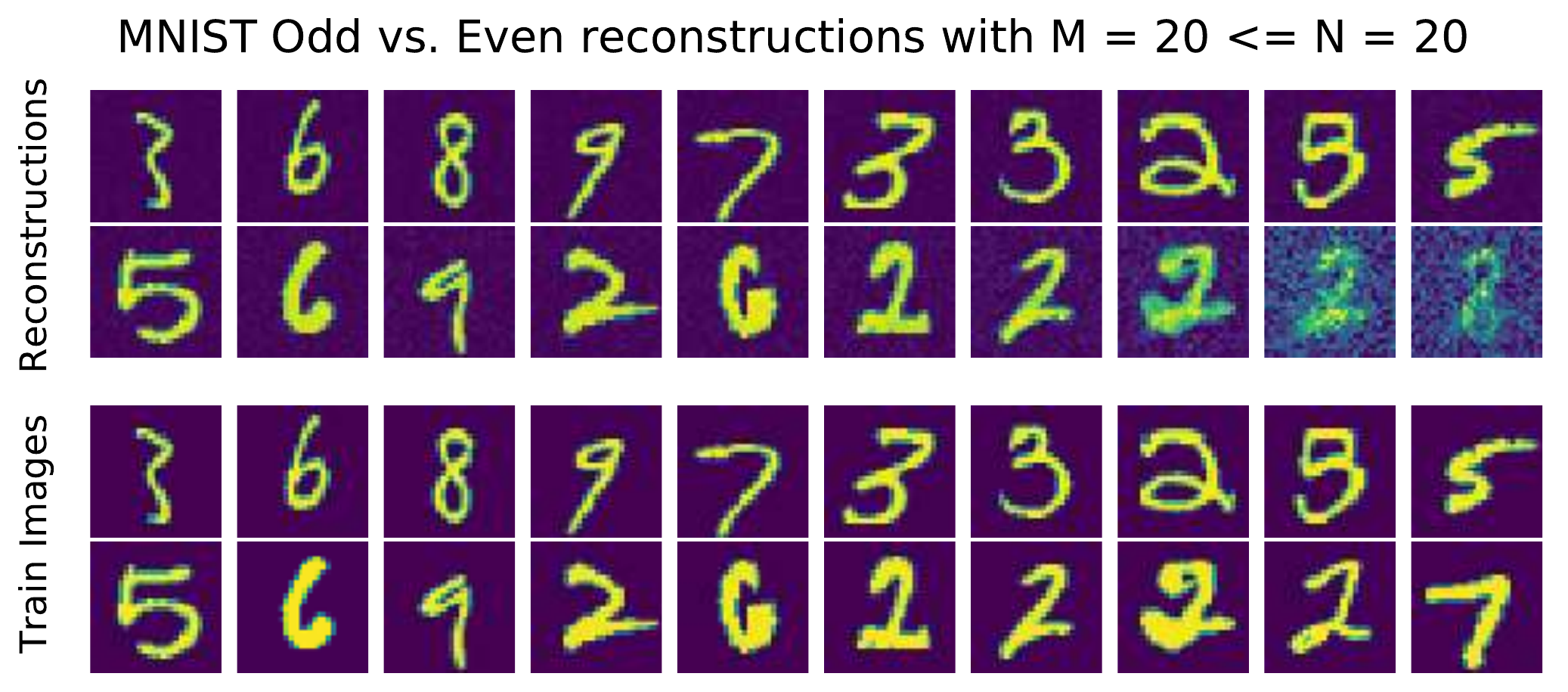}
\includegraphics[width = 0.45\linewidth]{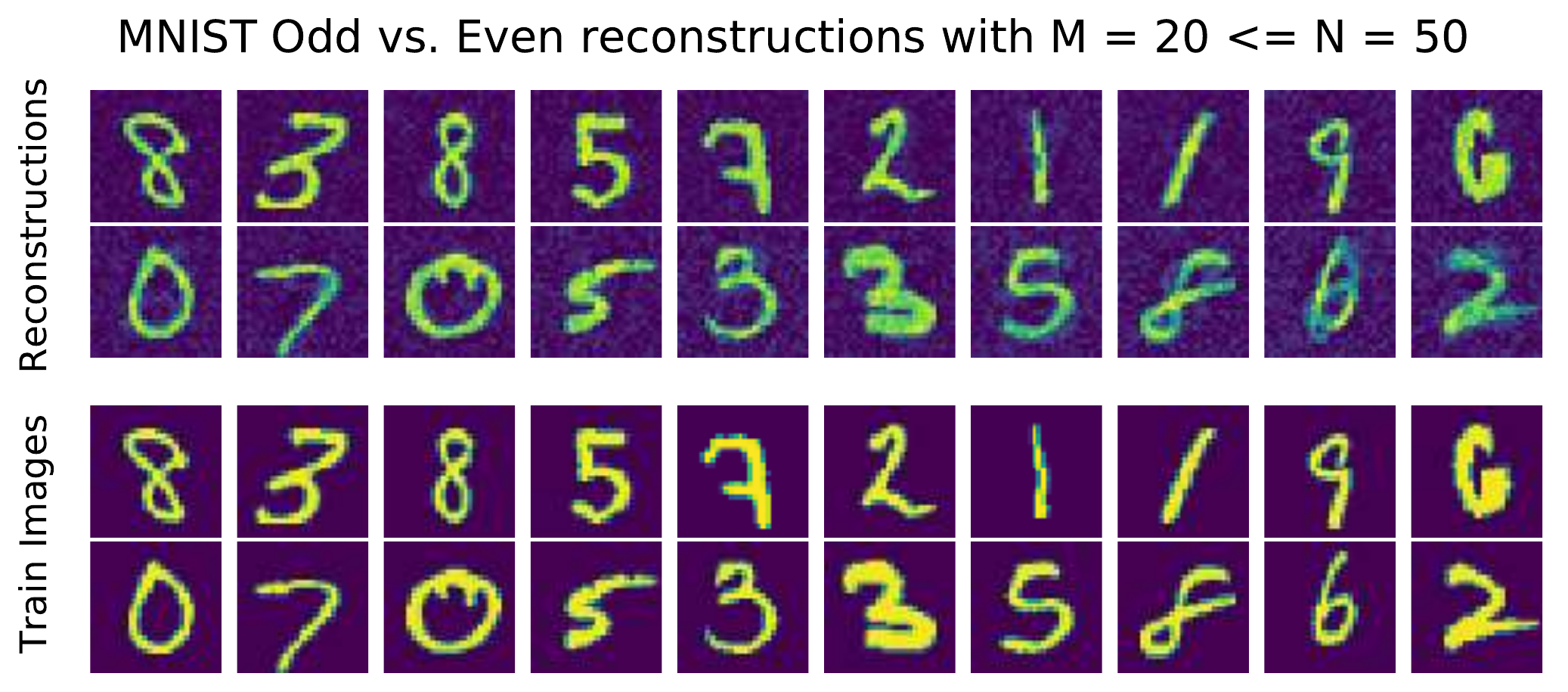}
\includegraphics[width = 0.45\linewidth]{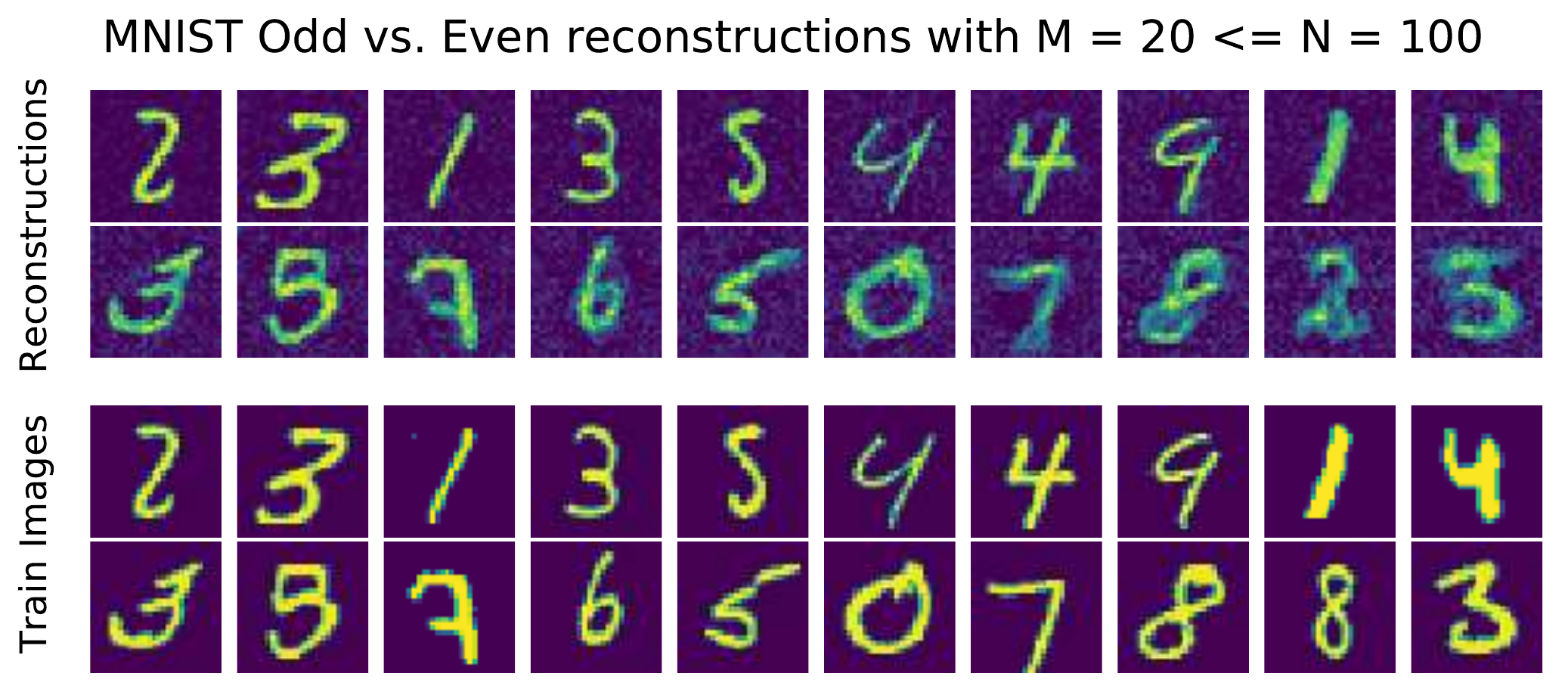}
\includegraphics[width = 0.45\linewidth]{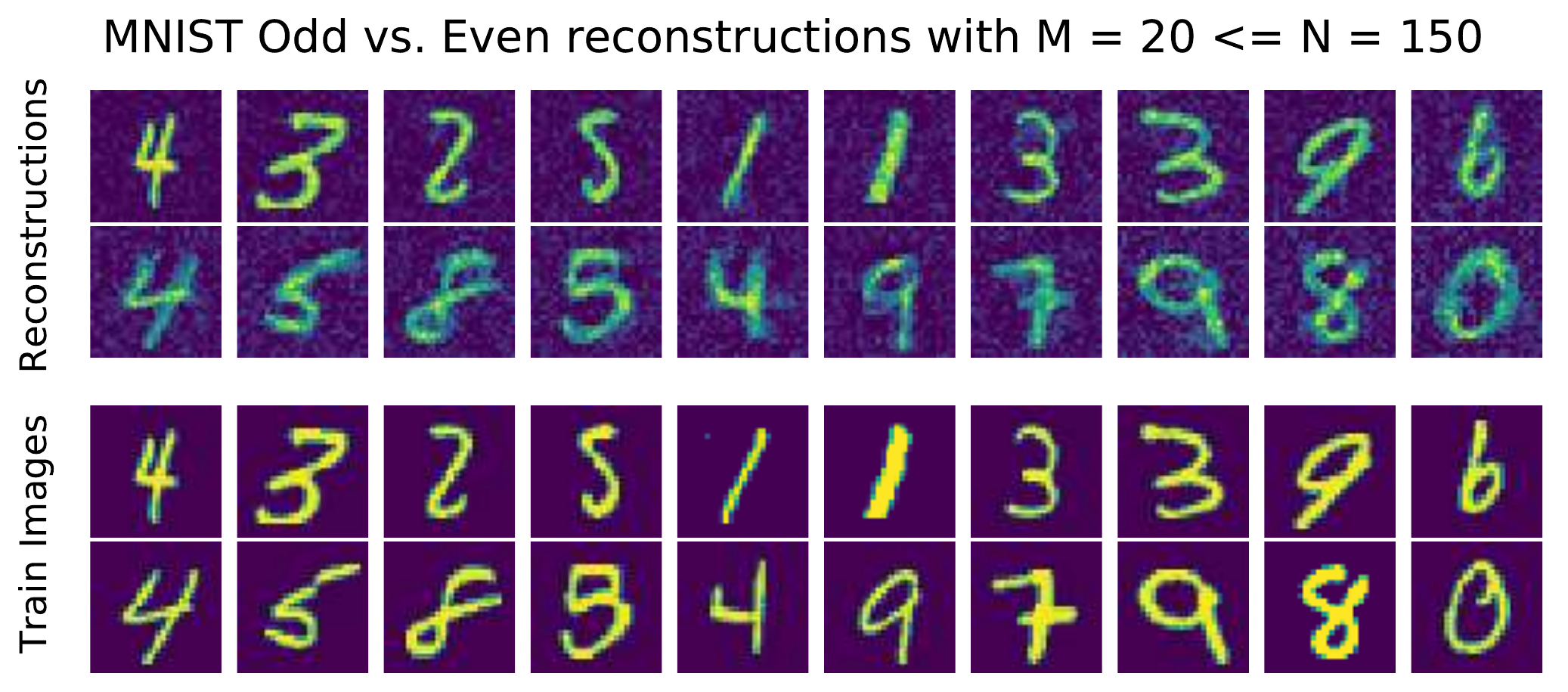}
\includegraphics[width = 0.45\linewidth]{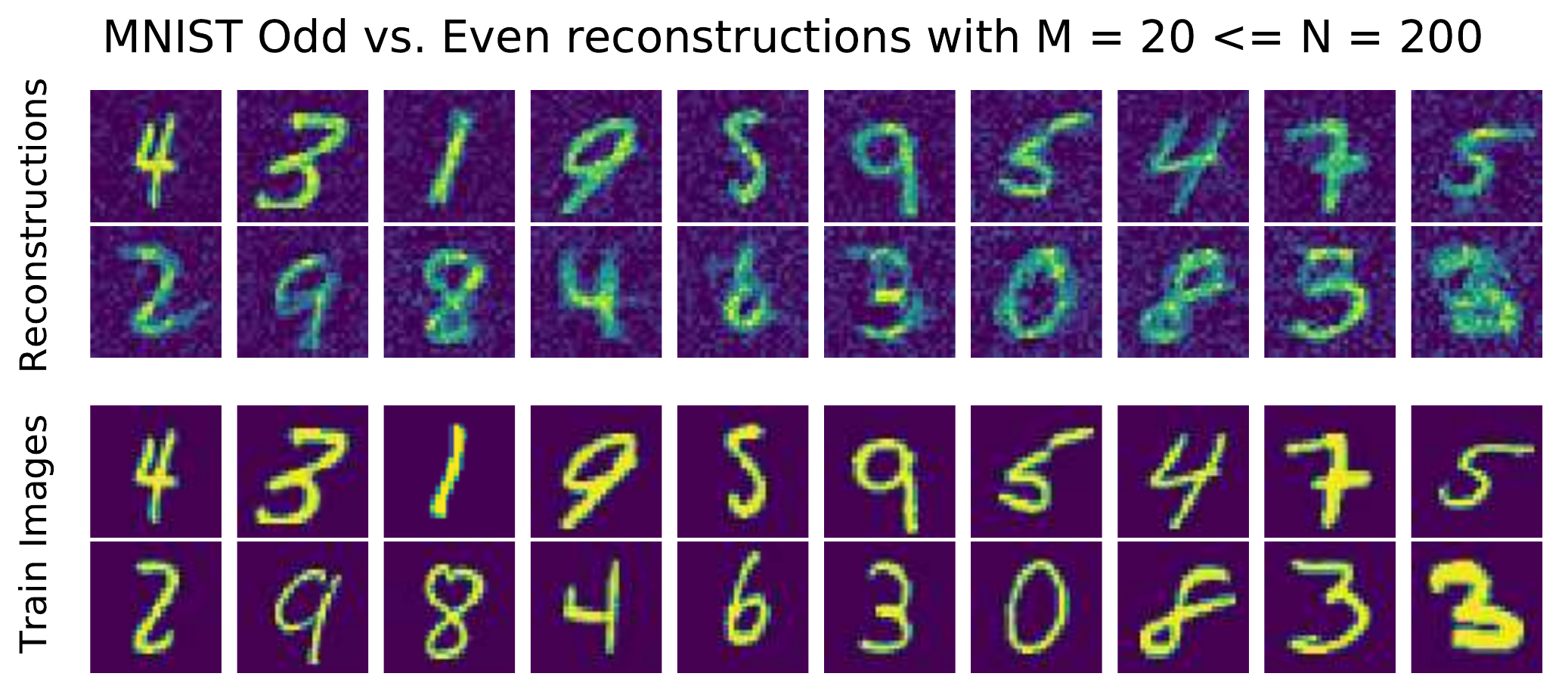}
\includegraphics[width = 0.45\linewidth]{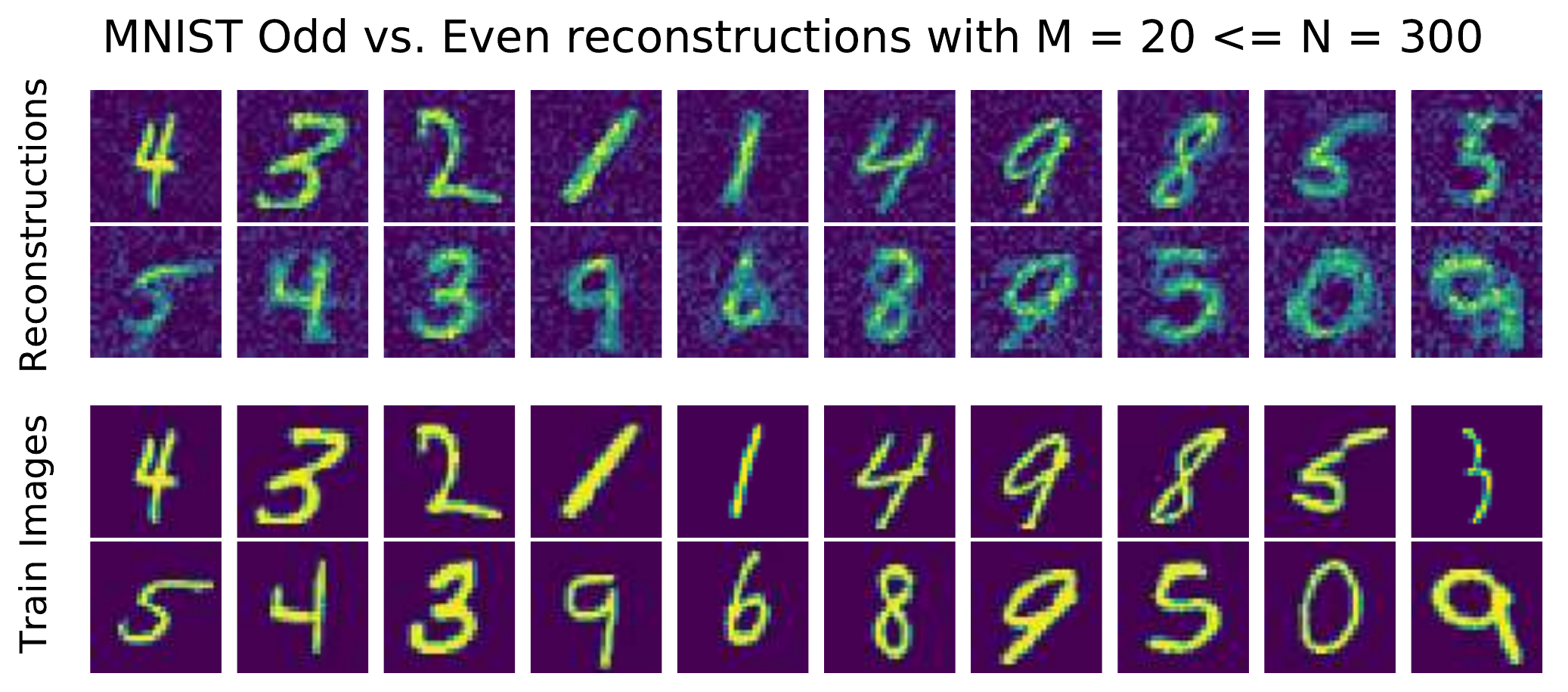}
\includegraphics[width = 0.45\linewidth]{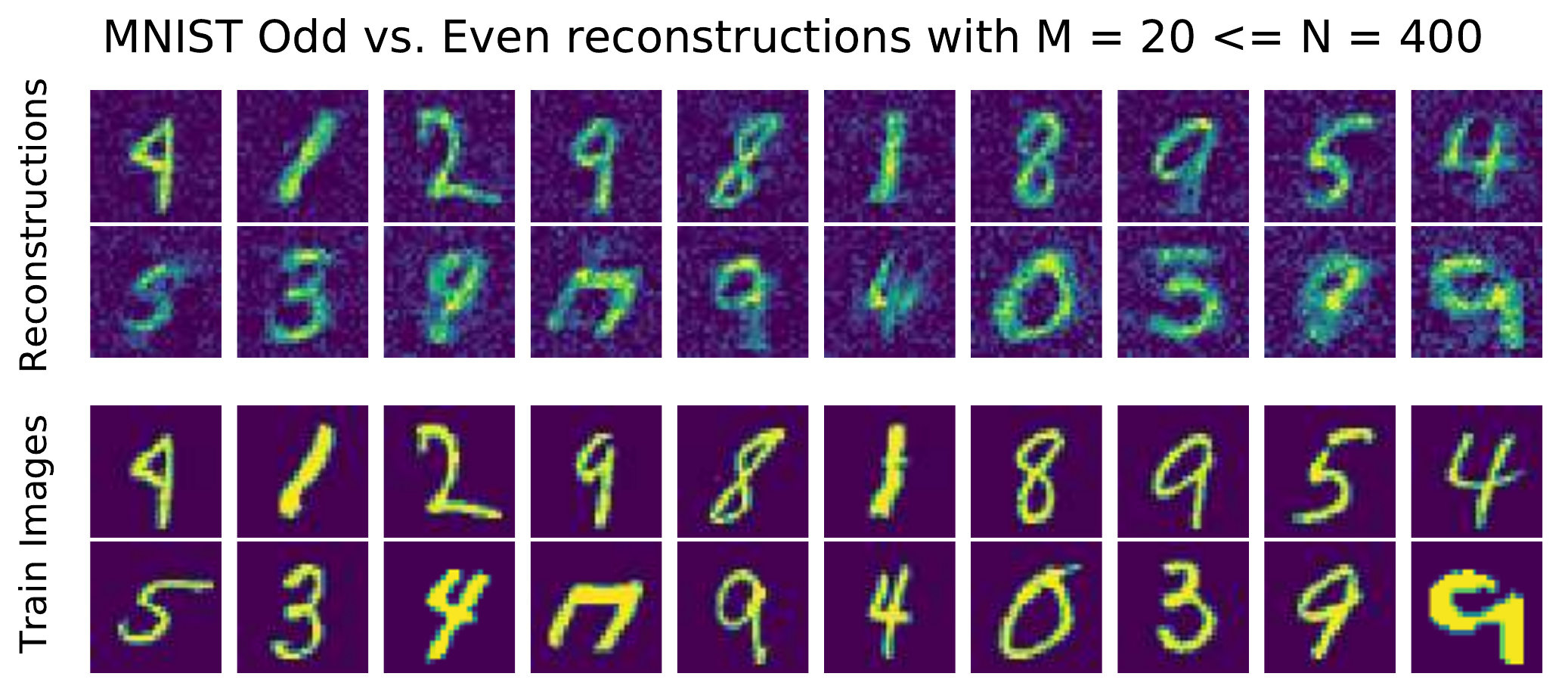}
\includegraphics[width = 0.45\linewidth]{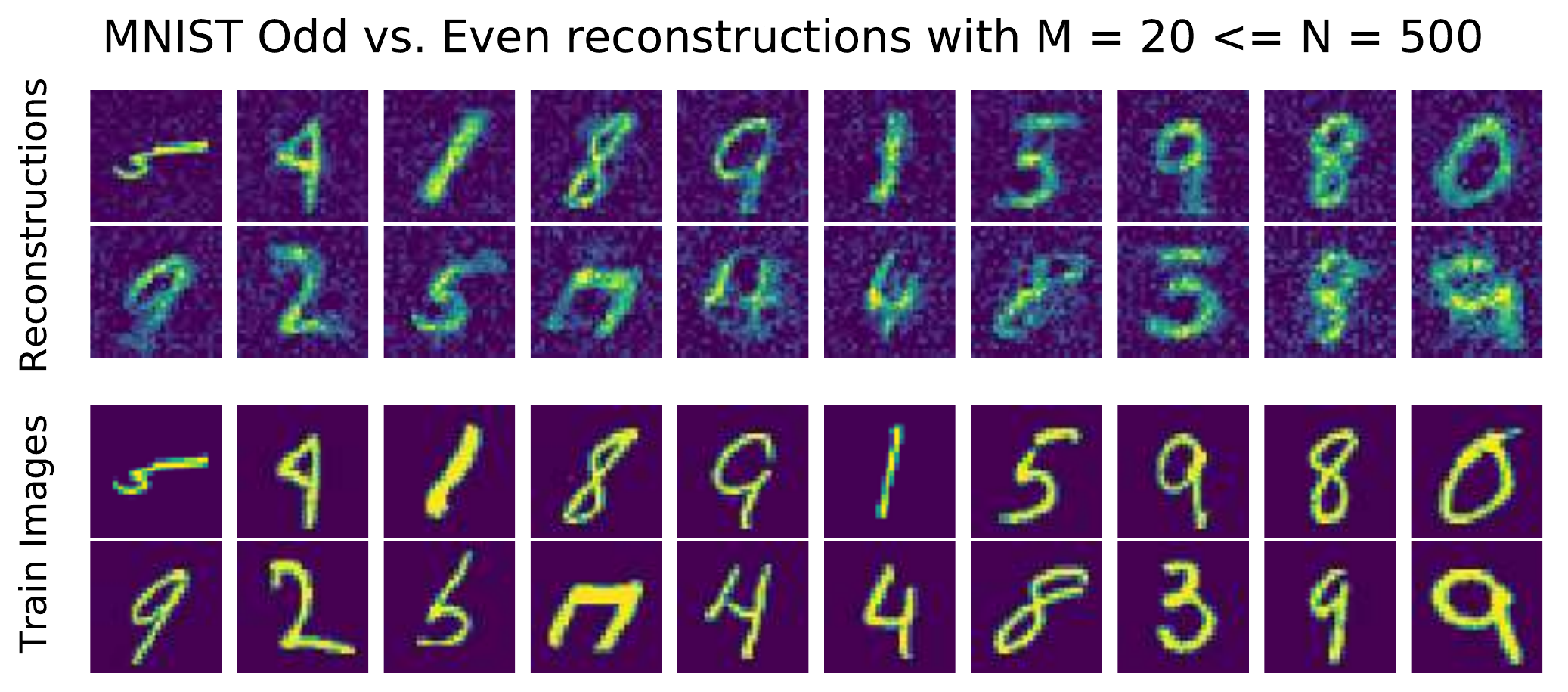}
\vskip -0.1in
\caption{Reconstructing 20 images from a dataset that may be larger than 20 images (MNIST Odd vs. Even)}
\label{app:fig:rkip_finite_mnist}
\end{center}
\vspace{-6mm}
\end{figure*}

\begin{figure*}[h!]
\vskip 0.1in
\begin{center}
\includegraphics[width = 0.45\linewidth]{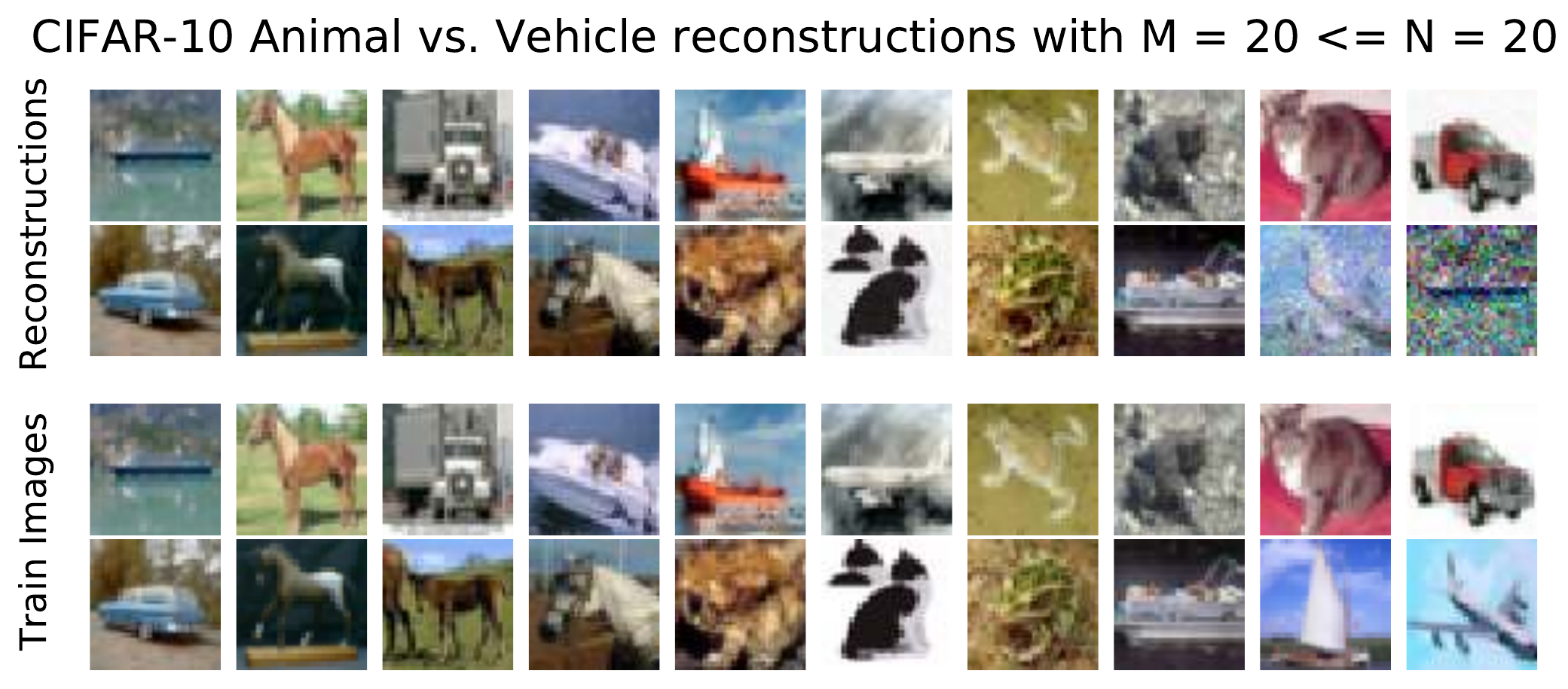}
\includegraphics[width = 0.45\linewidth]{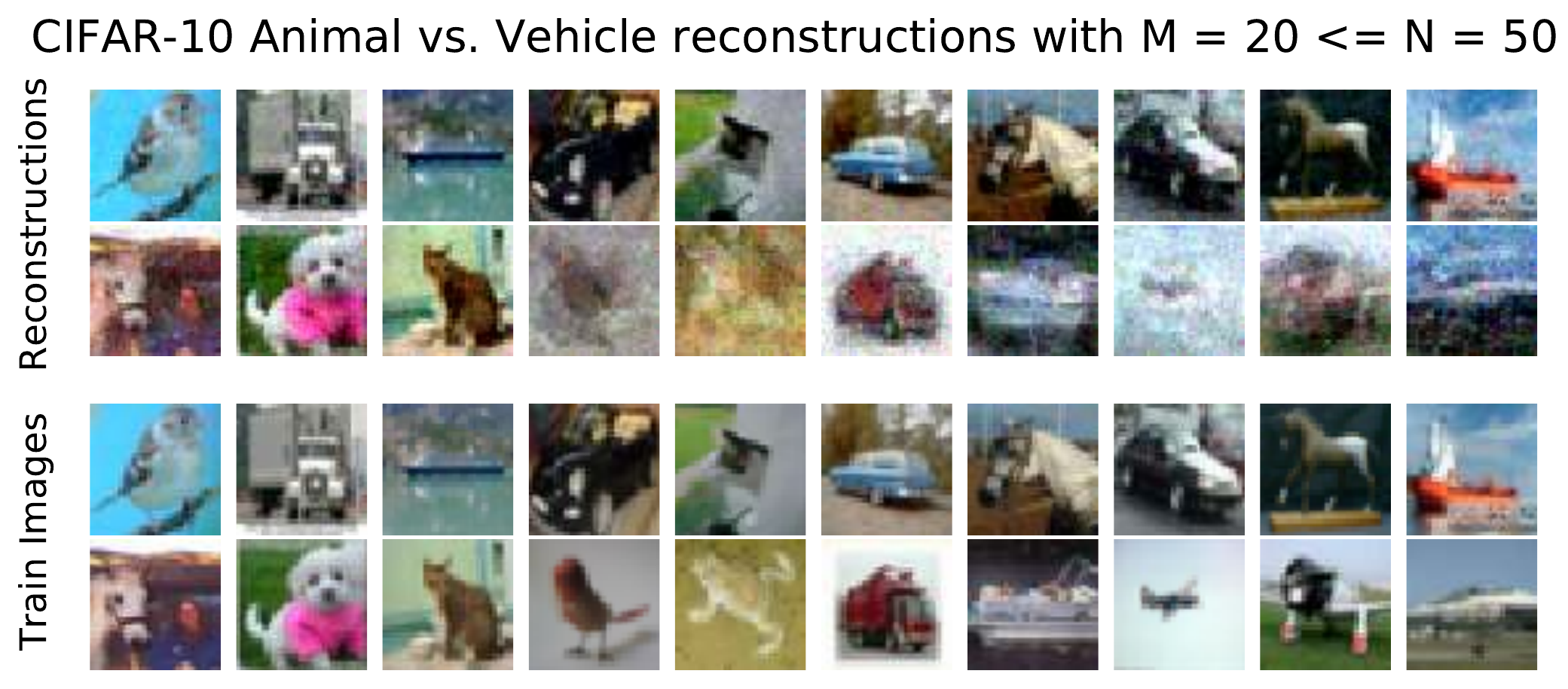}
\includegraphics[width = 0.45\linewidth]{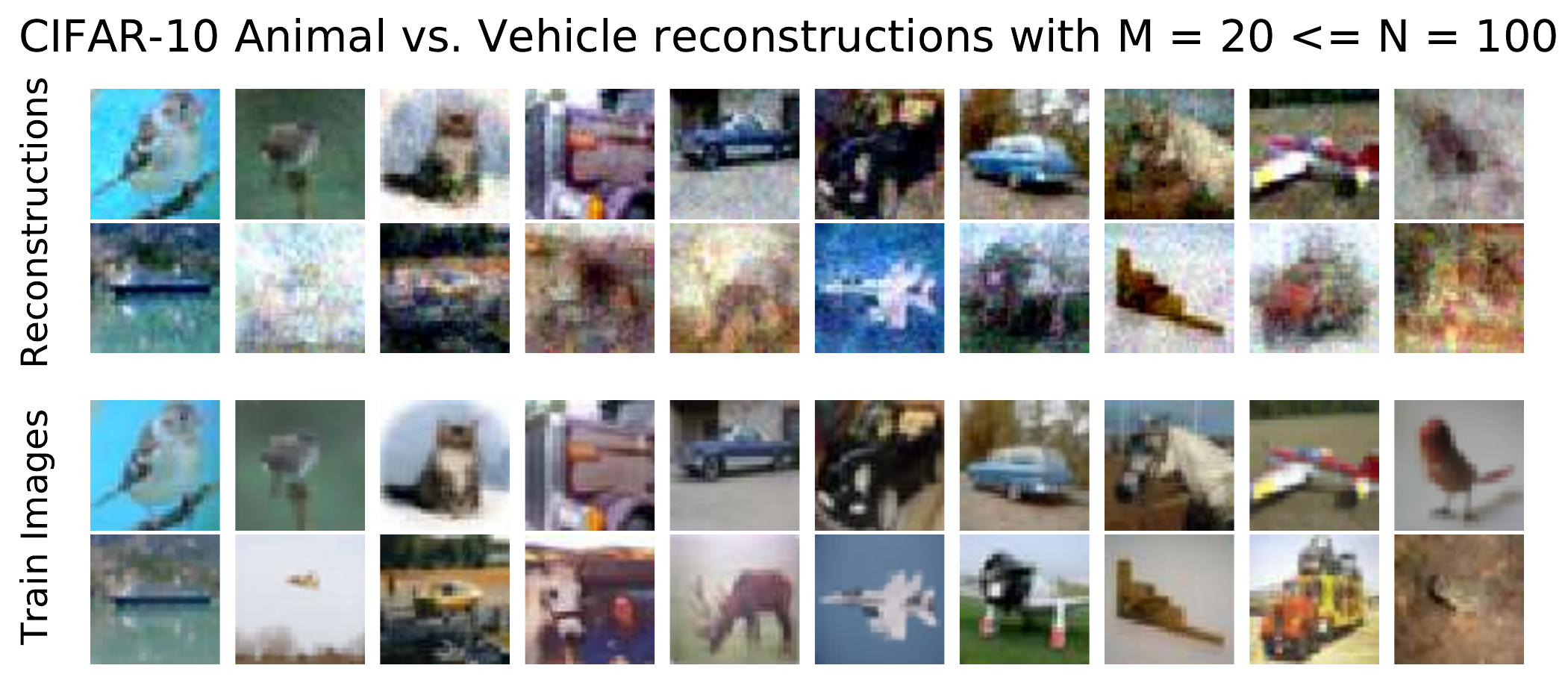}
\includegraphics[width = 0.45\linewidth]{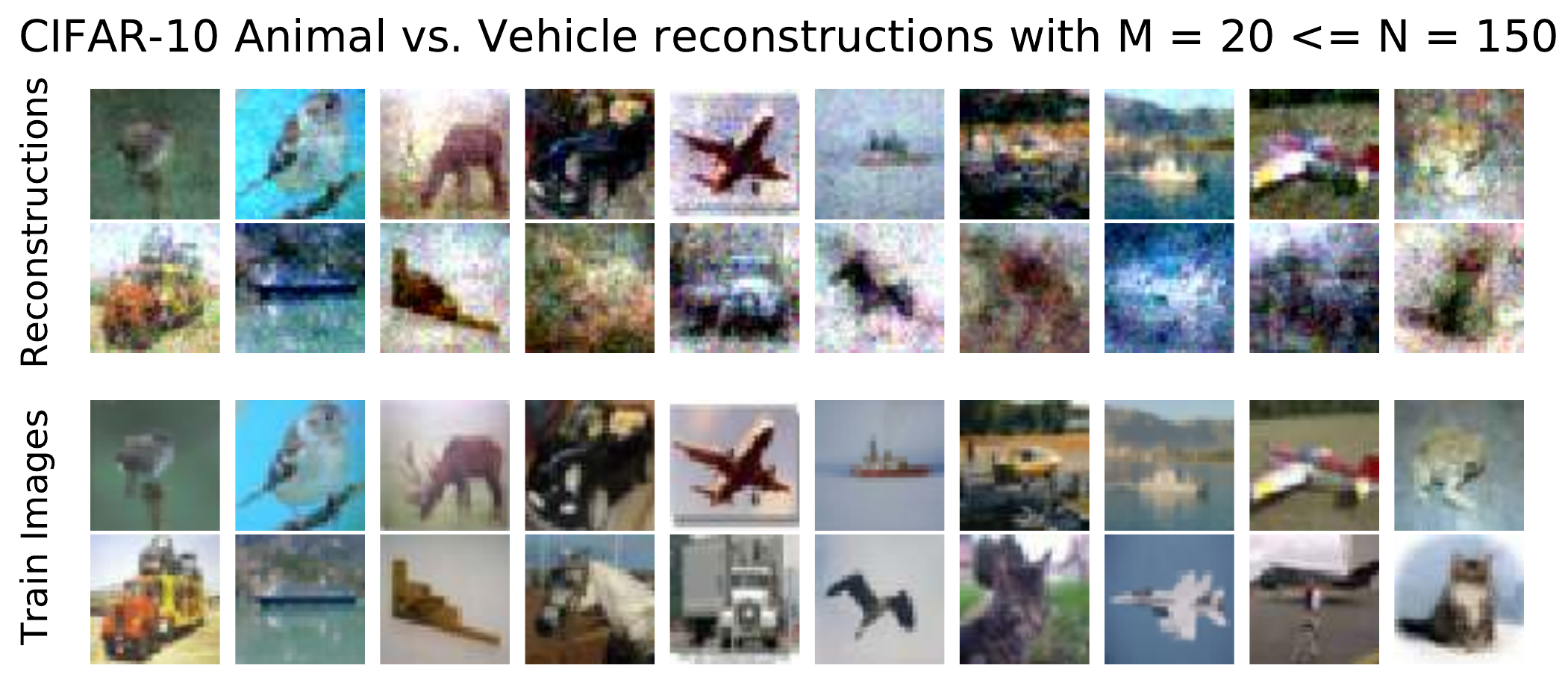}
\includegraphics[width = 0.45\linewidth]{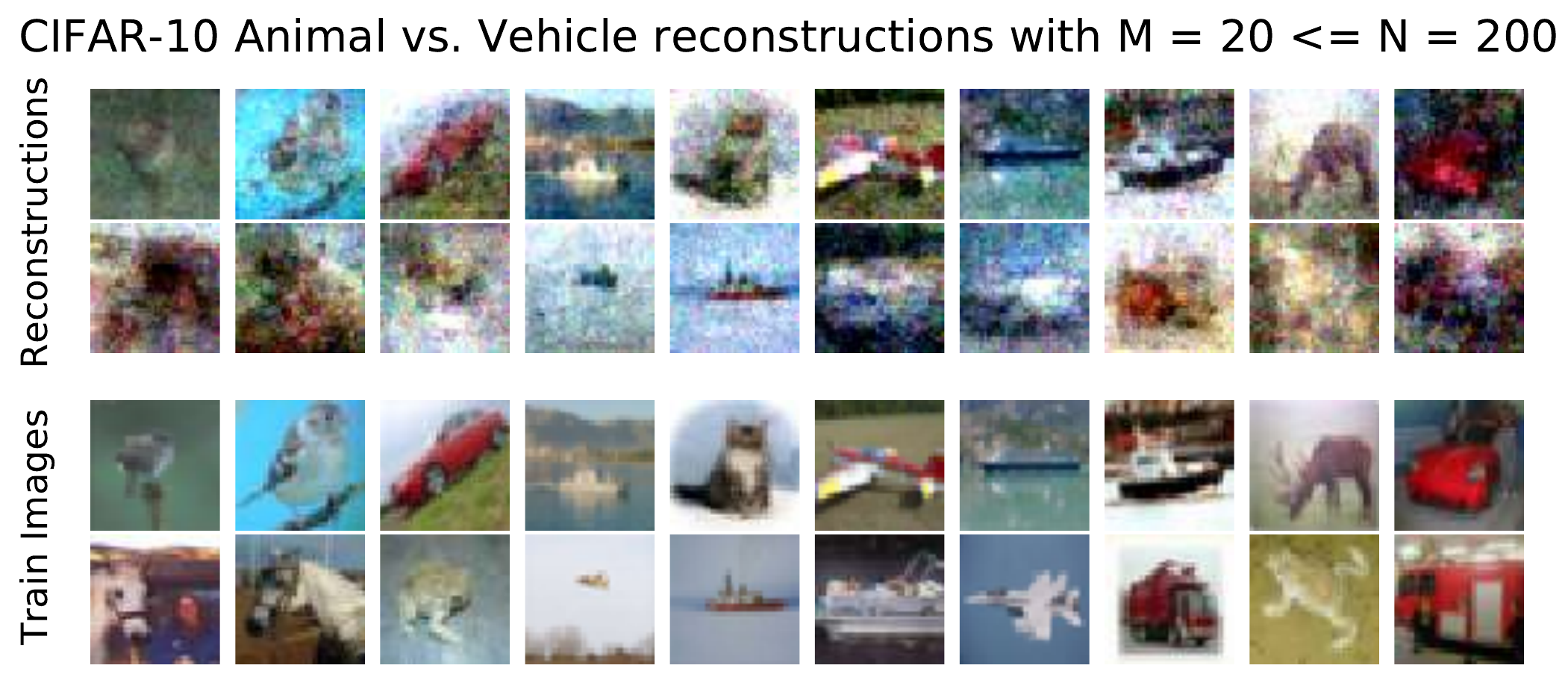}
\includegraphics[width = 0.45\linewidth]{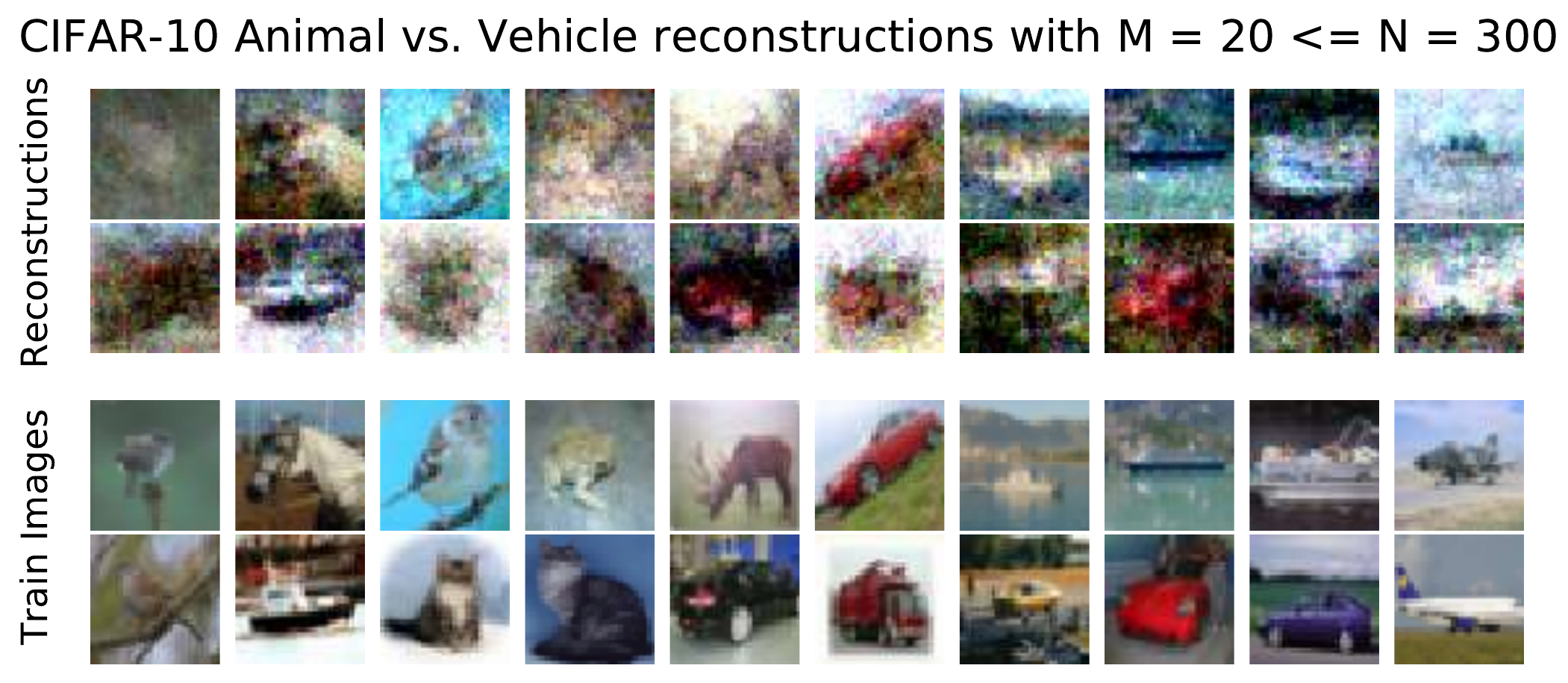}
\includegraphics[width = 0.45\linewidth]{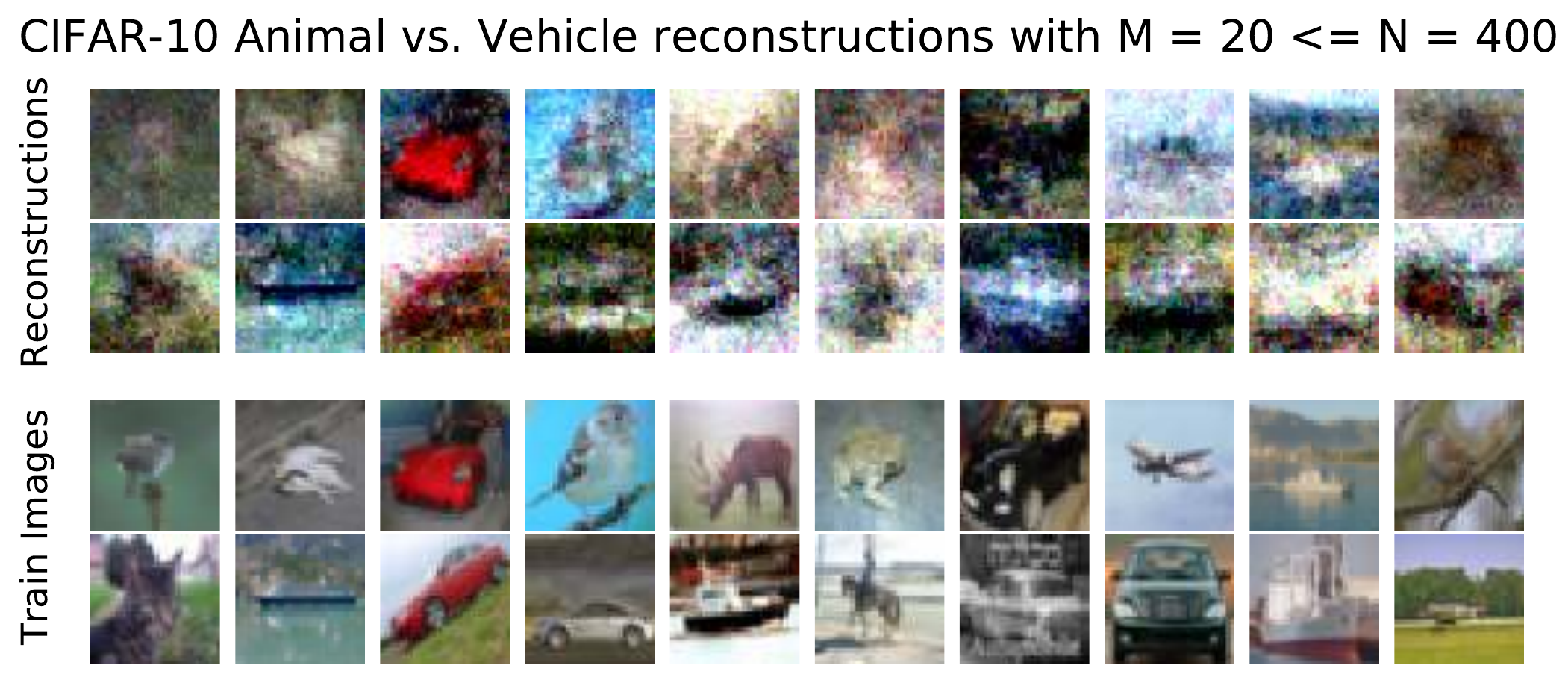}
\includegraphics[width = 0.45\linewidth]{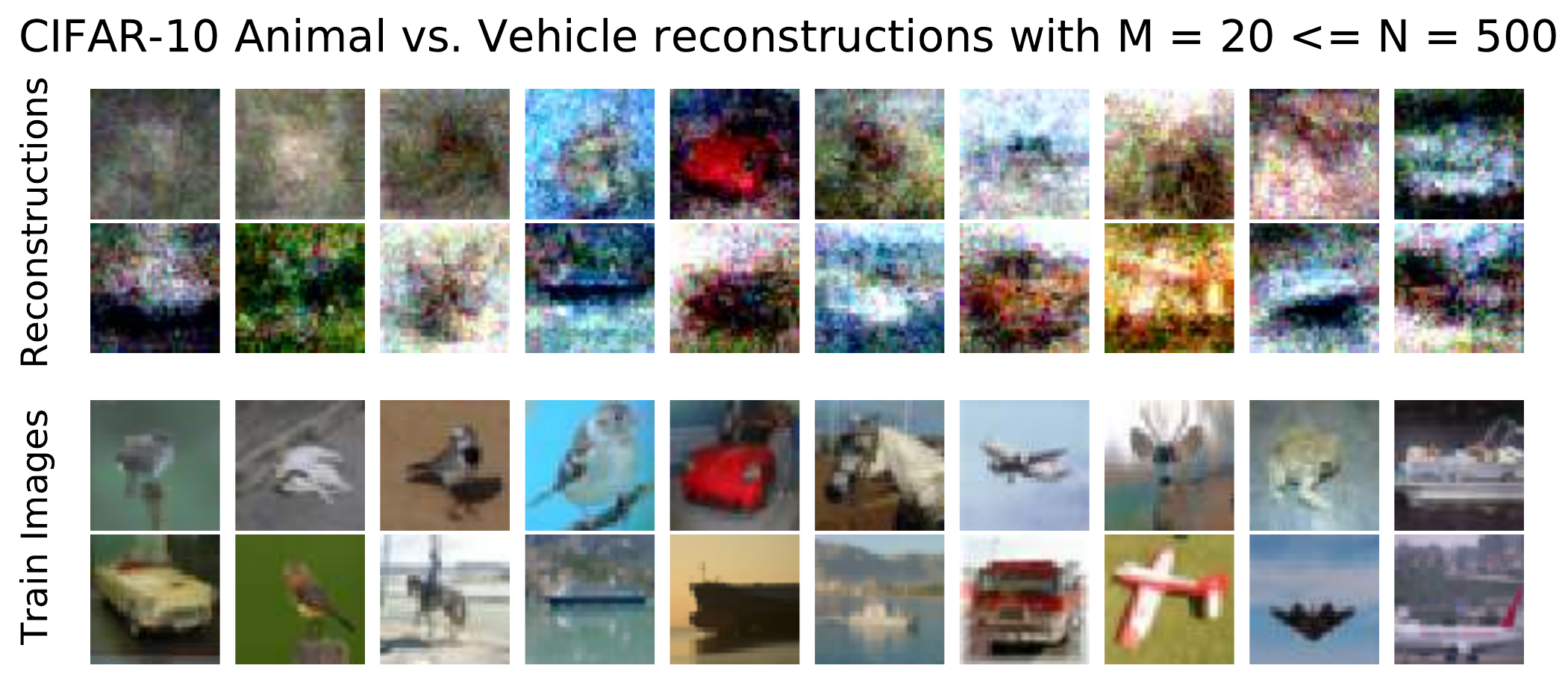}
\vskip -0.1in
\caption{Reconstructing 20 images from a dataset that may be larger than 20 images (CIFAR-10 Animal vs. Vehicle)}
\label{app:fig:rkip_finite_cifar10}
\end{center}
\vspace{-6mm}
\end{figure*}

\cref{app:fig:rkip_finite_mnist} and \cref{app:fig:rkip_finite_cifar10} show the resulting reconstructions when reconstructing 20 images from a dataset that may contain up to 500 images. Reconstructions are made from 4096 networks with linearized dynamics.

\section{Multiclass Reconstruction Loss}
\label{app:multiclass_loss}

Here we derive the multi-class reconstruction loss which we use in \cref{sec:finite_width_attack}. Our least-norm predictor satisfies the following conditioned (assuming that the network behaves in the linear regime):

\begin{align}
    \argmin_{\Delta \theta} \frac{1}{2} \|\Delta \theta\|^2_2 \quad &s.t.  \nonumber\\ 
    \forall c\in [C], \quad \Delta\theta^\intercal\nabla_\theta f_{\theta_0}^c(X_T) &= y_T^c - f_{\theta_0}^c(X_T) .
\end{align}

With $f_{\theta_0}^c$ referring to the network output on the $c$th class and $y_T^c$ referring to the training labels for the $c$th class. The network will converge to the least norm solution (norm of the difference from initialization), due to the network behaving in the lazy regime with gradient flow \citep{least_norm_stackexchange}. Writing the equation with dual variables our full Lagrangian is

\begin{align}
    \mathcal{L}(\Delta\theta, \alpha) = \frac{1}{2} \|\Delta \theta\|^2_2 + \sum_{c = 1}^C \alpha^{c \intercal} \left(\Delta\theta^\intercal\nabla_\theta f_{\theta_0}^c(X_T) - (y_T^c - f_{\theta_0}^c(X_T))\right).
\end{align}

With $\alpha$ our set of dual parameters $\in R^{C \times M}$, that is, we have a set of $M$ dual parameters for each class. Taking derives w.r.t $\Delta \theta$:

\begin{align}
     0 = \nabla_{\Delta \theta}\mathcal{L}(\Delta\theta, \alpha) = \Delta\theta + \sum_{c = 1}^C \alpha^{c \intercal} \left(\nabla_\theta f_{\theta_0}^c(X_T)\right).
\end{align}

So our multiclass reconstruction loss is

\begin{align}
     \mathcal{L}_{\textrm{reconstrution}} = \left\|\Delta\theta - \sum_{c = 1}^C \alpha^{c \intercal} \nabla_\theta f_{\theta_0}^c(X_T) \right\|^2_2 .
\end{align}

We can use the same argument as in \cref{sec:infinite_width_attack} to show this attack is exact in the infinite width limit.

\section{Early Stopping and Cross-Entropy}
\label{app:early_stopping_xent}

\subsection{Early Stopping}

\begin{figure}[h]
\begin{center}
\includegraphics[width = 1.0\linewidth]{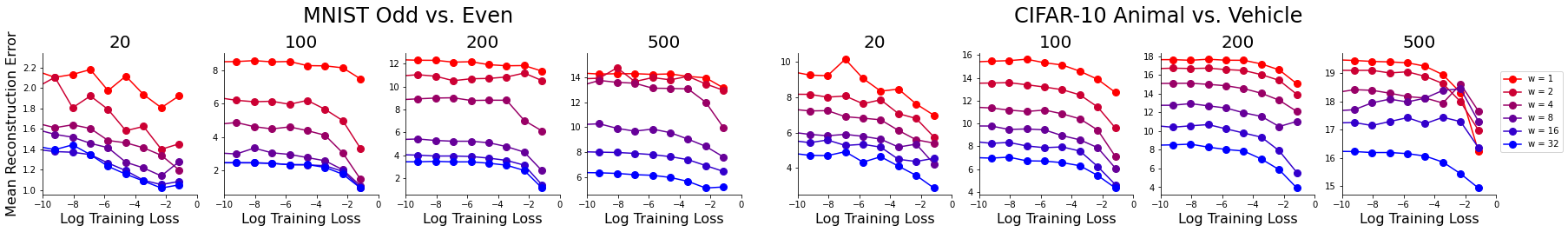}
\vskip -0.1in
\caption{Mean reconstruction errors for networks trained to various early stopping losses from $10^{-1}$ to $10^{-6}$}
\label{fig:reconstruction_curve_early_stopping_more}
\end{center}
\vspace{-5mm}
\end{figure}

\begin{figure}[h]
\begin{center}
\includegraphics[width = 0.45\linewidth]{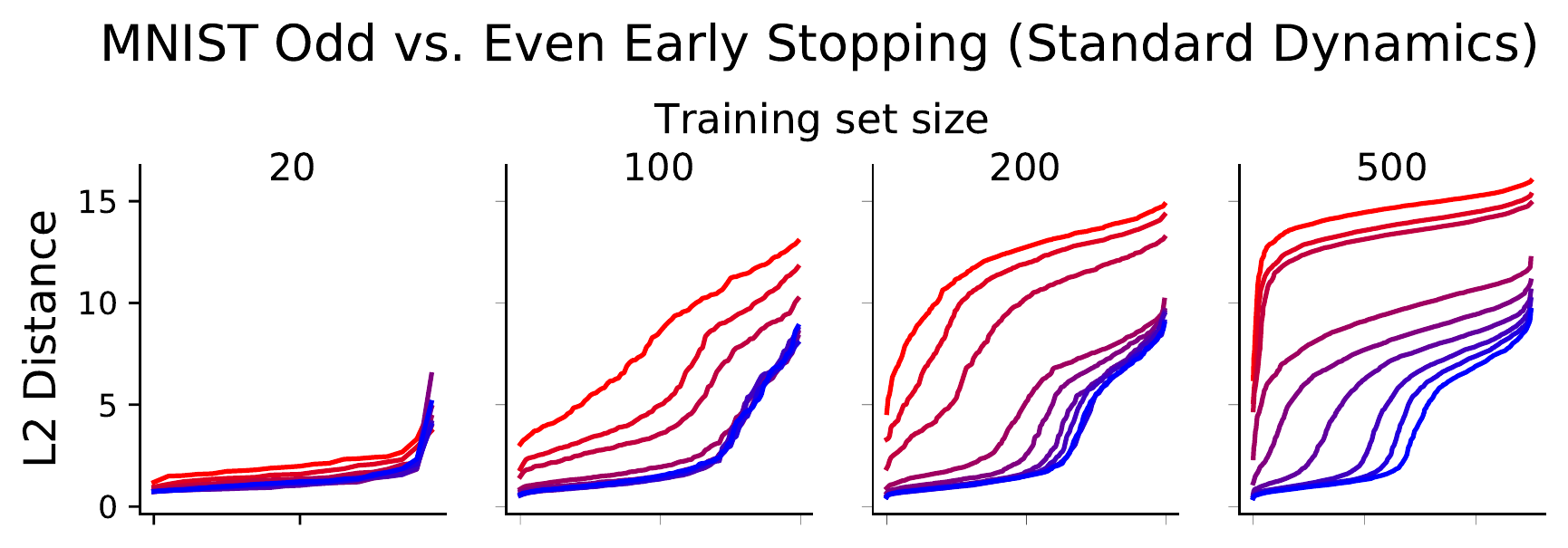}
\includegraphics[width = 0.45\linewidth]{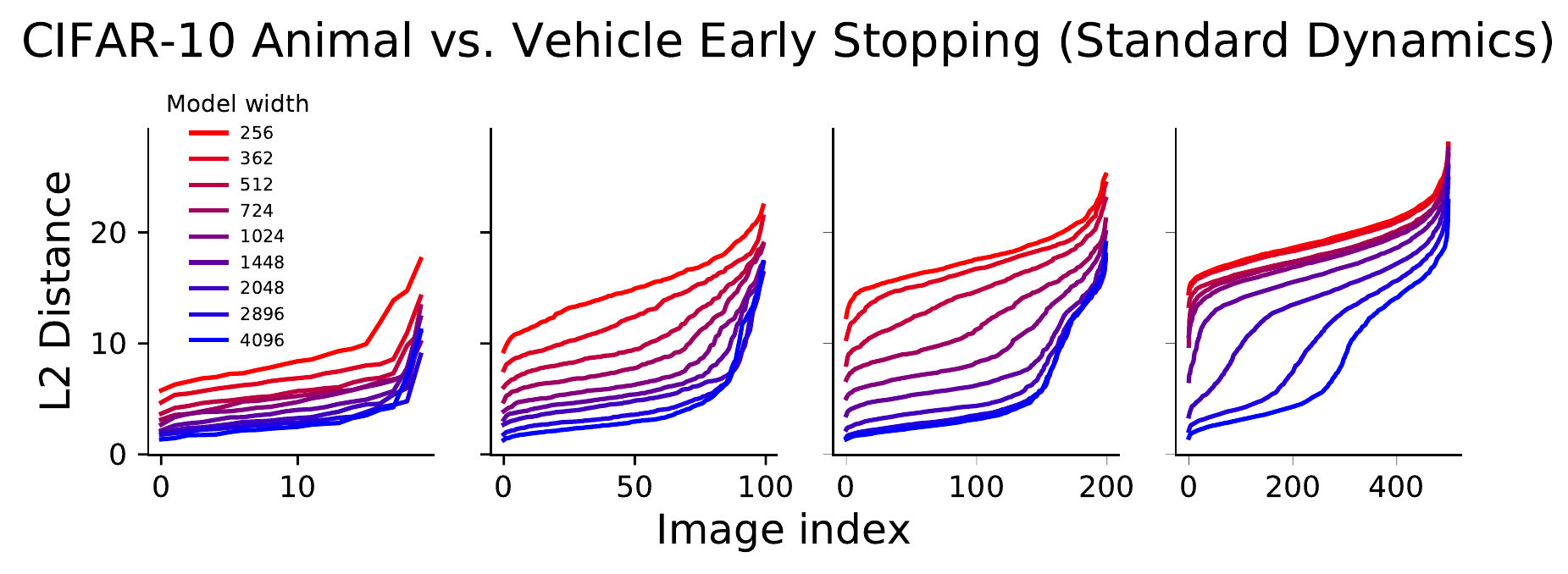}
\vskip -0.1in
\caption{Reconstruction curves for networks trained to a loss of 1e-2, i.e. significant underfitting (under standard dynamics). Compared to \cref{fig:reconstruction_curve_main}, we see that reconstruction quality is unaffected by early stopping, consistent with the theory.}
\label{fig:reconstruction_curve_early_stopping}
\end{center}
\vspace{-5mm}
\end{figure}

In the main text, we considered models trained with low learning rates for $10^6$ epochs, so that we achieve the KKT conditions described in \cref{sec:attack_description}. In practice, this is computationally expensive, and often ill-advised due to overfitting, with early stopping being a common regularization technique. One major limitation of the attack proposed in \citet{reconstruction} is that it requires the network to reach the KKT point to perform the attack. Our method \textbf{does not require the model to reach convergence} for the attack to work. Again we note that the time evolution of network parameters is given by:

{\small
\begin{align*}
    \Delta \theta(t) =\nabla_\theta f_{\theta_0}(X_T)^\intercal \underbrace{K_0^{-1} \left(I - e^{-\eta K_0 t}\right)\Big( y_T - f_{\theta_0}(X_T) \Big) }_{\text{time-dependent weights, $\alpha (t)$}}
\end{align*}
}
Notably, even at finite time, the change in network parameters is still a linear combination of the finite-width NTK feature maps of the training set, $\nabla_\theta f_{\theta_0}(X_T)$, with the indicated  \textit{time dependent} weights, $\alpha (t)$. Note that the success of the attack in infinite width relies on the injectivity of the kernel measure embedding, not that $\alpha(t)$ be at its converged value, implying the attack works with early stopping. The caveat with the early stopping attack is that we cannot necessarily hope to recover the original training labels.

We verify that this attack works in practice by repeating the attack procedure in \cref{sec:finite_width_attack}, with early stopping. We apply the attack on networks that achieve a mean training loss of from $10^{-1}$ to ${10^{-8}}$ (note that this means that on for $\{+1, -1\}$ labels, the outputs were around $0.45$ off, i.e. quite underfit, in the case of $\mathcal{L} = 10^{-1}$), with results shown in \cref{fig:reconstruction_curve_early_stopping_more}, with the specific reconstruction curve for $\mathcal{L} = 10^{-2}$ in \cref{fig:reconstruction_curve_early_stopping}. We observe that early stopping in general \textit{improves} reconstruction quality. We posit that there are two possible reasons for this: firstly, that early in training there is less time for the kernel to evolve, so the network exhibits network dynamics closer to the lazy regime. Secondly, we hypothesize that this could be that early in training all datapoints have a roughly equal contribution to the parameter changes, whereas later in training, certain datapoints have a stronger influence (see \cref{sec:which_get_attacked} and \cref{app:alpha_as_loss}). When some datapoints have a much stronger influence than others, this could cause the the less influential datapoints to be "drowned out" by the signal of the more influential ones. Future work could study the complex relationship between early stopping, outlier influence, and network vulnerability more closely. We also note that in the limit of a single gradient step, our attack corresponds to \textit{gradient leakage} attacks \citep{gradient_leakage}.

\subsection{Cross Entropy Loss}

Following \citep{wide_linear_models}, we have that wide networks trained under cross-entropy loss also exhibit lazy dynamics provided they are of sufficient width. The corresponding ODE is:

\begin{align*}
    \mathcal{L}(\theta_t) &= - \sum_i y_i\log\sigma(f_{\theta_t}(x_i)) \\
    \frac{\partial \theta_t}{\partial t} &= -\eta \frac{\partial \mathcal{L}(\theta_t)}{\partial \theta_t} \\
    \frac{\partial \theta_t}{\partial t} &= -\eta \sum_i (\sigma(f_{\theta_t}(x_i) - y_i) \frac{\partial f_{\theta_t} (x_i)}{\partial \theta_t}\\
\end{align*}

Unlike \cref{eq:recon_solution_to_opt_problem}, there is not a closed form solution to this, however the key point is that $\Delta\theta$ still is a linear combination of $\nabla_\theta f_\theta (x_i)$.

\section{Experimental Details}
\label{app:experiment_details}

\subsection{Libraries and Hardware}
We use the JAX, Optax, Flax, and neural-tangents libraries \citep{jax2018github, deepmind2020jax, flax, neural_tangents, fast_finite_NTK}. All experiments were run on Nvidia Titan RTX graphics cards with 24Gb VRAM.

\subsection{Network training}
Unless otherwise stated, networks trained on real data are trained for $10^6$ iterations of full batch gradient descent, with SGD with momentum 0.9. For the learning rate, we set $\eta = N \times 2\mathrm{e}{-7}$, where $N$ is the number of training images. For distilled data, we use a learning rate of $\eta = N \times 6\mathrm{e}{-6}$, where $N$ is now the distilled dataset size. We did not find that results were heavily dependent on the learning rates used during training. Additionally, if the training loss was less than $1\mathrm{e}{-10}$, we terminated training early. Every reconstruction curve in the main text is the average of 3 unique networks trained on 3 unique splits of training data.

For binary classification, we use labels in $\{+1, -2\}$, and for 10-way multiclass classification, we use labels of 0.9 corresponding to the selected class and -0.1 for other classes.

\subsection{Reconstructions}
To create reconstructions, we initialize reconstruction images with a standard deviation of 0.2, and dual parameters to be uniform random within $[-0.5, 0.5]$. We use Adam optimizer \citep{adam}, with a learning rate of 0.02 for all reconstructions. As stated in \cref{app:comparison_to_haim}, these could be fine-tuned to improve performance. We optimize the images for 80k iterations. Like with \citet{reconstruction}, we found that it was useful to use a \texttt{softplus} rather than a \texttt{Relu} during reconstruction, owing to the smoothing gradient loss. We annealed \texttt{softplus} temperature from 10 to 200 over the course of training, so that we are effectively using \texttt{ReLU}s by the end of training. Unless otherwise stated, we aim to reconstruct $M = 2N$ reconstruction images, with $N$ the training set size.

\subsection{Distillation}
We initialize distilled images with a standard deviation of 0.2 and distill for 50k iterations with Adam optimizer with a learning rate of 0.001.

\subsection{Fine-Tuning Experiments}
For the fine-tuning experiments in \cref{sec:fine_tuning}, we use the pretrained ResNet-18 from flaxmodels \citep{flaxmodels}. For these experiments, we used 64-bit training, as we found it improved reconstruction quality. We train for 30000 iterations with a learning rate of $1e-5$. We use SGD with no momentum, and also freeze the batchnorm layers to the parameters used by the initial model. We use the hybrid loss described in \cref{app:hybrid_loss}.

\subsection{Pruning Experiments}
For the pruning experiments we train for $10^5$ epochs for each training run with a learning rate of $n \times 3\textrm{e}-6$ with SGD with momentum 0.9. For reconstruction we use the same reconstruction attack in the main text for 30000 iterations. For each training iteration we set with width of the network to be $w = 55\sqrt{n}$, as we found that generally kept the strength of the attack the same for different values of $n$. If we make the attack too strong, then too many training points will be reconstructed and we will no longer have any notion of ``easy to reconstruct" examples, as all examples will be reconstructed equally well.

\subsection{Reconstruction Post-Processing}
We do not post-process our reconstructions and the reconstruction curves and visualization are based on unmodified reconstructions. Note that \citet{reconstruction} has a more complex reconstruction scheme involving rescaling the reconstructions and averaging, which is detailed in \citet{reconstruction}'s appendix.

\section{Choice of Kernel}
\label{app:hybrid_loss}
During reconstruction, there is a small choice we can make, specifically whether we using the initialization gradient tangent space, $\nabla_{\theta_0} f_{\theta_0}(x_i)$ or the final gradient tangent vector space, $\nabla_{\theta_f} f_{\theta_f}(x_i)$. Specifically, we can choose to optimize $\mathcal{L}_f$ or $\mathcal{L}_0$:

\begin{align*}
    \mathcal{L}_f &= \Big\|\Delta \theta - \sum_{\mathclap{s_j^R \in S^R}} \alpha_j^R \nabla_{\theta_f} f_{\theta_f} (x_j^R)\Big\|^2_2     \\
    \mathcal{L}_0 &= \Big\|\Delta \theta - \sum_{\mathclap{s_j^R \in S^R}} \alpha_j^R \nabla_{\theta_0} f_{\theta_0} (x_j^R)\Big\|^2_2     
\end{align*}

Under linearized dynamics, these two are equal as $\nabla_\theta f_\theta(x)$ does not change. However for standard dyanmcis there is a small difference, with the difference increasing the more the kernel changes. For the results in the main text, we use $\mathcal{L}_f$. We found that this has 6marginally better performance than $\mathcal{L}_0$, but this difference is rather minor. This also leads to a third choice of reconstruction loss, which we call the ``hybrid" loss $\mathcal{L}_h$, which considers the change in parameters a mixture of both the final and initial kernel:

\begin{align*}
    \mathcal{L}_h &= \Big\|\Delta \theta - \sum_{\mathclap{s_j^R \in S^R}} \alpha_{j, 0}^R \nabla_{\theta_f} f_{\theta_f} (x_j^R) - \sum_{\mathclap{s_j^R \in S^R}} \alpha_{j, f}^R \nabla_{\theta_0} f_{\theta_0} (x_j^R)\Big\|^2_2     \\
\end{align*}

In which we have two sets of dual parameters $\alpha_0$ and $\alpha_f$. This further increases performance, but in general is twice as slow as optimizing $\mathcal{L}_0$ or $\mathcal{L}_f$, and we chose to use this loss for the ResNet-18 fine-tuning experiments, since that setting is more challenging. Note that this attack could be generalized to include multiple checkpoints $\theta_t$ along the trajectory and more dual parameters, however of course this would require access to more training checkpoints.
\section{Additional Results}
\label{app:additional_results}

\subsection{Convolutional Architectures}
\label{app:conv_results}

\begin{figure}
\vspace{-5mm}
\vskip 0.2in
\begin{center}
\includegraphics[width = \linewidth]{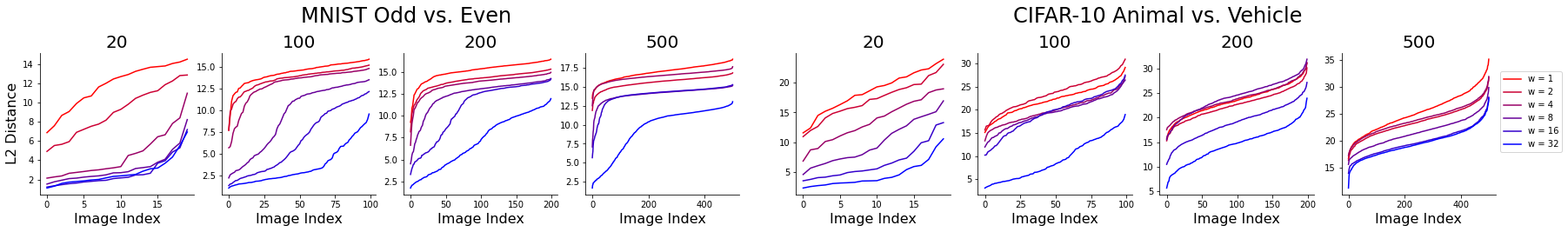}
\vskip -0.1in
\caption{Reconstruction curves for convolutional architectures trained on MNIST Odd vs. Even and CIFAR-10 Animal vs. Vehicle classification with varying width multipliers from $1-32$}
\label{app:fig:lenet_curve}
\end{center}
\vskip -0.2in
\end{figure}

Here we provide results for our attack applied to convolutional architectures. We trained networks on binary MNIST and CIFAR-10 classifications tasks, as we did in the main text, but trained on a convolutional architecture from scratch. We use the common LeNet-5 architecture with width multipliers ranging from 1-32. \cref{app:fig:lenet_curve} shows the results. The findings observed on two-layer networks still apply in this settings, however our attack struggles more in deeper architectures.

\subsection{Tiny-ImageNet}
\label{app:timagenet_results}

\begin{figure}
\vspace{-5mm}
\vskip 0.2in
\begin{center}
\includegraphics[width = \linewidth]{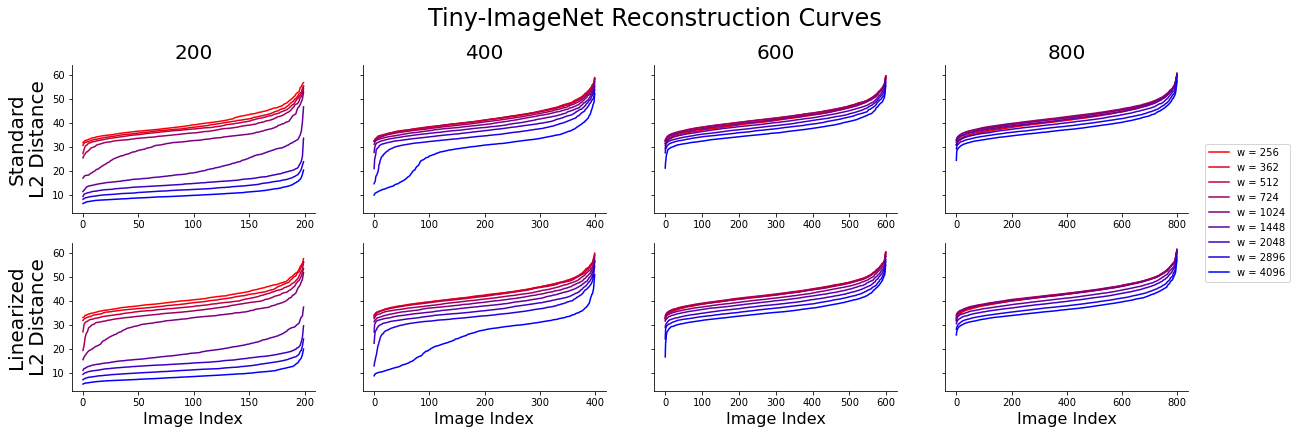}
\vskip -0.1in
\caption{Reconstruction curves for networks trained on Tiny-ImageNet 200-way classification}
\label{app:fig:timagenet_curve}
\end{center}
\vskip -0.2in
\end{figure}

\begin{figure}
\vspace{-5mm}
\vskip 0.2in
\begin{center}
\includegraphics[height=2.0in]{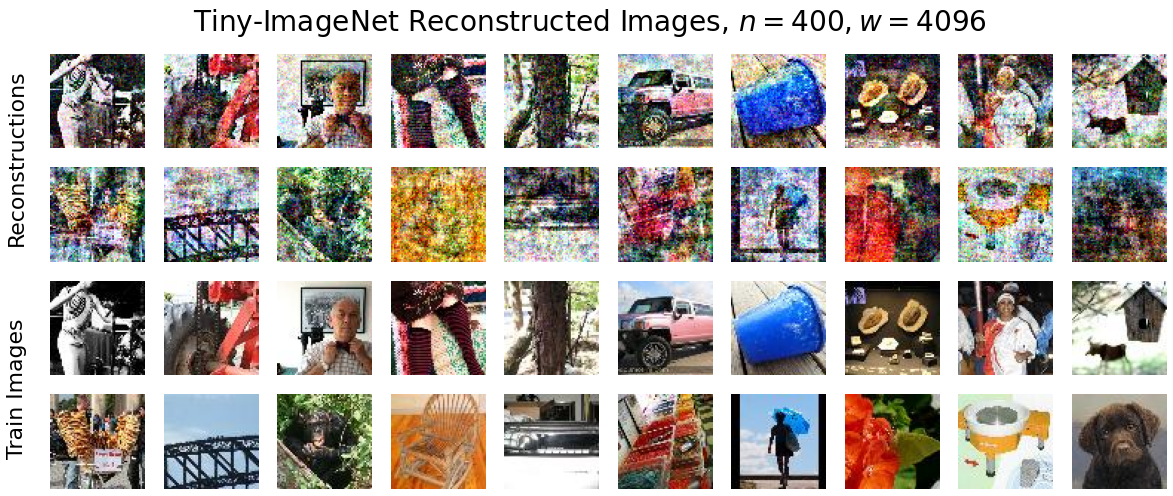}
\vskip -0.1in
\caption{Reconstructed images and their nearest train iamge counterparts for Tiny-ImageNet reconstruction for a $w=4096$ network with $n=400$}
\label{app:fig:timagenet_images}
\end{center}
\vskip -0.2in
\end{figure}

The main text works mainly with classification on small, low-resulution datasets such as MNIST and CIFAR-10. Here we consider more complex datasets with higher resolution by applying our attack to Tiny-ImageNet classification. Tiny-ImageNet consists of 200 classes with images of resolution $64\times 64$ \citep{tinyimagenet}. As the quality of the NTK approximation is negatively affected by image resolution \citep{jacotntk}, this experiment serves as an important testing ground of the viability of the attack for higher resolution images. We show the results on few-shot classification, consider $1-4$ images per classes in \cref{app:fig:timagenet_curve} and with the reconstructed images in \cref{app:fig:timagenet_images}. Reconstructing higher resolution images is more challenging and improving this attack on high resolution images is an interesting direction for future work.

\subsection{Kernel Distance vs. Reconstruction Quality Scatter plots for Multiclass Classifications}
\label{app:kernel_distance_scatter_multiclass}

\begin{figure*}[h]
\vskip 0.1in
\begin{center}
\includegraphics[height = 2.0in]{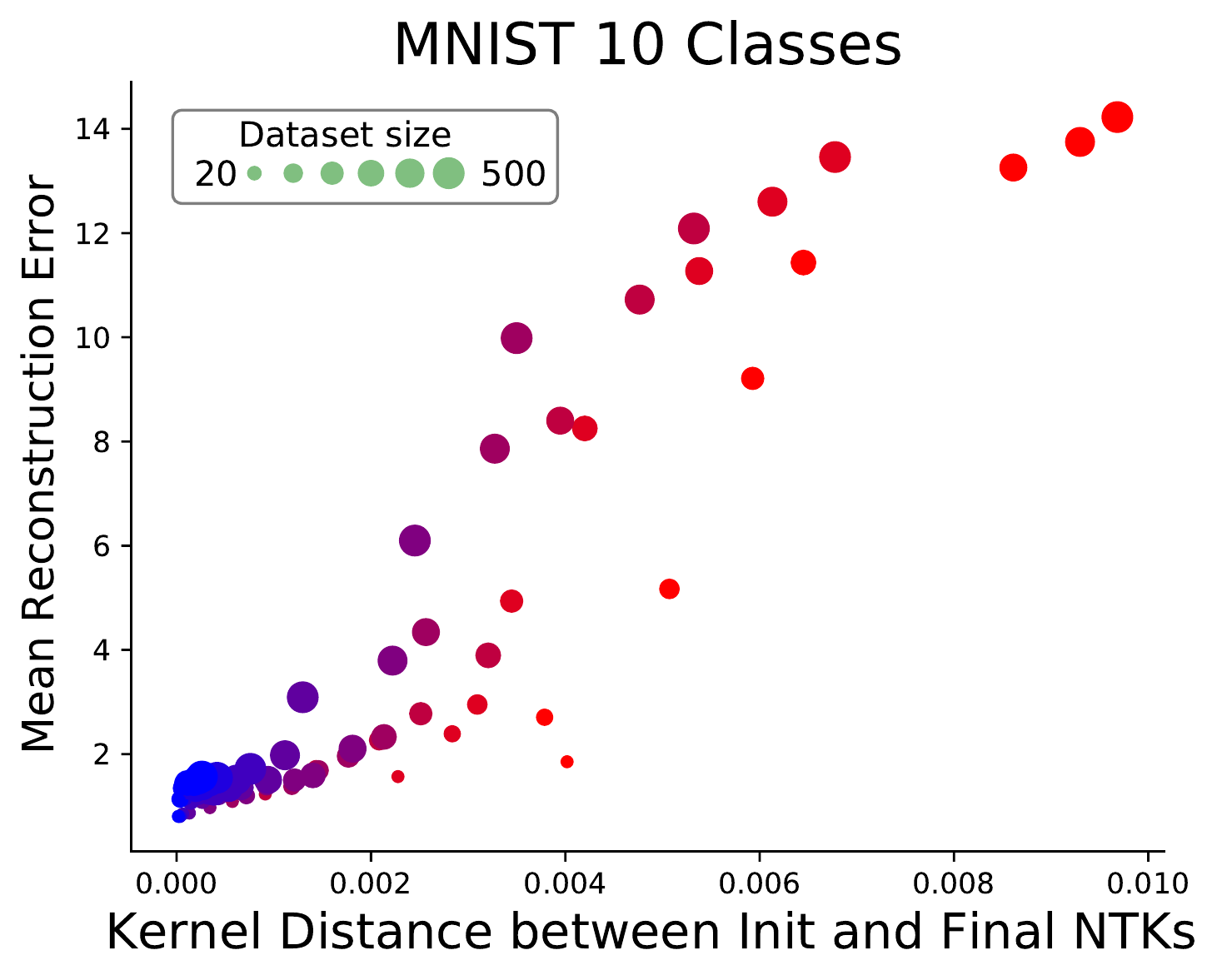}
\includegraphics[height = 2.0in]{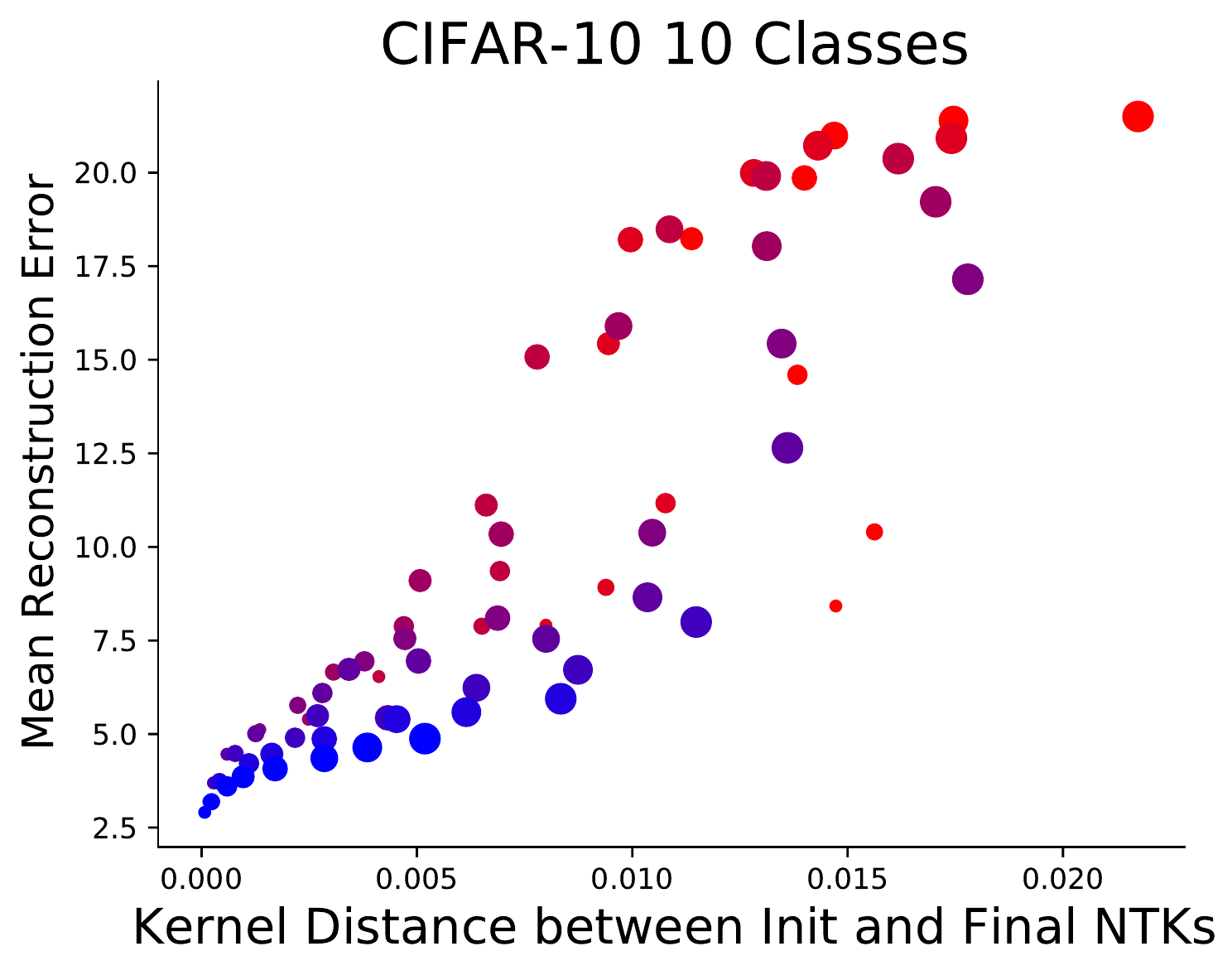}
\vskip -0.1in
\caption{Mean reconstruction error vs. the kernel distance from the initialization kernel to the final kernel for multiclass classification. The mean reconstruction error, measured as the average value of the reconstruction curve, is strongly correlated with how much the finite-width NTK evolves over training. Dataset size is given by dot size, while the color indicates model width. Multiclass classification sees significantly lower kernel distances (up to 1e-2 for MNIST and 2e-2 for CIFAR-10) compared to binary classification (up to 3e-2 for MNIST and 9e-2 for CIFAR-10, see \cref{fig:reconstruction_kernel_dist_scatter}), which may be the cause of better reconstruction quality.}
\label{fig:scatter_multiclass}
\end{center}
\vspace{-6mm}
\end{figure*}

\cref{fig:scatter_multiclass} shows the corresponding \cref{fig:reconstruction_kernel_dist_scatter} for multiclass classification. As we observed, multiclass classification has improved reconstruction quality. From \cref{fig:scatter_multiclass}, we see that multiclass classification sees significantly lower kernel distances (up to 1e-2 for MNIST and 2e-2 for CIFAR-10) compared to binary classification (up to 3e-2 for MNIST and 9e-2 for CIFAR-10, see \cref{fig:reconstruction_kernel_dist_scatter}), which may explain why the reconstructions have better quality.

\subsection{Extra Pruning Experiments}
\begin{figure}
\vspace{-5mm}
\begin{center}
\includegraphics[height=2.0in]{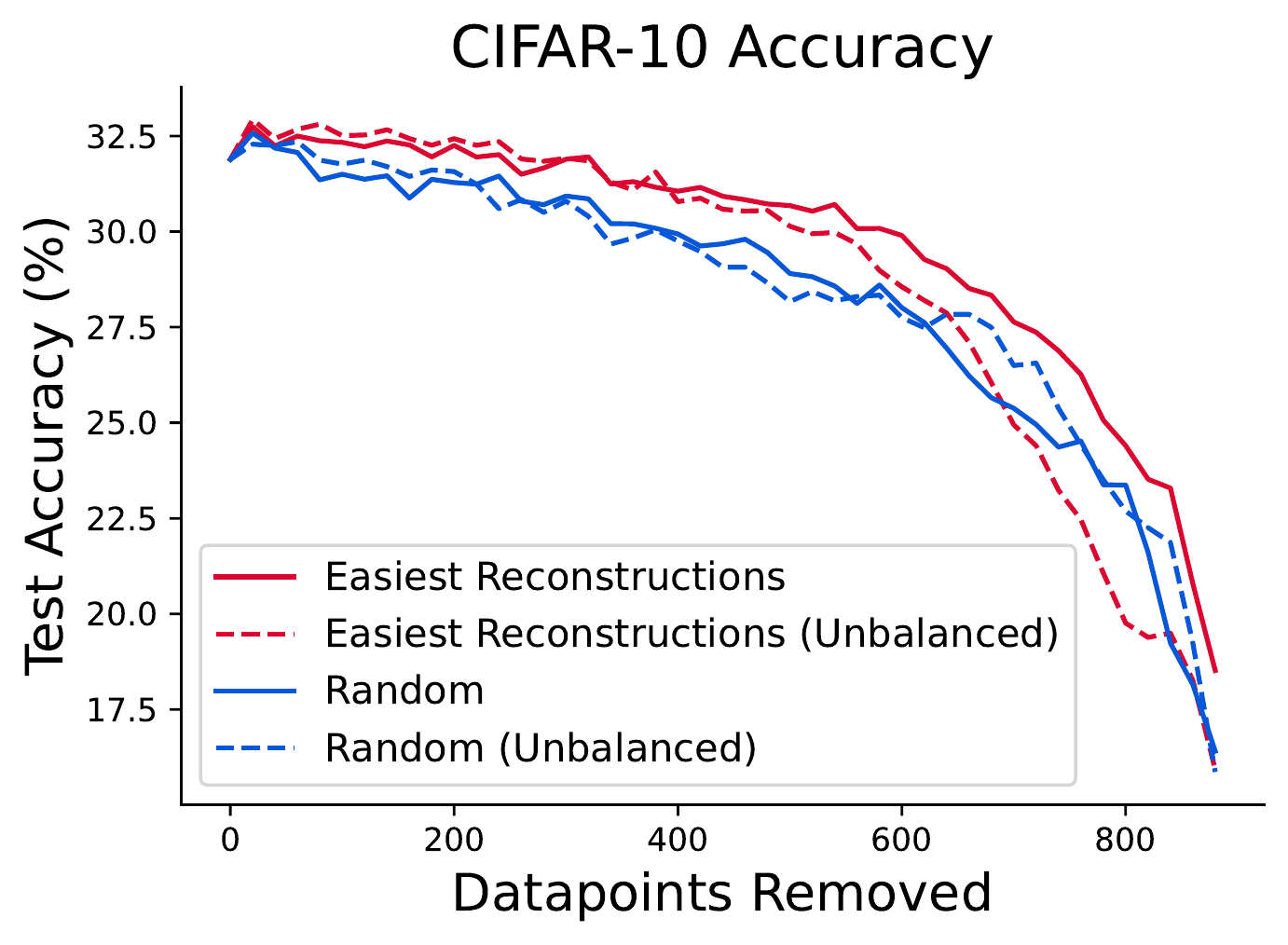}
\vskip -0.1in
\caption{Test accuracy of iteratively pruned CIFAR-10 using either random pruning or pruning based on easily reconstructed images with either class balanced subsets or non-balanced subsets.}
\label{app:fig:pruning_no_balance}
\end{center}
\vskip -0.2in
\end{figure}
In \cref{sec:which_get_attacked}, we considered removed class balanced subsets of the training data at each training iteration, either by random or by ease of reconstruction. If instead we allow class imbalance, we see the same behaviour as in \cref{sec:which_get_attacked} initially, but as more datapoints are removed, we see in \cref{app:fig:pruning_no_balance} that removing easy reconstructions results in strongly imbalanced classes, resulting in poor test accuracy. Understanding why some classes are more susceptible to reconstruction is likely related to the discussion in \cref{sec:which_get_attacked}. Additionally, we found that if we underfit the data, then we do not observe any difference in test accuracy for pruning random vs. reconstructions. This suggests that the effect of large $\alpha$ values only shows later in training, when some datapoints are well fit and others still underfit.

\newpage
\subsection{Additional Reconstruction Curves}

\begin{figure*}[h]
\vskip 0.1in
\begin{center}
\includegraphics[width = 0.9\linewidth]{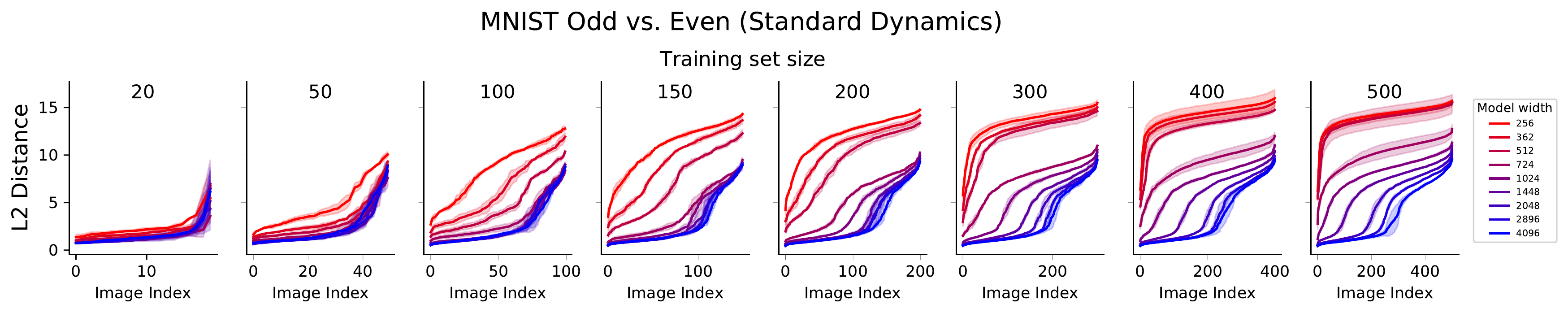}
\includegraphics[width = 0.9\linewidth]{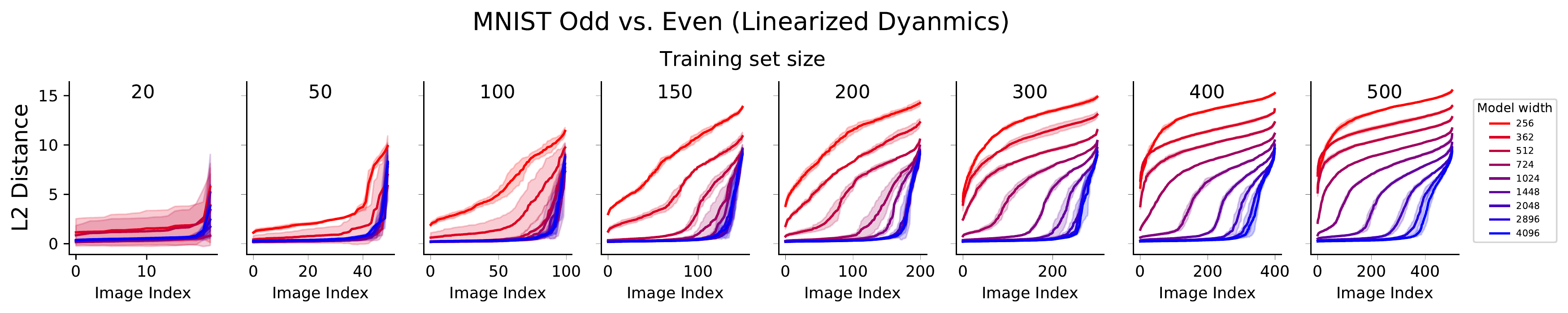}
\includegraphics[width = 0.9\linewidth]{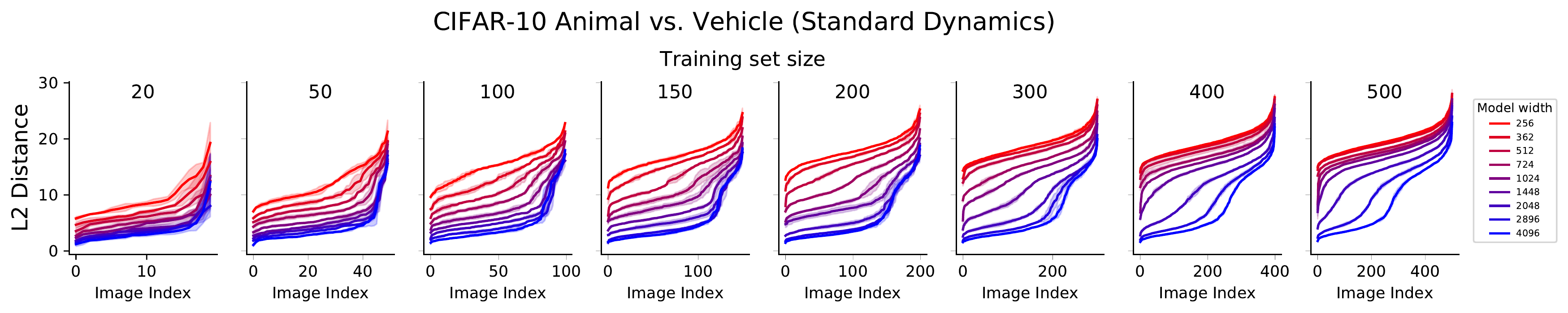}
\includegraphics[width = 0.9\linewidth]{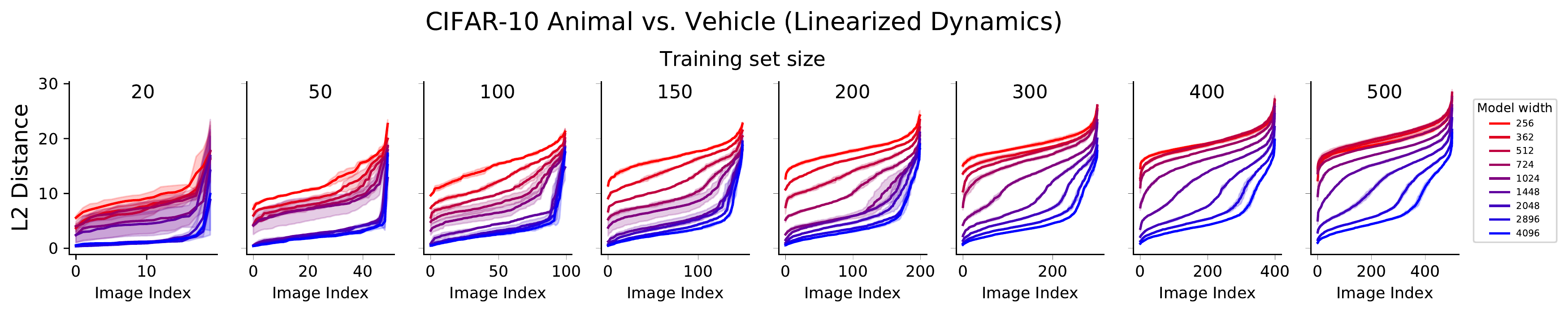}
\vskip -0.1in
\caption{Reconstruction curves for binary classification tasks}
\label{app:fig:binary_recons}
\end{center}
\vspace{-6mm}
\end{figure*}

\begin{figure*}[h]
\vskip 0.1in
\begin{center}
\includegraphics[width = 0.9\linewidth]{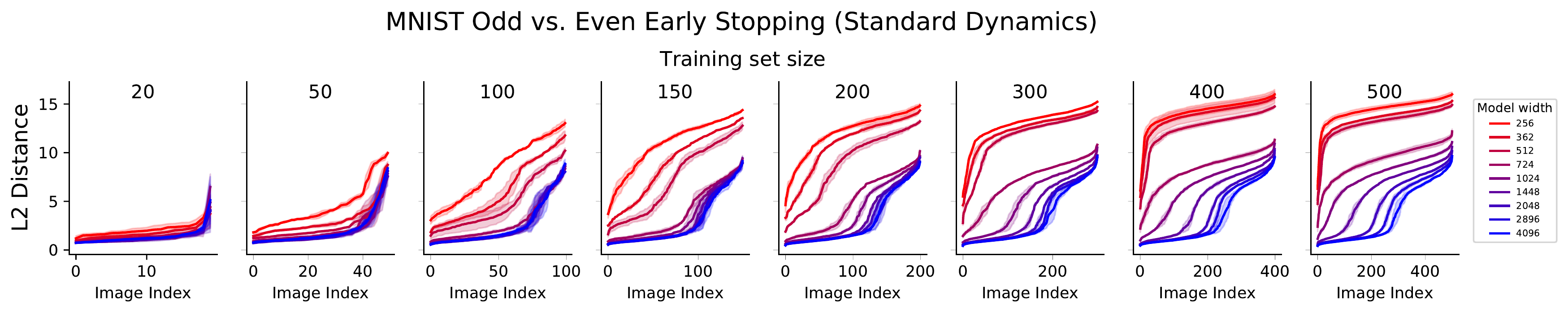}
\includegraphics[width = 0.9\linewidth]{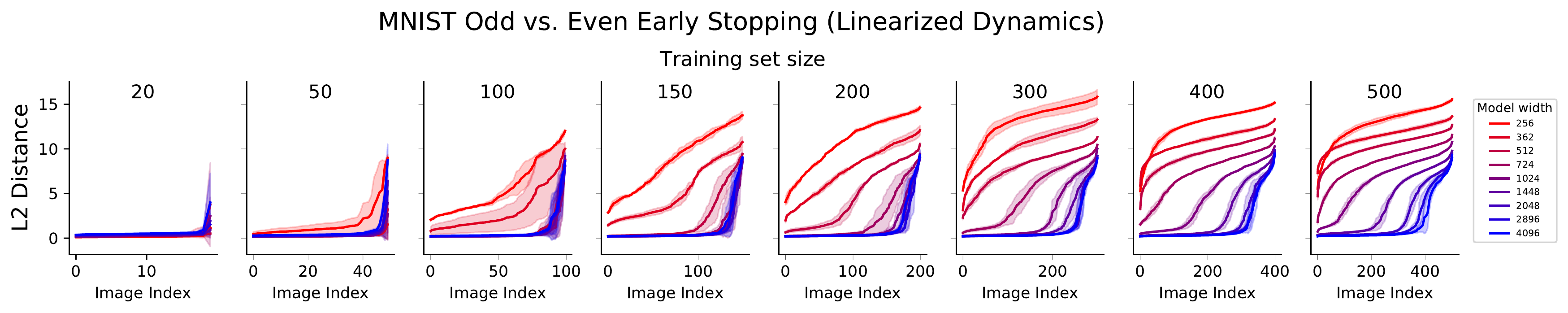}
\includegraphics[width = 0.9\linewidth]{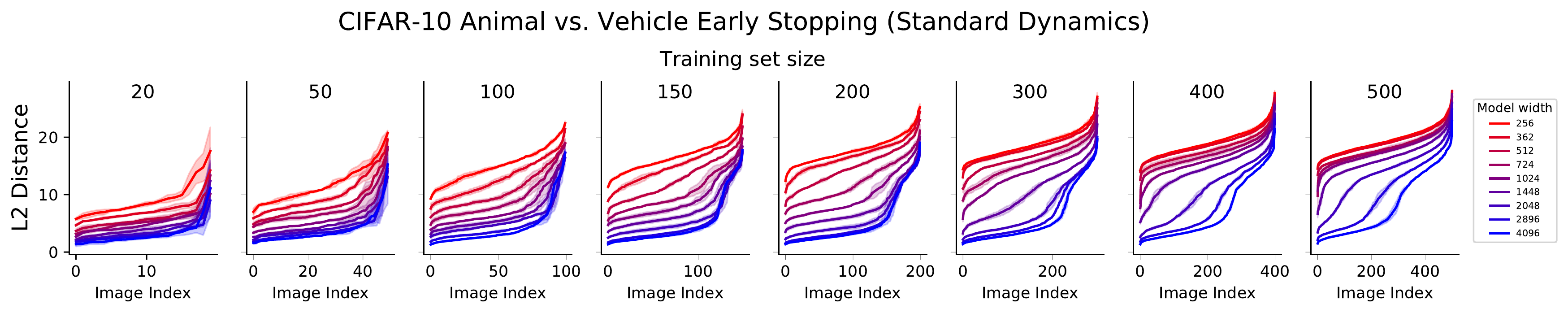}
\includegraphics[width = 0.9\linewidth]{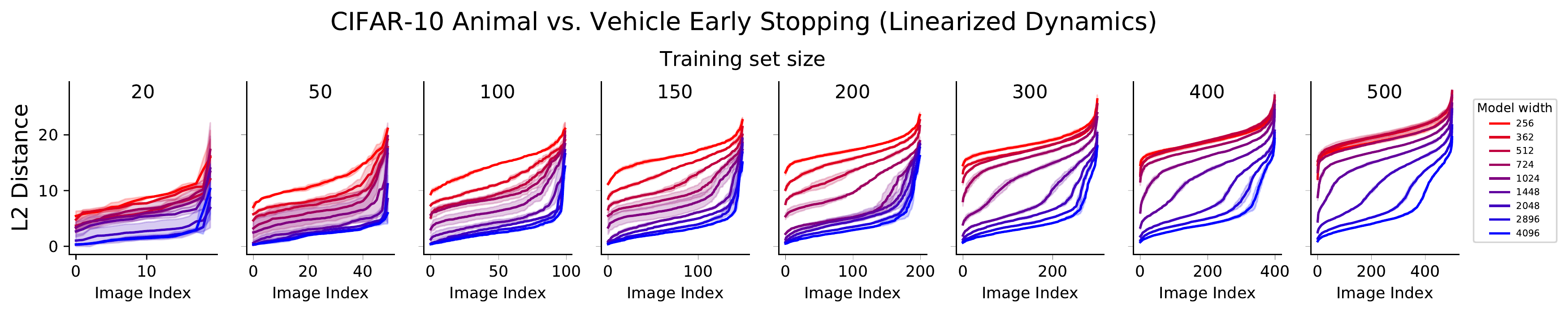}
\vskip -0.1in
\caption{Reconstruction curves for binary classification tasks with early stopping}
\label{app:fig:binary_recons_with_early_stop}
\end{center}
\vspace{-6mm}
\end{figure*}

\begin{figure*}[h]
\vskip 0.1in
\begin{center}
\includegraphics[width = 0.9\linewidth]{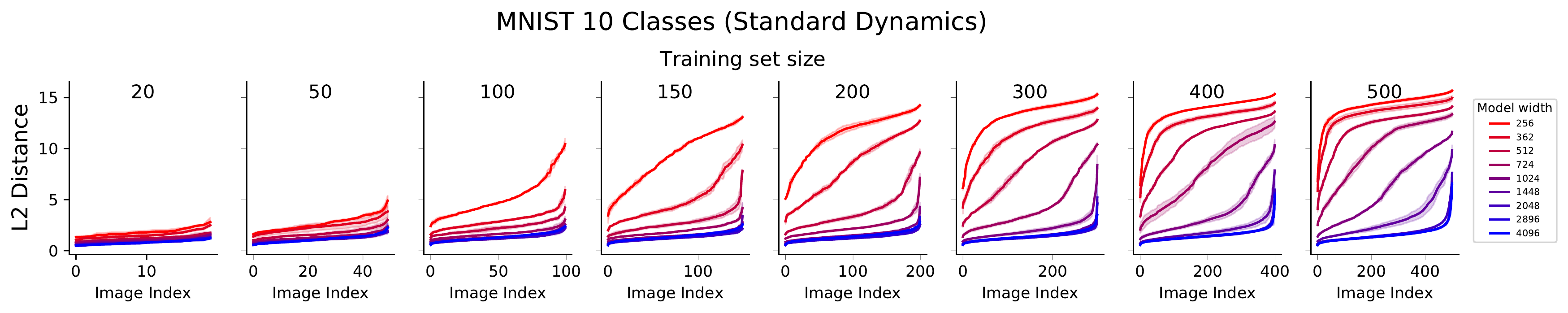}
\includegraphics[width = 0.9\linewidth]{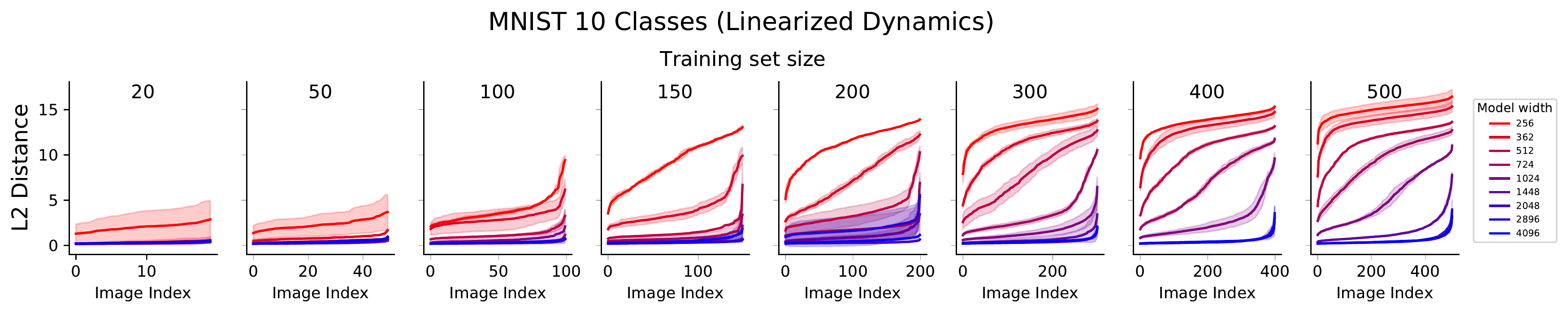}
\includegraphics[width = 0.9\linewidth]{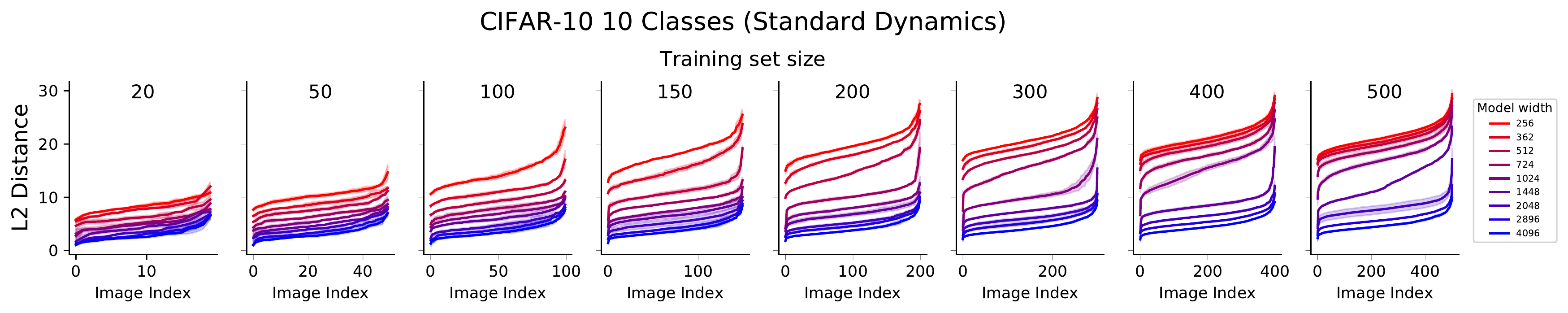}
\includegraphics[width = 0.9\linewidth]{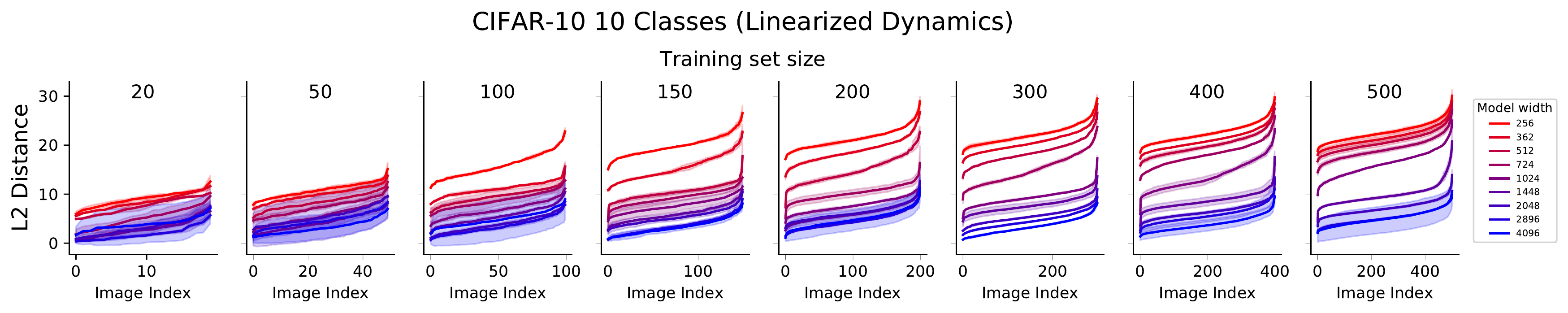}
\vskip -0.1in
\caption{Reconstruction curves for multiclass classification}
\label{app:fig:multiclass_recons}
\end{center}
\vspace{-6mm}
\end{figure*}

We show additional reconstruction curves for all dataset sizes in $[20, 50, 100, 150, 200, 300, 400, 500]$ for MNIST Odd vs. Even and CIFAR-10 Animal vs. Vehicle in \cref{app:fig:binary_recons}. We show the same reconstruction curves with the distillation reconstructions in \cref{app:fig:distilled_curves}. \cref{app:fig:binary_recons_with_early_stop} shows the same reconstruction curves for early stopping. Finally, \cref{app:fig:multiclass_recons} shows the same curves for multi-class classification.

\subsection{Reconstruction Images}

Here we show all the reconstruction images and their nearest training images in terms of $L_2$ distance. Images are sorted based on their rank in the reconstruction curve.

\subsubsection{Binary Classification}
\begin{figure*}[h]
\vskip 0.1in
\begin{center}
\includegraphics[width = 0.6\linewidth]{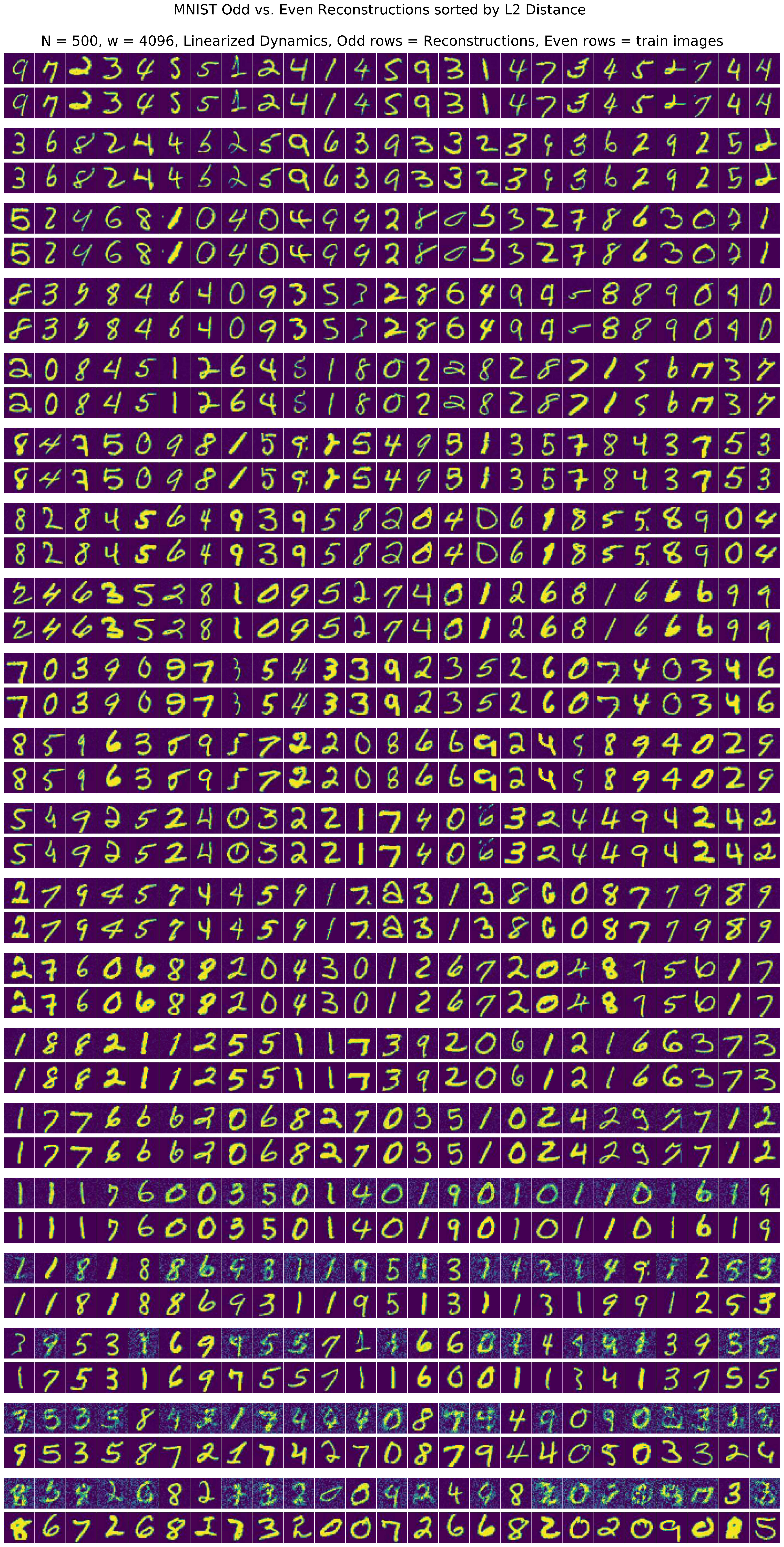}
\vskip -0.1in
\caption{Reconstructions for MNIST Odd vs. Even, Linearized Dynamics, 4096 width.}
\label{app:fig:recon_show_start_binary}
\end{center}
\vspace{-6mm}
\end{figure*}

\begin{figure*}[h]
\vskip 0.1in
\begin{center}
\includegraphics[width = 0.6\linewidth]{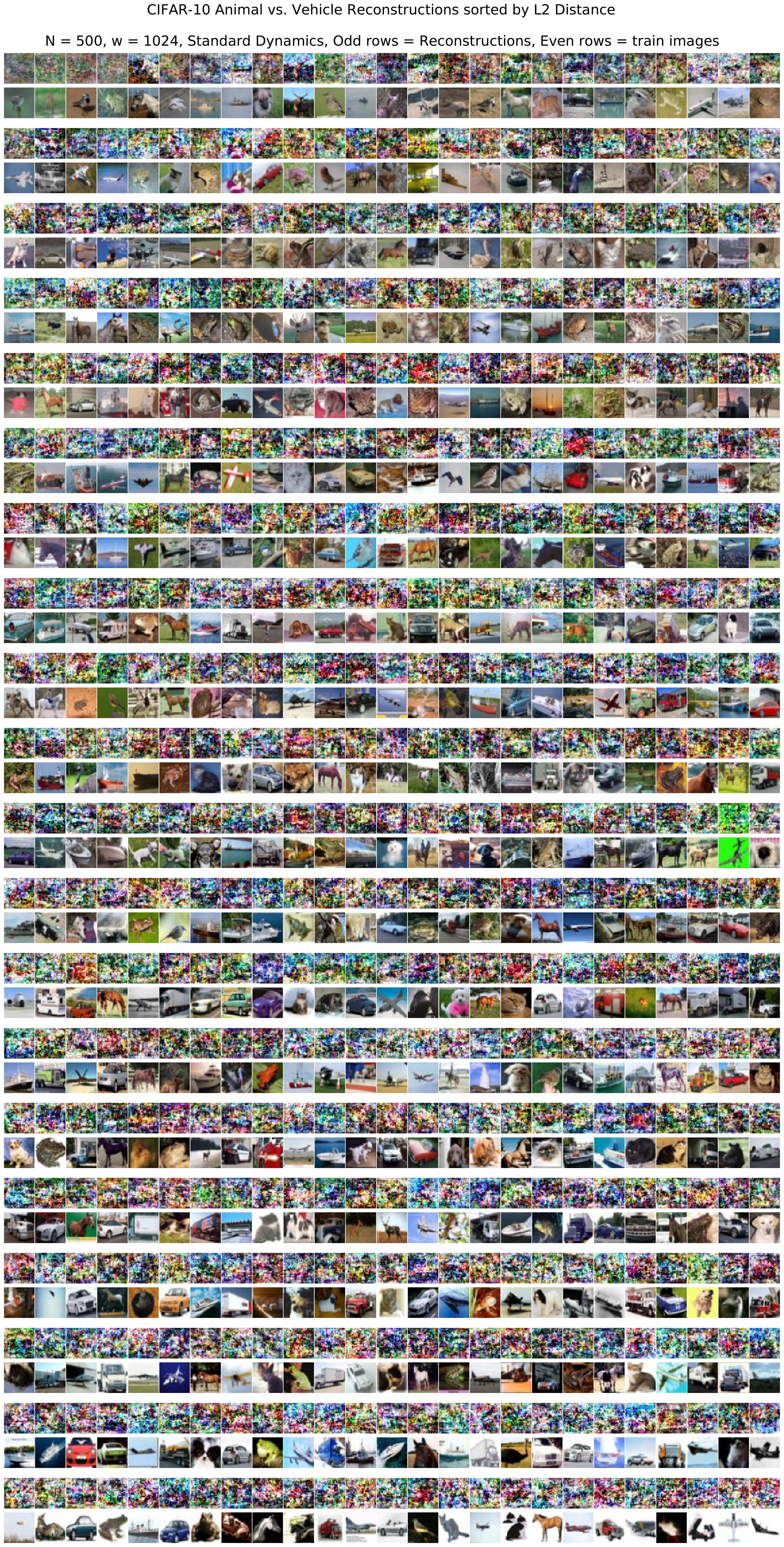}
\vskip -0.1in
\caption{Reconstructions for CIFAR-10 Animal vs. Vehicle, Standard Dynamics, 1024 width.}
\label{app:fig:recon_show_end_binary}
\end{center}
\vspace{-6mm}
\end{figure*}

We show the reconstruction curves for MNIST Odd vs. Even and CIFAR-10 Animal vs. Vehicle tasks for width 4096 and 1024 networks with linearized or standard dynamics in figures \ref{app:fig:recon_show_start_binary} to \ref{app:fig:recon_show_end_binary}.

\begin{figure*}[h]
\vskip 0.1in
\begin{center}
\includegraphics[width = 0.6\linewidth]{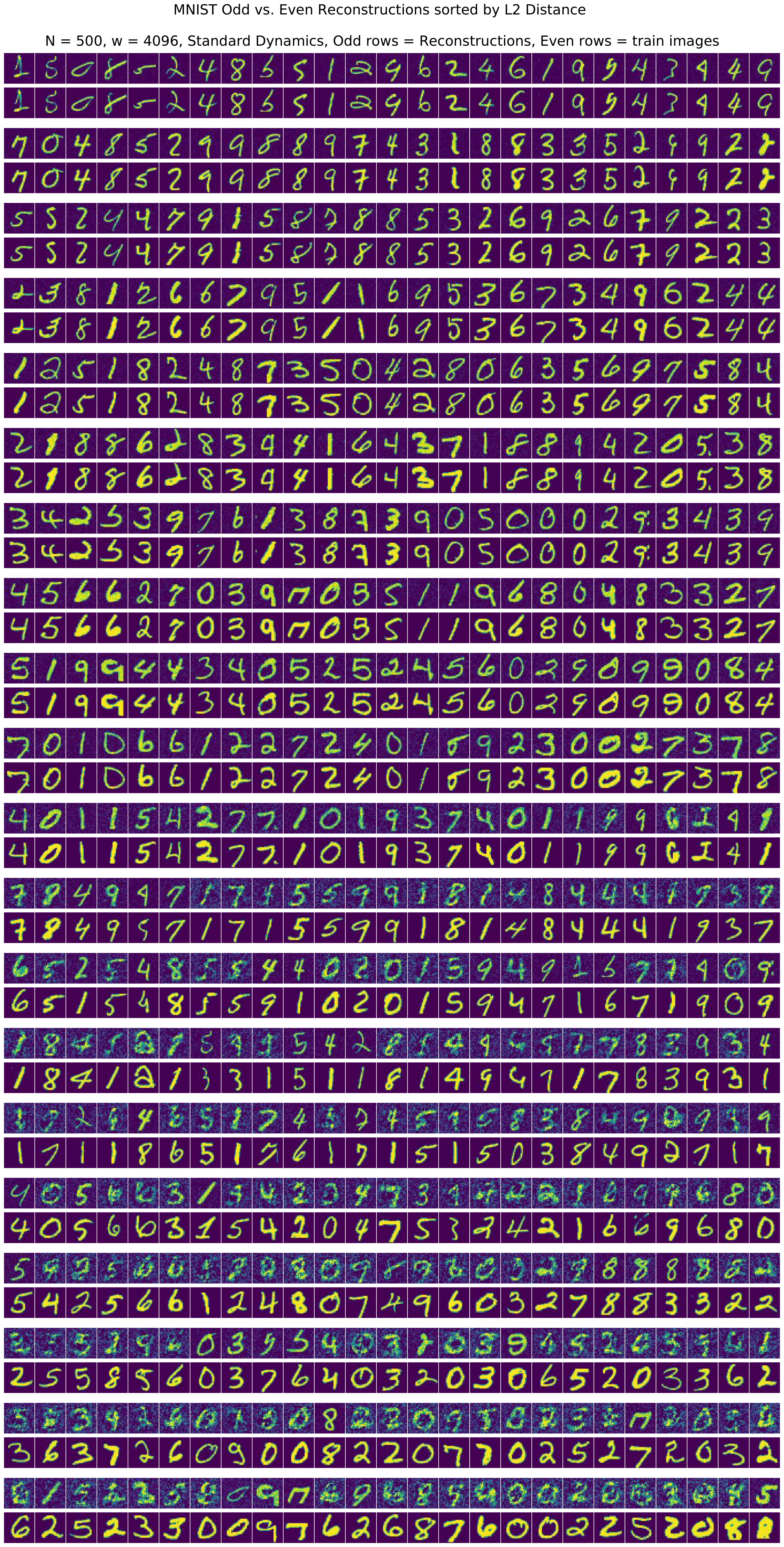}
\vskip -0.1in
\caption{Reconstructions for MNIST Odd vs. Even, Standard Dynamics, 4096 width.}
\end{center}
\vspace{-6mm}
\end{figure*}

\begin{figure*}[h]
\vskip 0.1in
\begin{center}
\includegraphics[width = 0.6\linewidth]{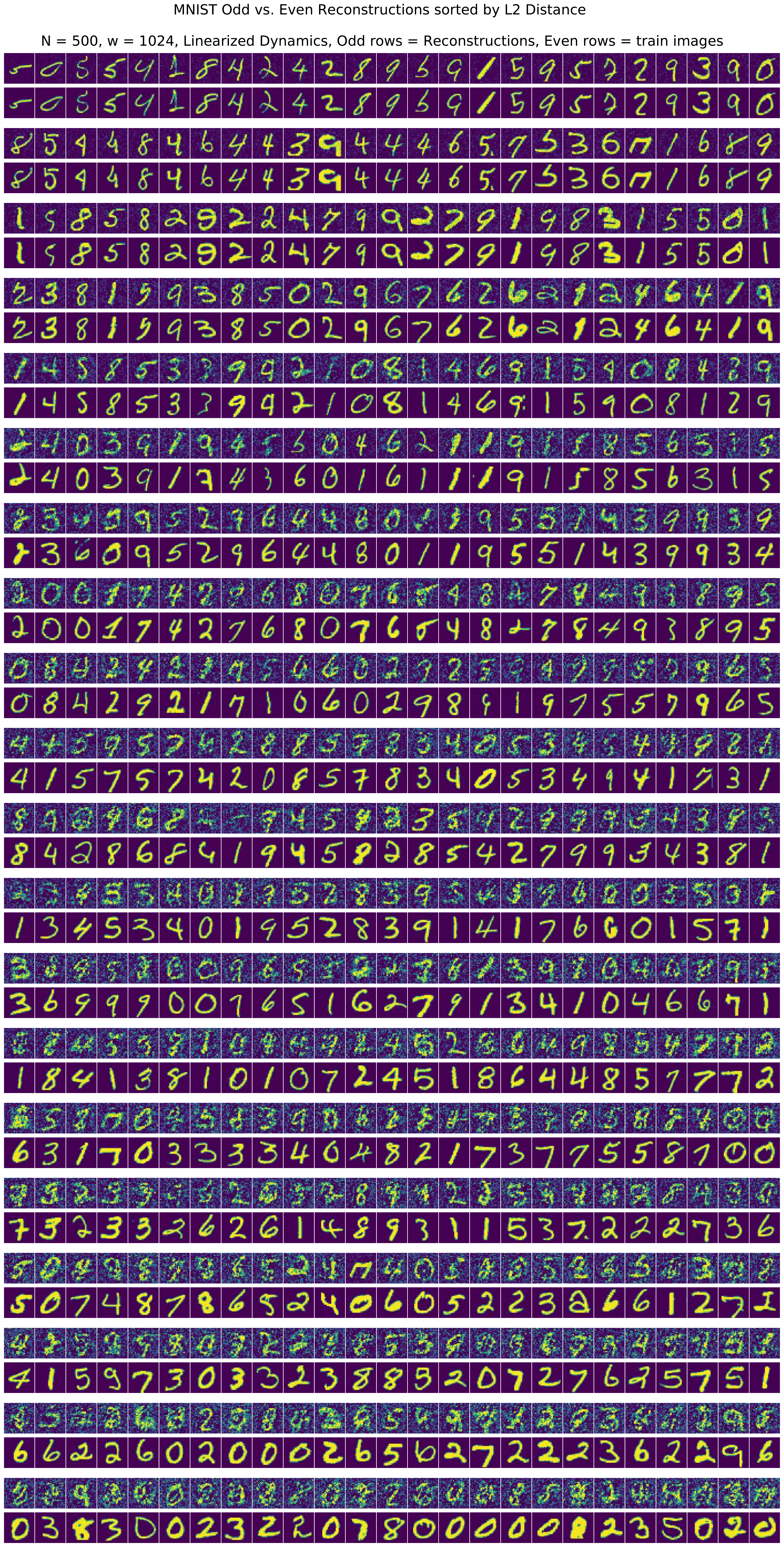}
\vskip -0.1in
\caption{Reconstructions for MNIST Odd vs. Even, Linearized Dynamics, 1024 width.}
\end{center}
\vspace{-6mm}
\end{figure*}

\begin{figure*}[h]
\vskip 0.1in
\begin{center}
\includegraphics[width = 0.6\linewidth]{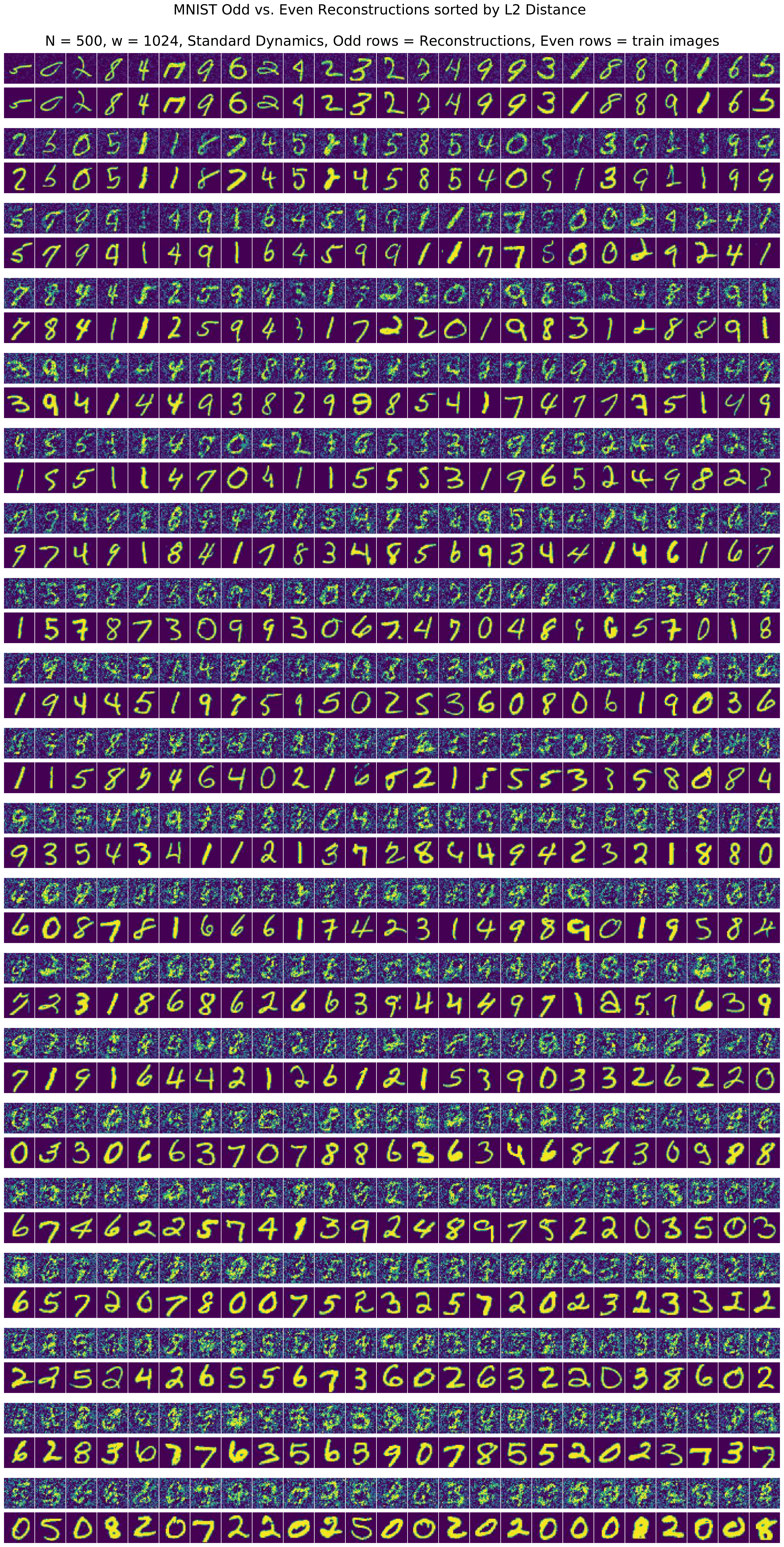}
\vskip -0.1in
\caption{Reconstructions for MNIST Odd vs. Even, Standard Dynamics, 1024 width.}
\end{center}
\vspace{-6mm}
\end{figure*}

\begin{figure*}[h]
\vskip 0.1in
\begin{center}
\includegraphics[width = 0.6\linewidth]{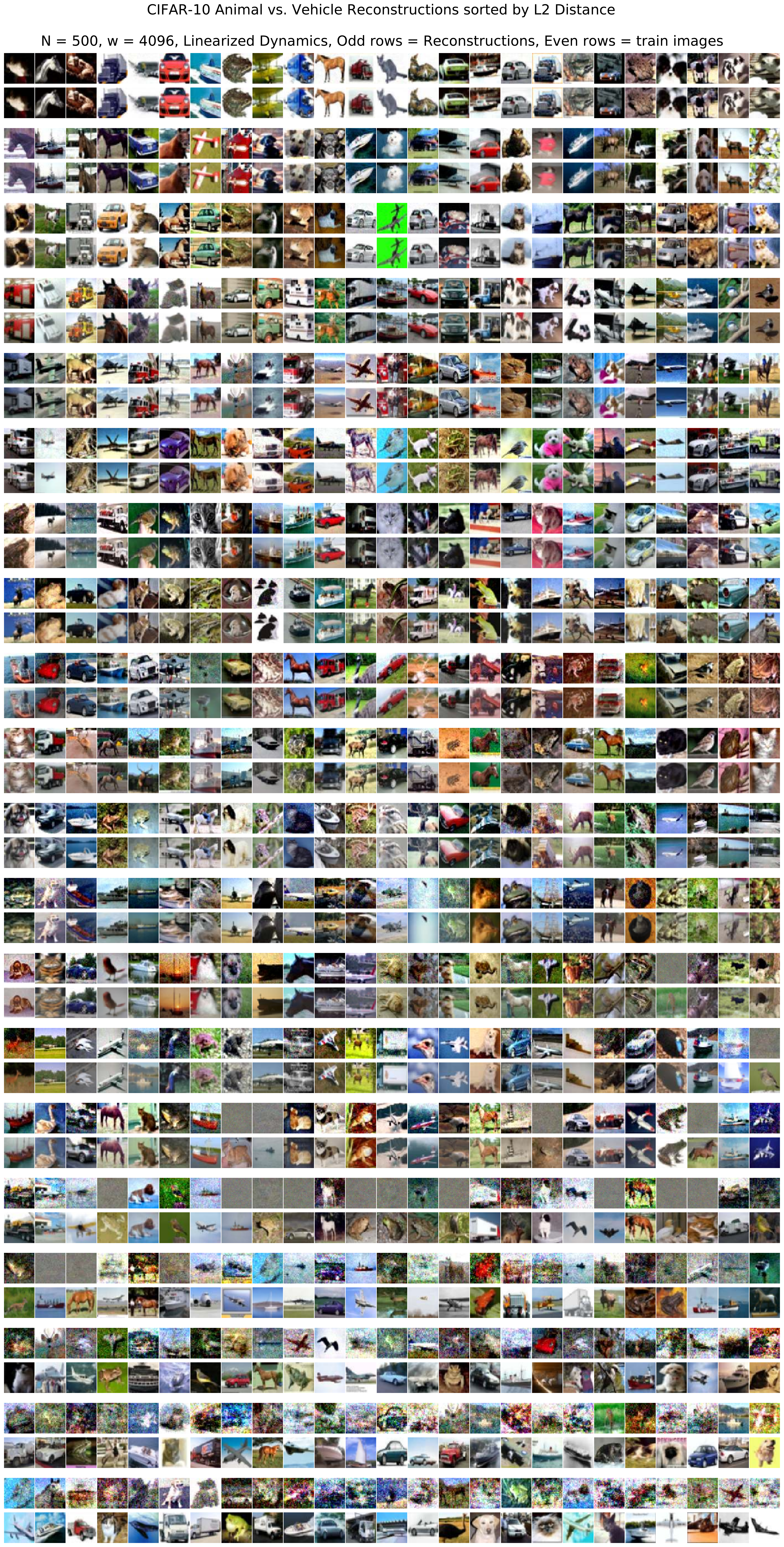}
\vskip -0.1in
\caption{Reconstructions for CIFAR-10 Animal vs. Vehicle, Linearized Dynamics, 4096 width.}
\end{center}
\vspace{-6mm}
\end{figure*}

\begin{figure*}[h]
\vskip 0.1in
\begin{center}
\includegraphics[width = 0.6\linewidth]{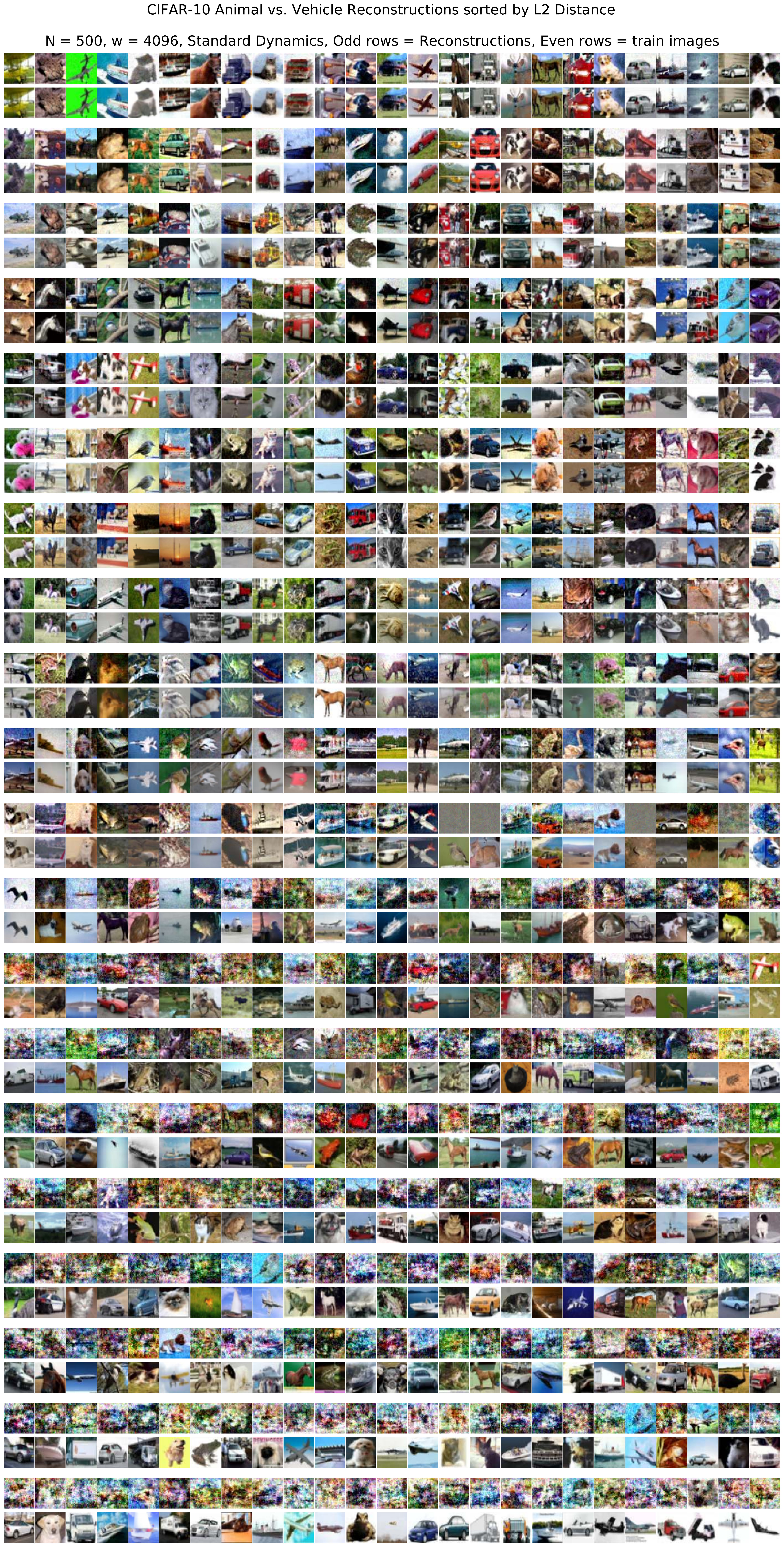}
\vskip -0.1in
\caption{Reconstructions for CIFAR-10 Animal vs. Vehicle, Standard Dynamics, 4096 width.}
\end{center}
\vspace{-6mm}
\end{figure*}

\begin{figure*}[h]
\vskip 0.1in
\begin{center}
\includegraphics[width = 0.6\linewidth]{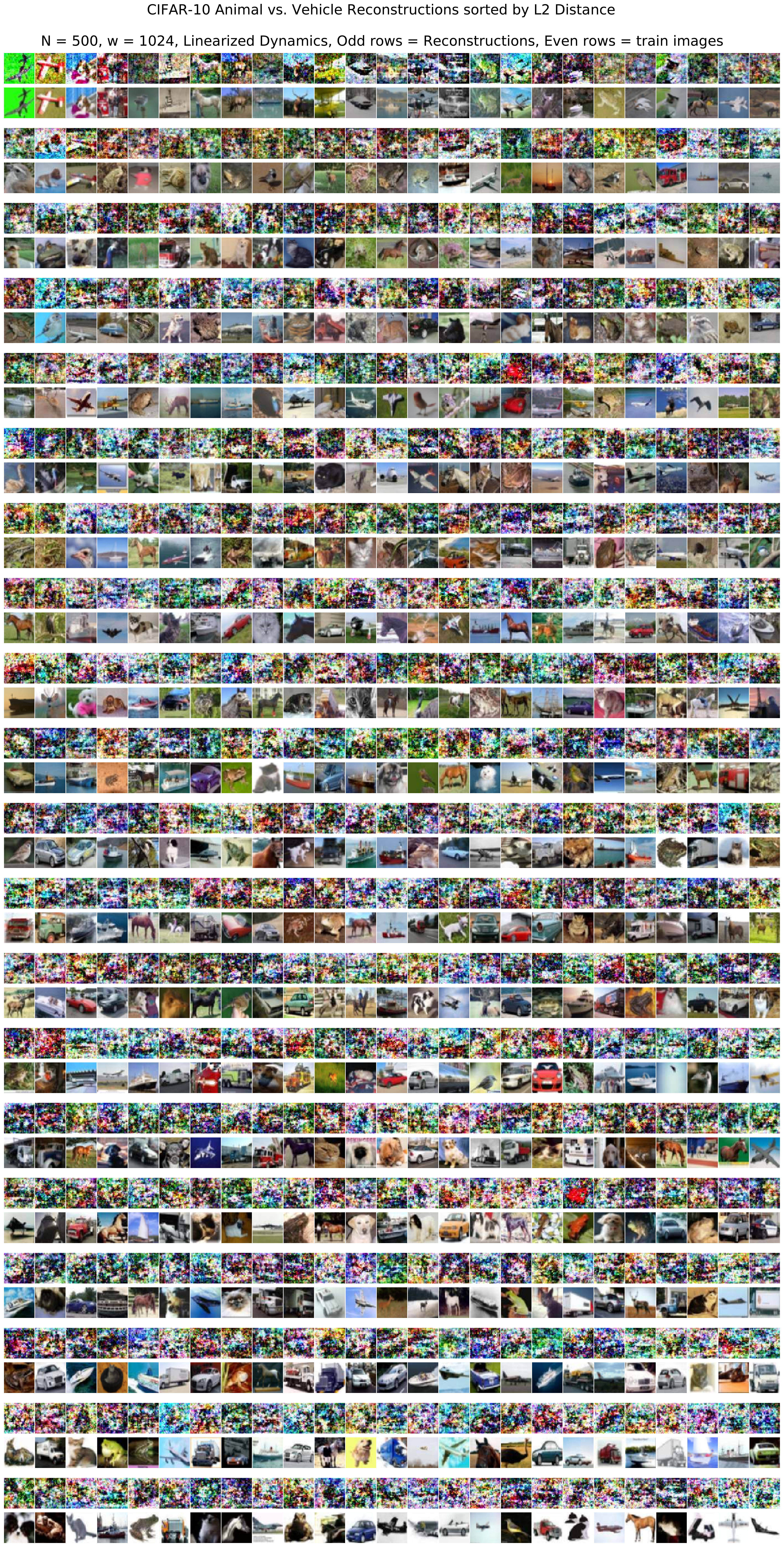}
\vskip -0.1in
\caption{Reconstructions for CIFAR-10 Animal vs. Vehicle, Linearized Dynamics, 1024 width.}
\end{center}
\vspace{-6mm}
\end{figure*}

\subsubsection{Multiclass Classification}
\begin{figure*}[h]
\vskip 0.1in
\begin{center}
\includegraphics[width = 0.6\linewidth]{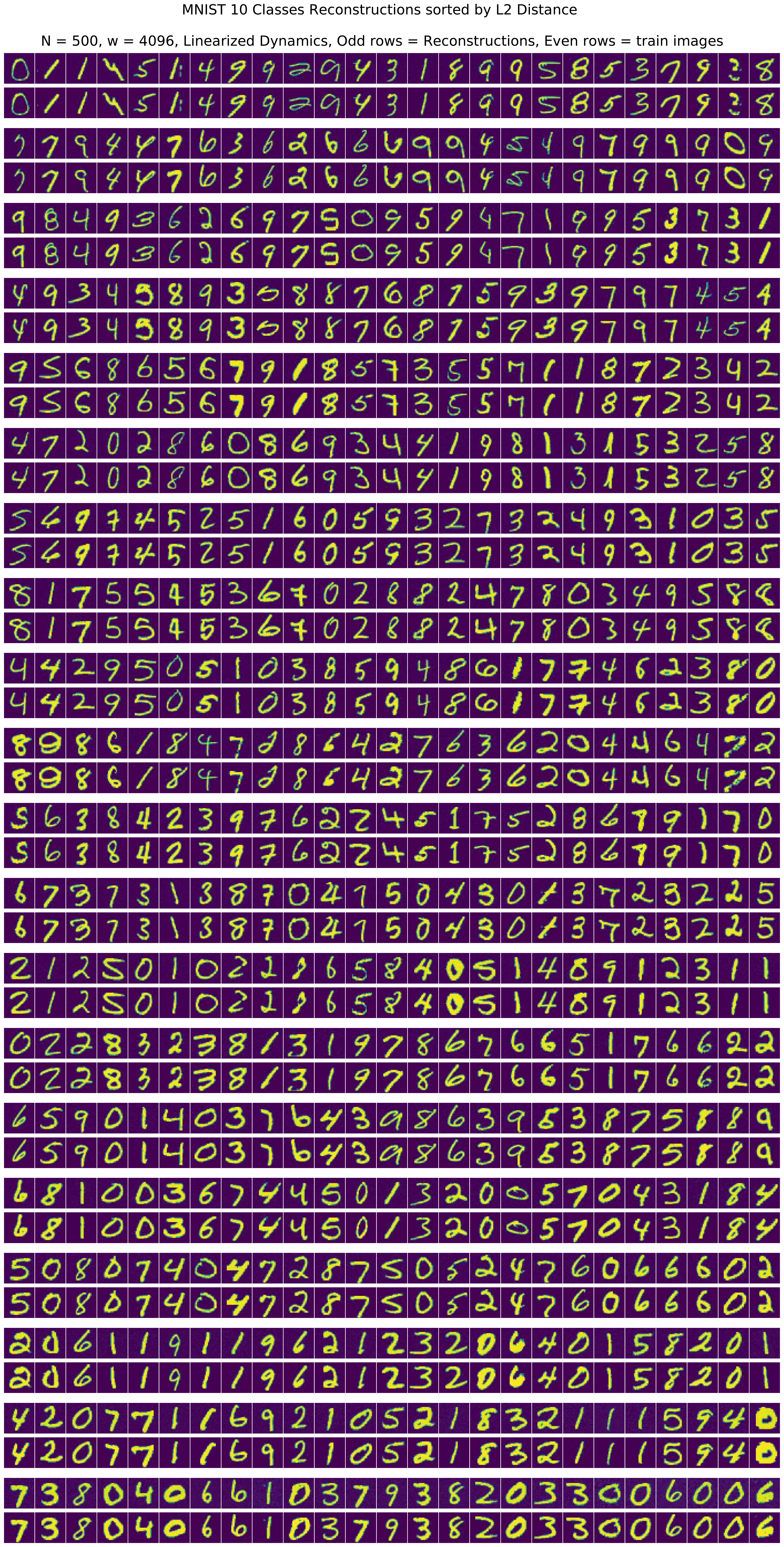}
\vskip -0.1in
\caption{Reconstructions for MNIST 10 Classes, Linearized Dynamics, 4096 width.}
\label{app:fig:recon_show_start_multiclass}
\end{center}
\vspace{-6mm}
\end{figure*}

\begin{figure*}[h]
\vskip 0.1in
\begin{center}
\includegraphics[width = 0.6\linewidth]{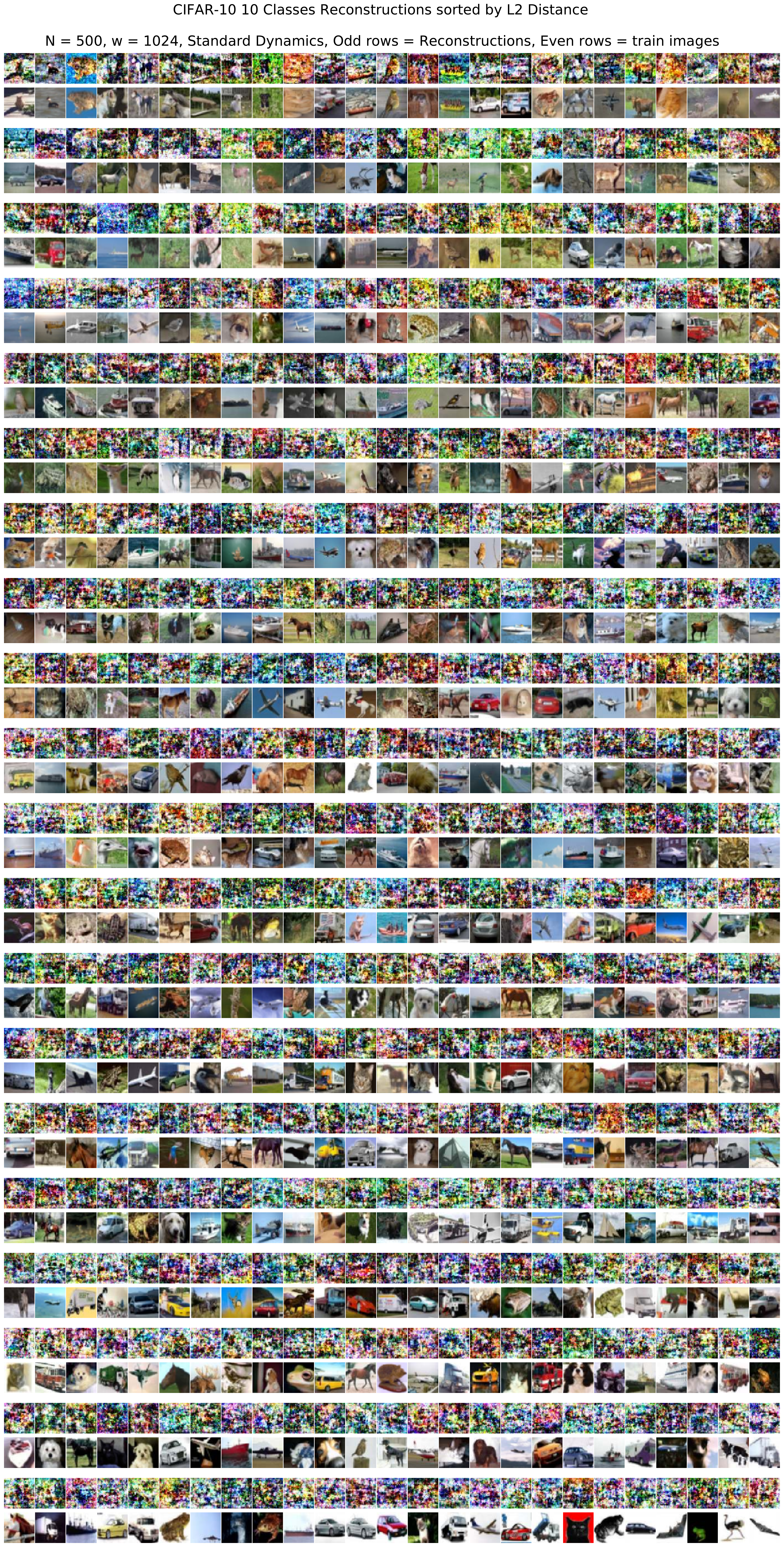}
\vskip -0.1in
\caption{Reconstructions for CIFAR-10 10 Classes, Standard Dynamics, 1024 width.}
\label{app:fig:recon_show_end_multiclass}
\end{center}
\vspace{-6mm}
\end{figure*}

We show the reconstruction curves for MNIST and CIFAR-10 10-way classification for width 4096 and 1024 networks with linearized or standard dyanmics in figures \ref{app:fig:recon_show_start_multiclass} to \ref{app:fig:recon_show_end_multiclass}.

\begin{figure*}[h]
\vskip 0.1in
\begin{center}
\includegraphics[width = 0.6\linewidth]{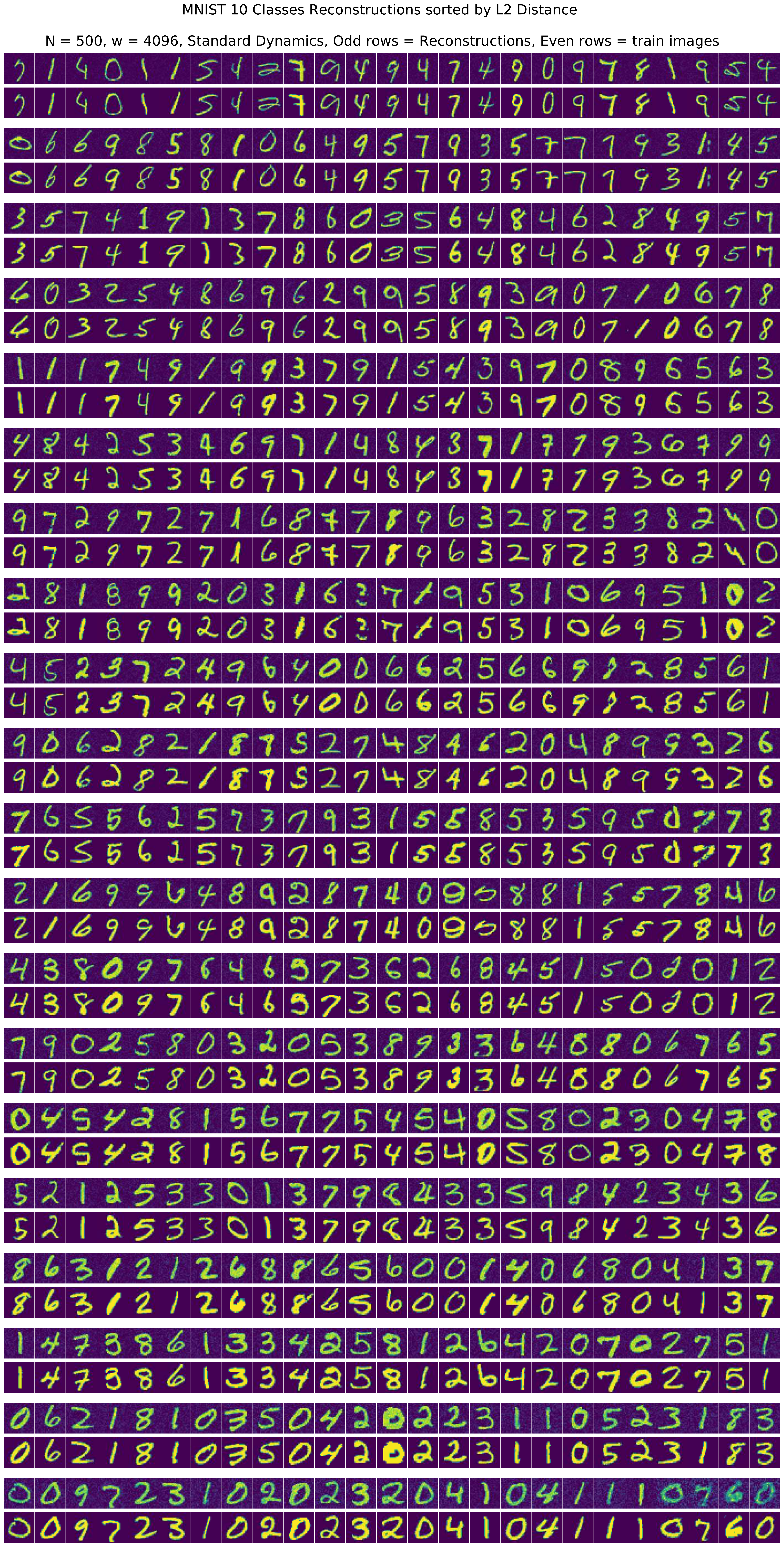}
\vskip -0.1in
\caption{Reconstructions for MNIST 10 Classes, Standard Dynamics, 4096 width.}
\end{center}
\vspace{-6mm}
\end{figure*}

\begin{figure*}[h]
\vskip 0.1in
\begin{center}
\includegraphics[width = 0.6\linewidth]{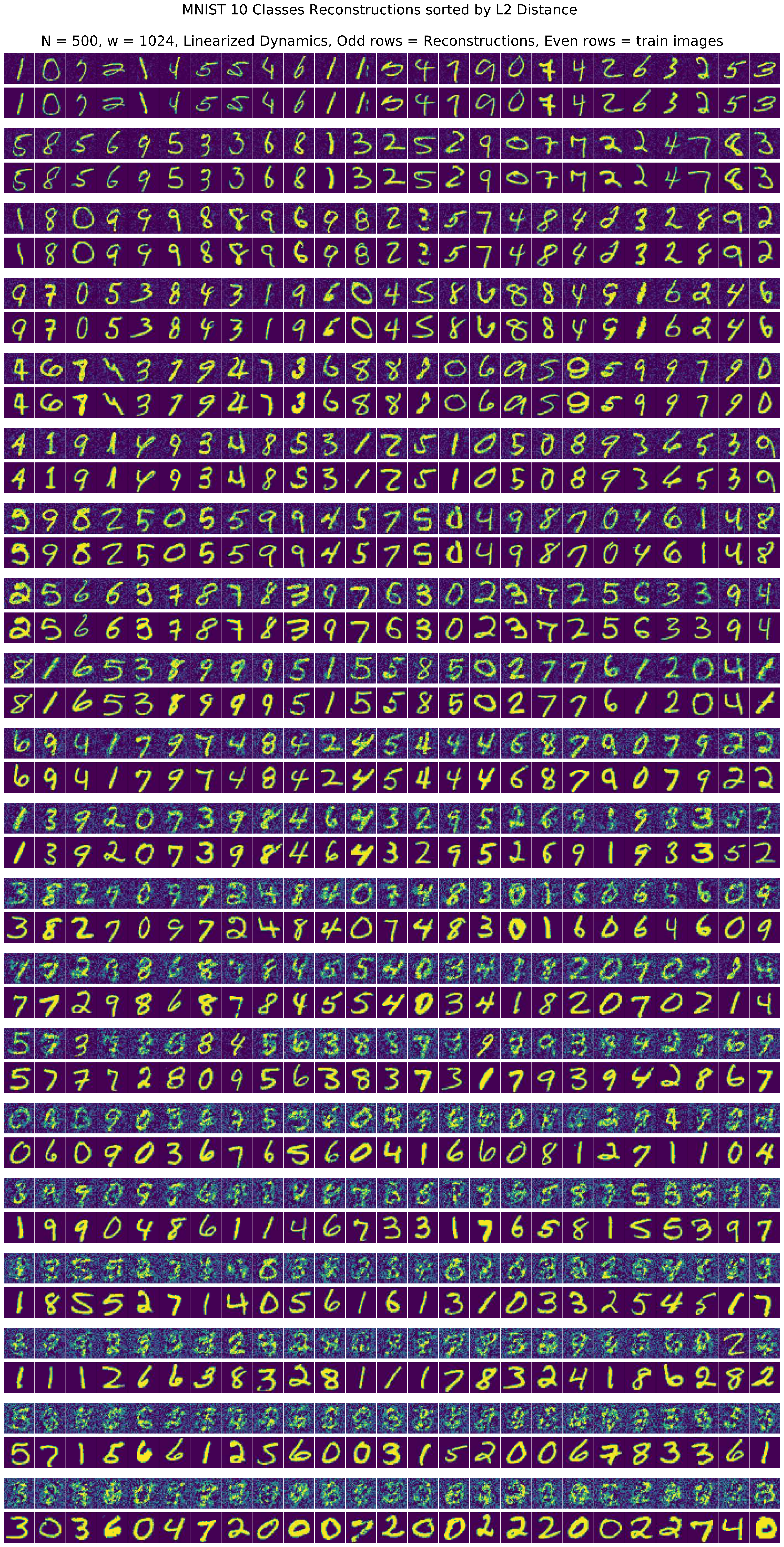}
\vskip -0.1in
\caption{Reconstructions for MNIST 10 Classes, Linearized Dynamics, 1024 width.}
\end{center}
\vspace{-6mm}
\end{figure*}

\begin{figure*}[h]
\vskip 0.1in
\begin{center}
\includegraphics[width = 0.6\linewidth]{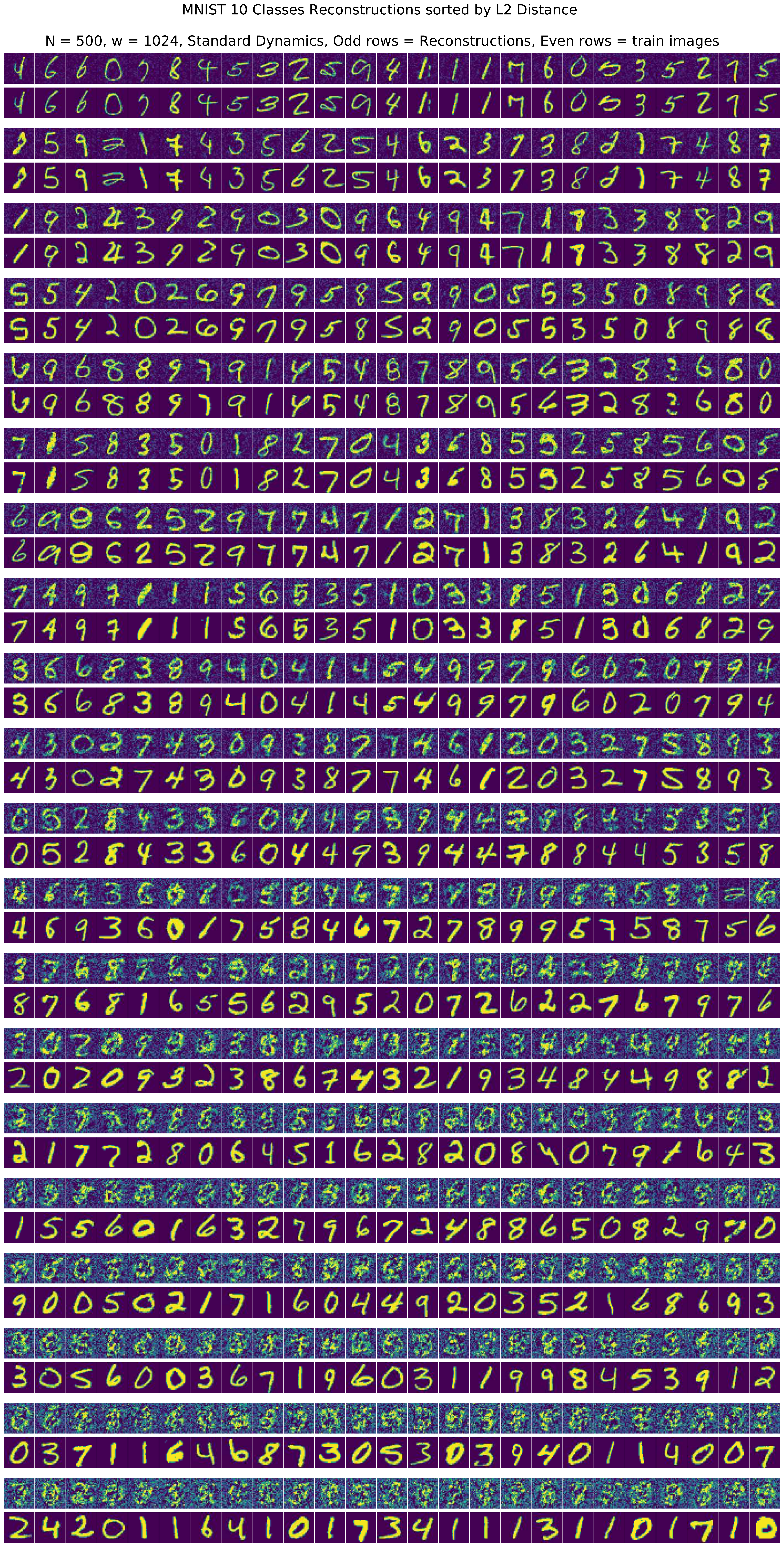}
\vskip -0.1in
\caption{Reconstructions for MNIST 10 Classes, Standard Dynamics, 1024 width.}
\end{center}
\vspace{-6mm}
\end{figure*}

\begin{figure*}[h]
\vskip 0.1in
\begin{center}
\includegraphics[width = 0.6\linewidth]{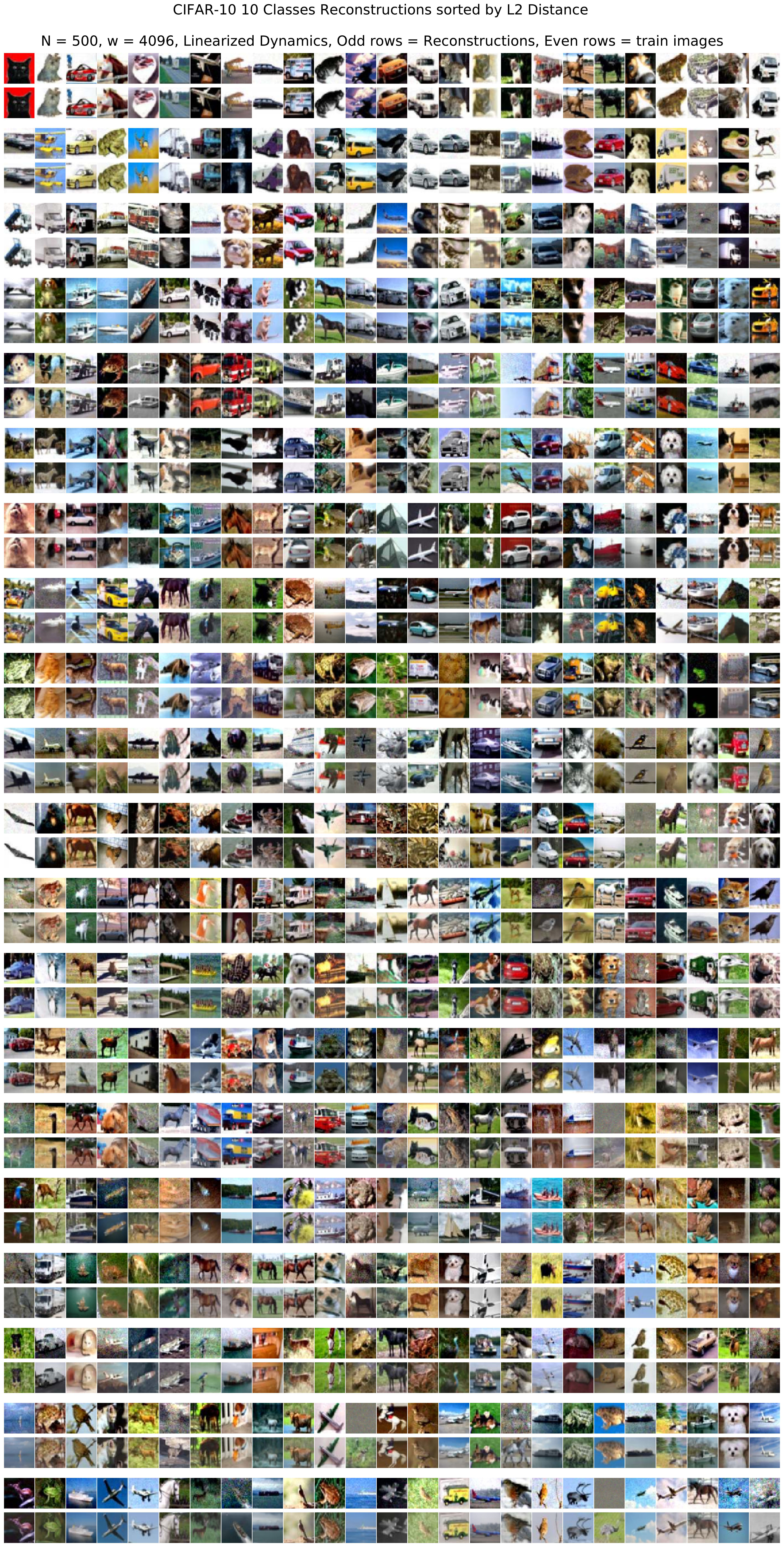}
\vskip -0.1in
\caption{Reconstructions for CIFAR-10 10 Classes, Linearized Dynamics, 4096 width.}
\end{center}
\vspace{-6mm}
\end{figure*}

\begin{figure*}[h]
\vskip 0.1in
\begin{center}
\includegraphics[width = 0.6\linewidth]{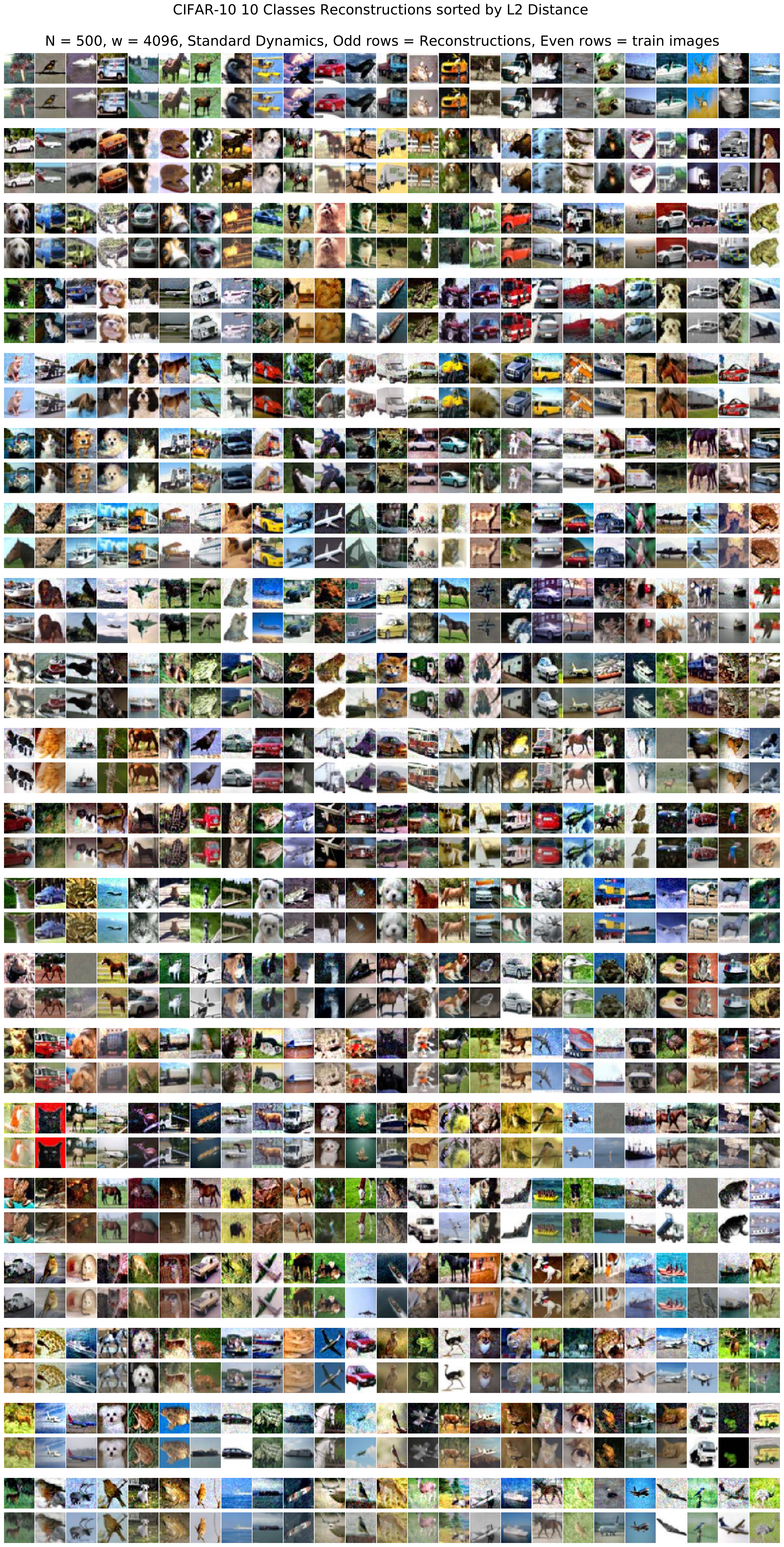}
\vskip -0.1in
\caption{Reconstructions for CIFAR-10 10 Classes, Standard Dynamics, 4096 width.}
\end{center}
\vspace{-6mm}
\end{figure*}

\begin{figure*}[h]
\vskip 0.1in
\begin{center}
\includegraphics[width = 0.6\linewidth]{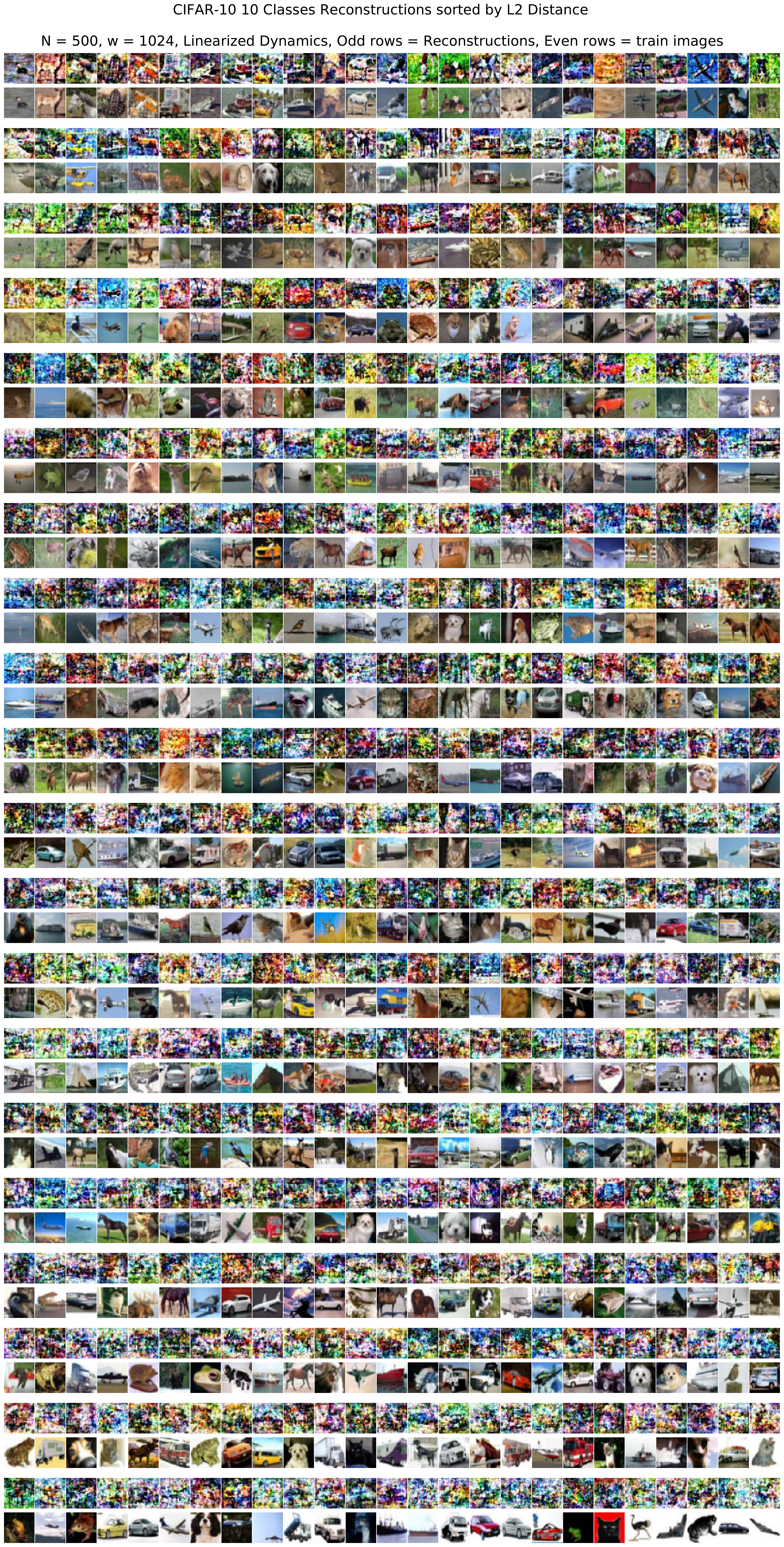}
\vskip -0.1in
\caption{Reconstructions for CIFAR-10 A10 Classes, Linearized Dynamics, 1024 width.}
\end{center}
\vspace{-6mm}
\end{figure*}

\end{document}